\documentclass[10pt,twocolumn,letterpaper]{article}

\usepackage{cvpr}

\makeatletter
\@namedef{ver@everyshi.sty}{}
\makeatother
\usepackage{float}
\usepackage{graphicx}
\usepackage{amsmath}
\usepackage{amssymb}
\usepackage{booktabs}
\usepackage{comment}
\usepackage{color}
\usepackage{tabularx}
\usepackage{colortbl}
\usepackage{tabu}
\usepackage{multirow}
\usepackage{setspace}
\usepackage{cuted}
\usepackage[font=small,labelsep=period]{caption}
\usepackage{fontawesome5} %
\usepackage[usenames,dvipsnames]{xcolor}
\usepackage{tikz}
\usepackage{pgfplots}
\usepackage{pgfkeys}
\usepackage{xcolor}

\usepackage[accsupp]{axessibility} %
\usepackage[moderate]{savetrees}

\DeclareCaptionLabelFormat{tab}{Tab.~#2}
\DeclareCaptionLabelFormat{fig}{Fig.~#2}
\makeatletter
\renewcommand{\fnum@figure}{Figure \thefigure}
\makeatother

\newcolumntype{Y}{>{\centering\arraybackslash}X}
\newcolumntype{R}{>{\raggedleft\arraybackslash}X}
\newcolumntype{L}{>{\raggedright\arraybackslash}X}

\newcommand{\mytilde}{\raise.17ex\hbox{$\scriptstyle\sim$}}

\definecolor{bblue}{rgb}{0.0,0.2,0.8}
\definecolor{ccol}{rgb}{0.91,0.91,0.91}

\definecolor{avgcol}{rgb}{1.0,0.91,0.722}
\definecolor{arcol}{rgb}{0.765,0.878,0.812}
\definecolor{timecol}{rgb}{0.941,0.749,0.737}

\definecolor{awardcol}{rgb}{0.765,0.878,0.812}

\usepackage{xspace}

\newcommand\customparagraph[1]{\vspace{0.5em}\noindent\textbf{#1}}

\def\addlegendimage{\csname pgfplots@addlegendimage\endcsname}

\newcommand{\iconBest}{\textcolor{darkgray}{\faCrown}}
\newcommand{\iconFast}{\textcolor{darkgray}{\faBolt}}
\newcommand{\iconOpen}{\textcolor{darkgray}{\faGithub}}
\newcommand{\iconRGB}{\textcolor{darkgray}{\faImage}}
\newcommand{\iconDefault}{\textcolor{darkgray}{\faCog}}

\usepackage[pagebackref=true,breaklinks=true,colorlinks=true,urlcolor=bblue,bookmarks=false,citecolor=bblue]{hyperref}

\usepackage[capitalize]{cleveref}
\crefname{section}{Sec.}{Secs.}
\Crefname{section}{Section}{Sections}
\Crefname{table}{Table}{Tables}
\crefname{table}{Tab.}{Tabs.}

\usepackage[percent]{overpic}

\def\cvprPaperID{8}
\def\confName{CVPR}
\def\confYear{2025}

\begin{document}

\def\greencheckmark{\textcolor{darkgreen}{\checkmark}}
\def\redxmark{\textcolor{darkred}{\text{\ding{55}}}}  %

\definecolor{darkgreen}{RGB}{0,110,0}
\definecolor{darkred}{RGB}{170,0,0}
\definecolor{DarkMagenta}{rgb}{0.7, 0.0, 0.7}
\definecolor{DarkOrange}{rgb}{1.0, 0.55, 0.0}
\newcommand{\nguyen}[1]{{\color{DarkMagenta}#1}}
\newcommand{\nguyenrmk}[1]{{\color{DarkMagenta} {\bf [VN: #1]}}}

\newcommand{\tom}[1]{{\color{DarkOrange}#1}}
\newcommand{\tomrmk}[1]{{\color{DarkOrange} {\bf [TH: #1]}}}

\newcommand{\martin}[1]{{\color{teal}#1}}
\newcommand{\martinrmk}[1]{{\color{teal} {\bf [BT: #1]}}}

\newcommand{\stephen}[1]{{\color{teal}#1}}
\newcommand{\stephenrmk}[1]{{\color{teal} {\bf [BT: #1]}}}

\newcommand{\taeyeop}[1]{{\color{pink}#1}}
\newcommand{\taeyeoprmk}[1]{{\color{pink} {\bf [BT: #1]}}}

\title{\mbox{BOP Challenge 2024 on Model-Based and Model-Free 6D Object Pose Estimation}}

\newcommand{\namesep}{\hspace{0.8em}}
\author{
Van Nguyen Nguyen$^{1}$\namesep
Stephen Tyree$^{2}$\namesep
Andrew Guo$^{3}$\namesep
M{\'e}d{\'e}ric Fourmy$^{4}$\namesep
Anas Gouda$^{5}$\namesep \\
Taeyeop Lee$^{6}$\namesep
Sungphill Moon$^{7}$\namesep 
Hyeontae Son$^{7}$\namesep
Lukas Ranftl$^{8, 9}$\namesep
Jonathan Tremblay$^{2}$\namesep \\
Eric Brachmann$^{10}$\namesep
Bertram Drost$^{8}$\namesep
Vincent Lepetit$^{1}$\namesep
Carsten Rother$^{11}$\namesep 
Stan Birchfield$^{2}$\namesep \\
Jiri~Matas$^{4}$\namesep
Yann Labb{\'e}$^{13}$\namesep
Martin Sundermeyer$^{12}$\namesep
Tomas~Hodan$^{13}$ \vspace{0.56em} \\
 {\normalsize
     {$^{1}$ENPC}\namesep
     {$^{2}$NVIDIA}\namesep
     {$^{3}$University of Toronto} \namesep
     {$^{4}$CTU Prague} \namesep
     {$^{5}$TU Dortmund} \namesep
     {$^{6}$KAIST} \namesep 
 } \\
 {\normalsize
     {$^{7}$NAVER LABS} \namesep
     {$^{8}$MVTec}\namesep
     {$^{9}$TU Munich}\namesep
     {$^{10}$Niantic}\namesep
     {$^{11}$Heidelberg University}\namesep
     {$^{12}$Google}\namesep
     {$^{13}$Meta}
}     
}

\maketitle

\begin{strip}
\begin{minipage}{\textwidth}\centering
\vspace{-33.5pt}

\begingroup
\centering
\scriptsize
\renewcommand{\arraystretch}{0.5}

\newcommand{\datasetsheight}{3.3cm}
\newcommand{\datasetswidth}{4.2cm}
\newcommand{\onboardingheight}{3.0cm}
\newcommand{\onboardingwidth}{4cm} %
\setlength{\tabcolsep}{1.5pt}
\newcommand{\framedimage}[2]{%
    \setlength{\fboxsep}{0pt}%
    \setlength{\fboxrule}{0.8pt}%
    \fbox{\includegraphics[width=\onboardingwidth, height=\onboardingheight, trim=#2, clip]{#1}}%
}

\begin{tabular}{m{3.5cm} m{4.4cm} m{4.4cm} m{0.4cm} m{3.5cm} m{1.5cm}} %
    \includegraphics[height=\datasetsheight, trim = 0 0 0 0, clip]{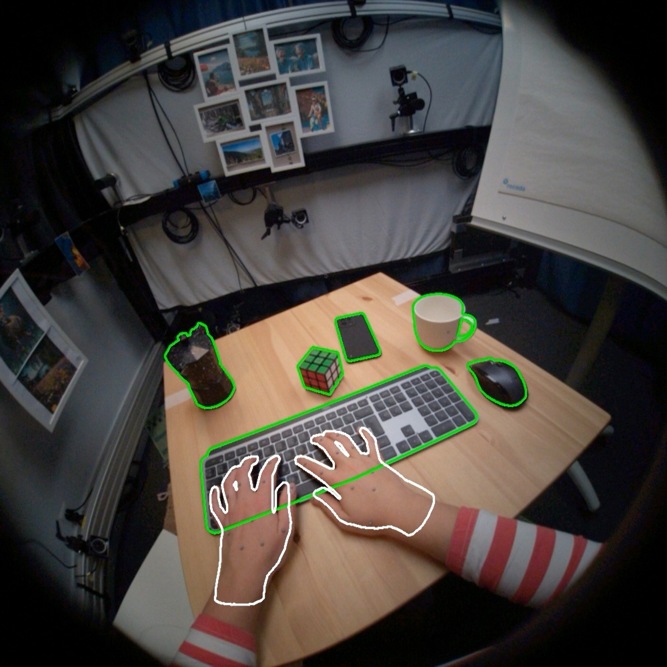} &
    \includegraphics[height=\datasetsheight,width=\datasetswidth, trim=120 0 120 0, clip]{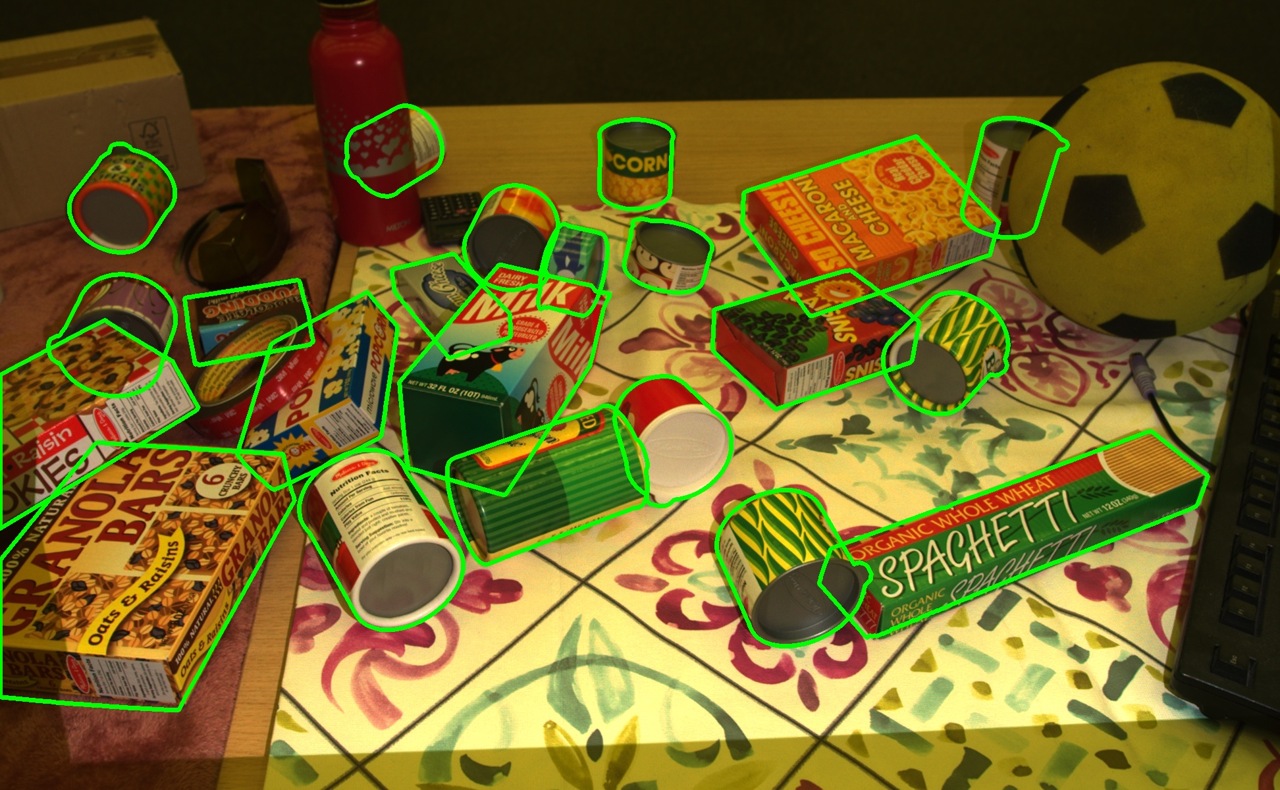} &
    \includegraphics[height=\datasetsheight,width=\datasetswidth]{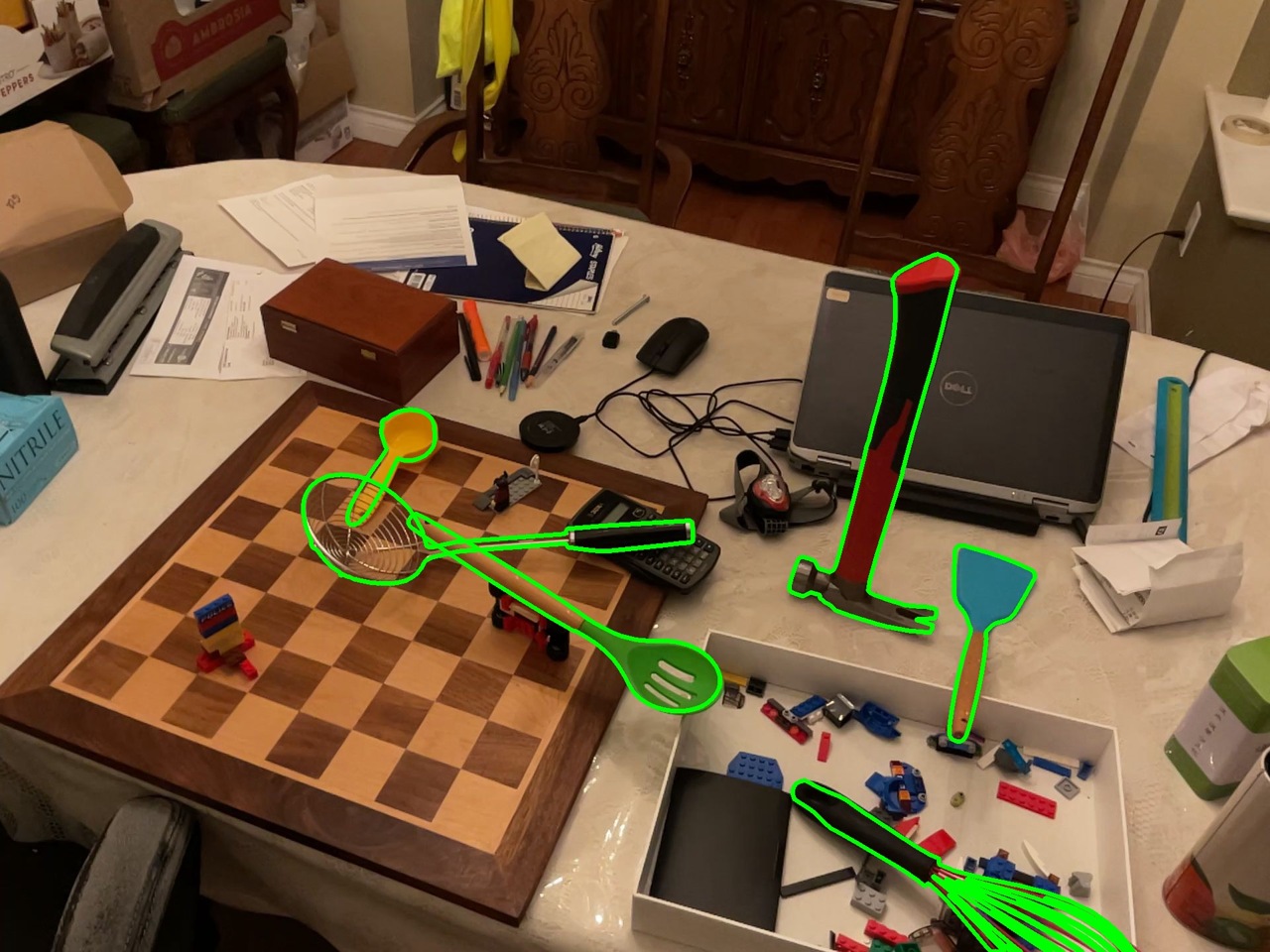} & &

    \vspace{0.1cm}
    \raisebox{0.3cm}{\begin{overpic}[width=\onboardingwidth, height=\onboardingheight, trim=0 0 0 0, clip]{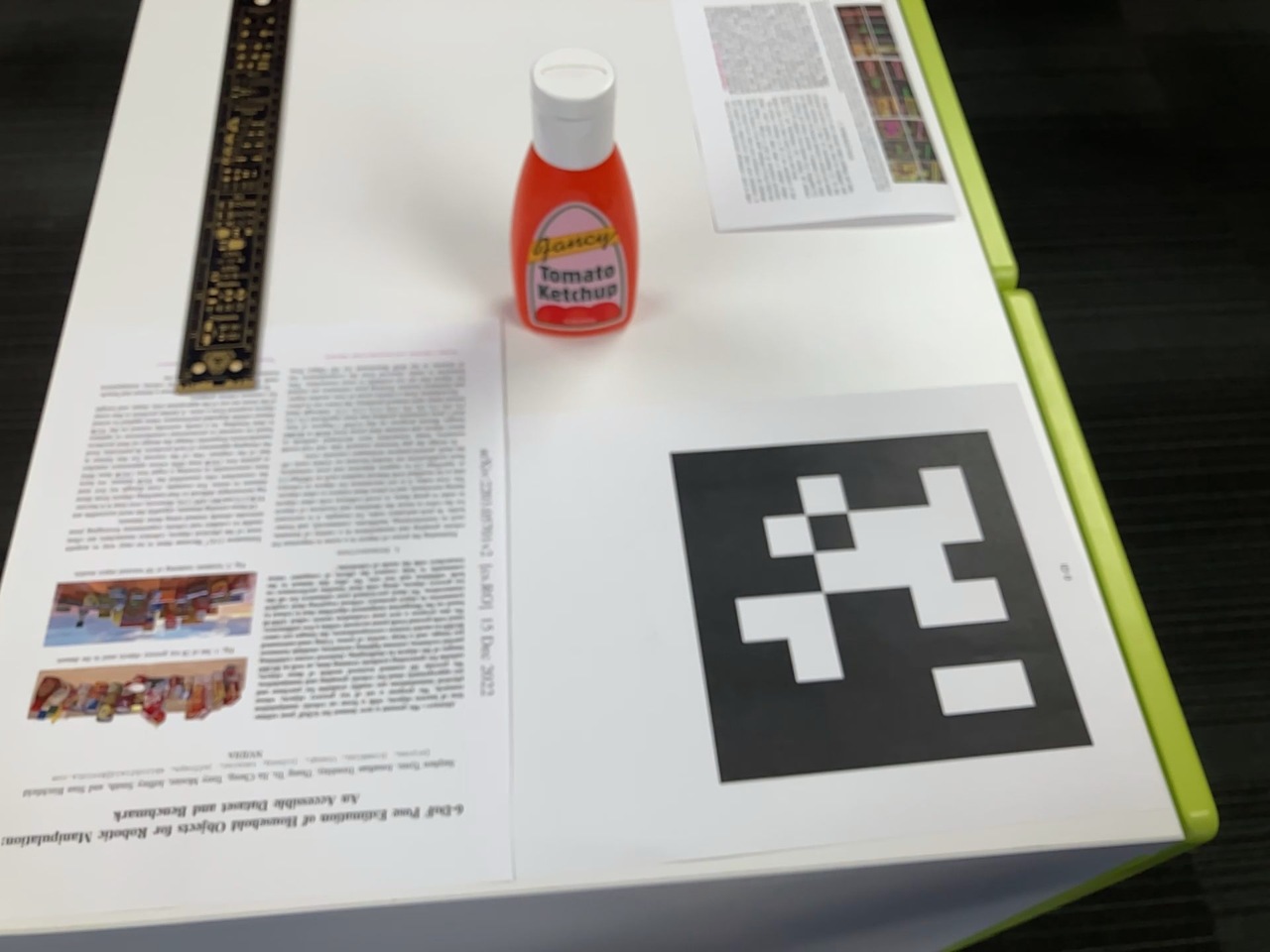}
        \put(0,0){\framedimage{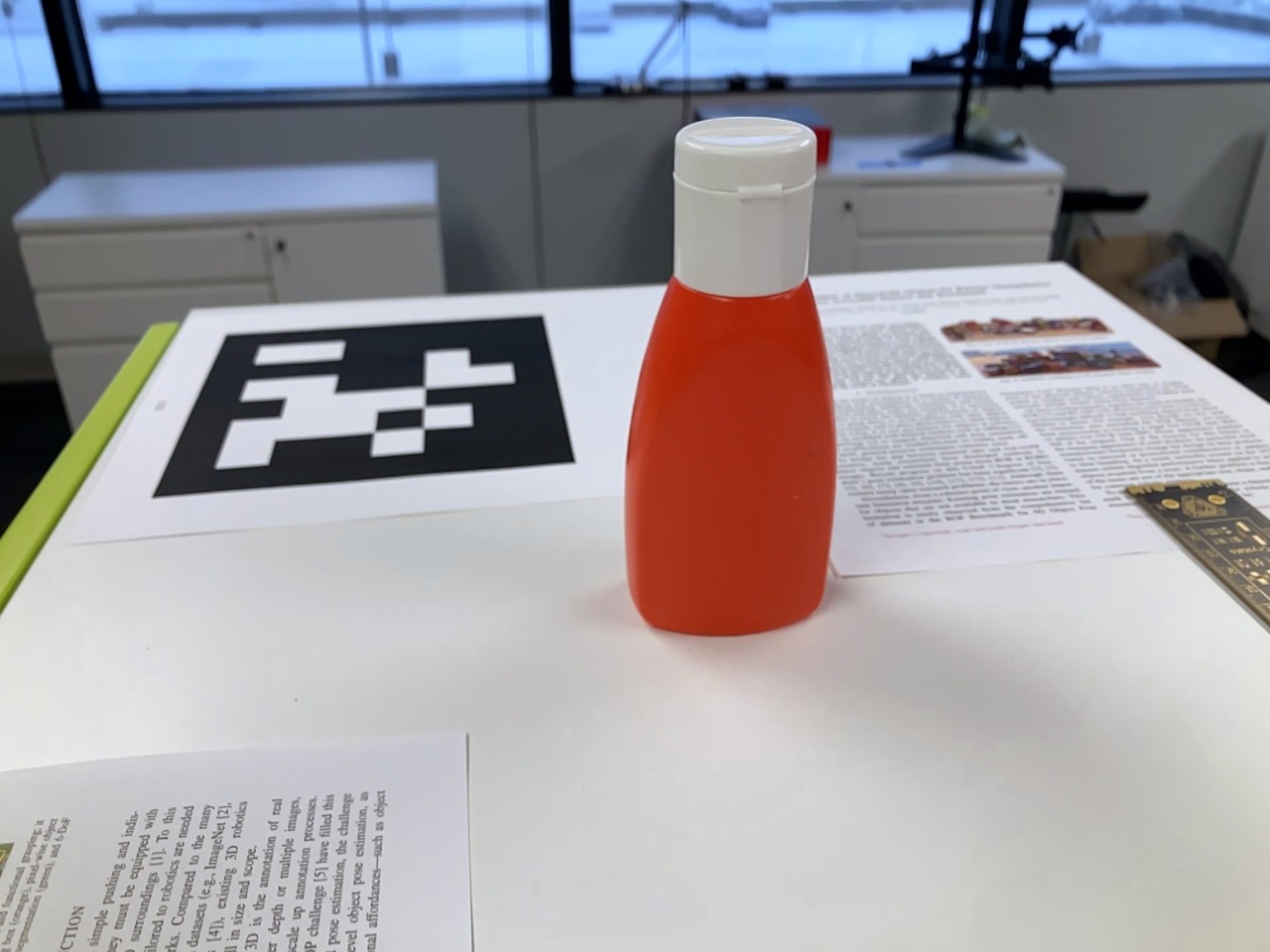}{0 0 0 0}}

        \put(3,-3){\framedimage{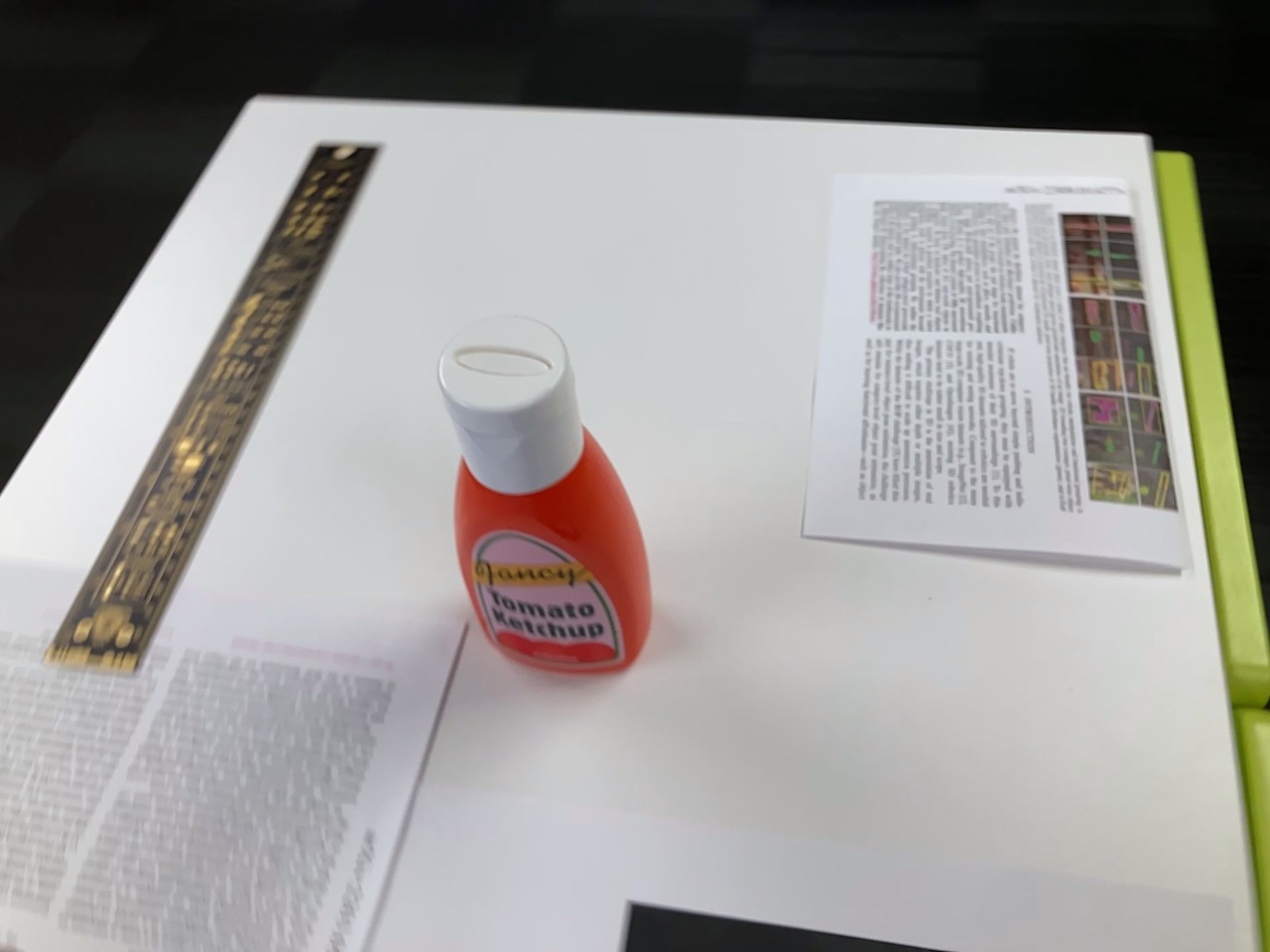}{0 0 0 0}}

        \put(6,-6){\framedimage{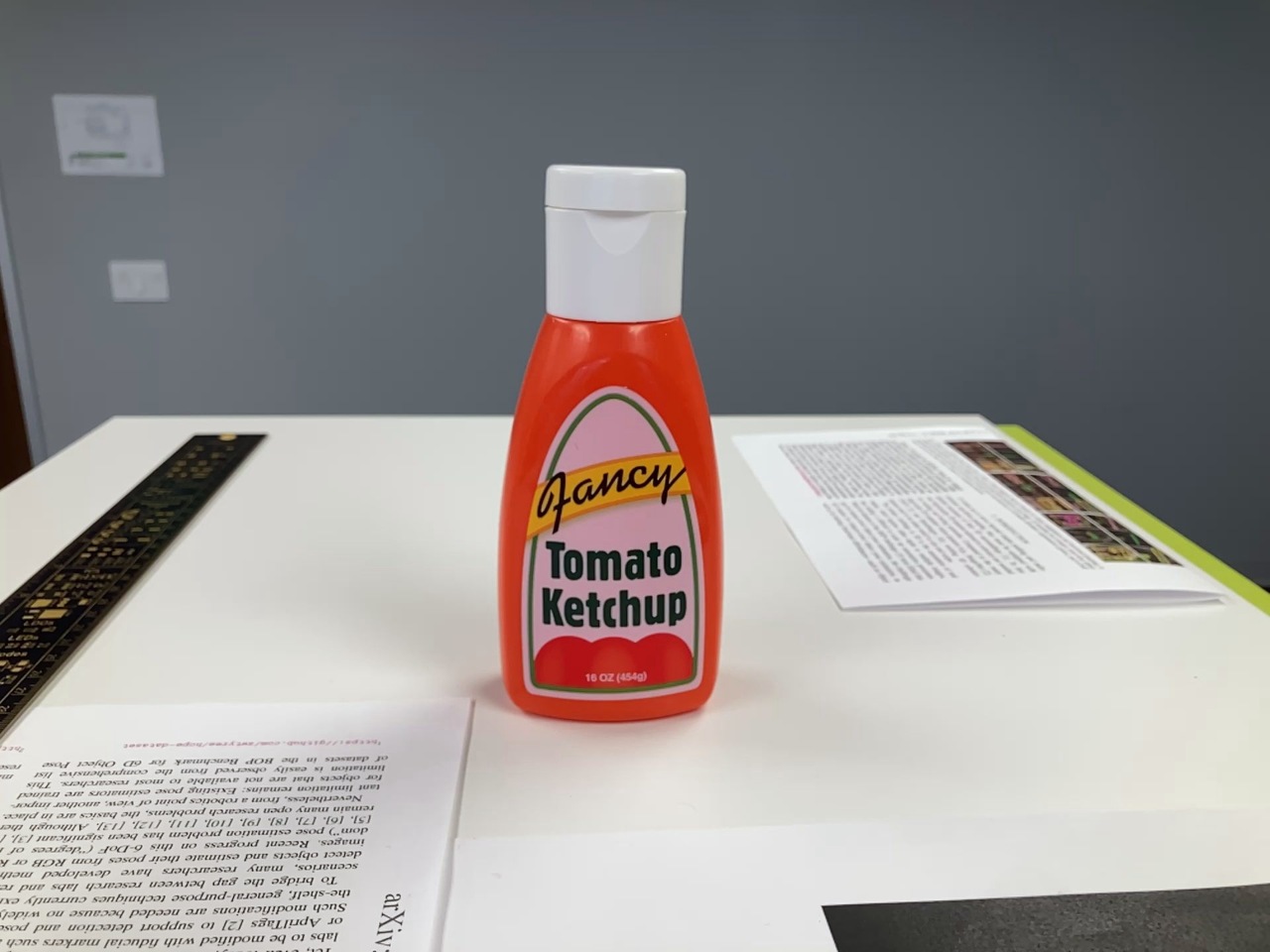}{0 0 0 0}} 
    \end{overpic}} &
\end{tabular}

\vspace{0.02cm}

\begin{tabular}{m{3.5cm} m{4.4cm} m{4.4cm} m{0.4cm} m{3.5cm} m{1.5cm}} %
    \includegraphics[height=\datasetsheight,width=\datasetsheight, trim=0 0 0 0, clip]{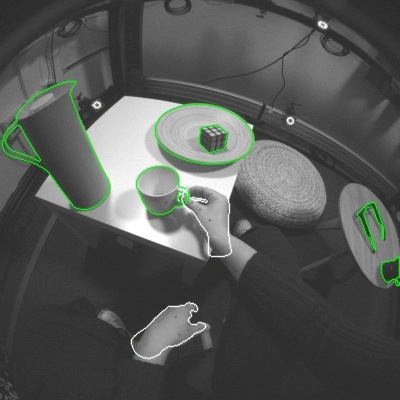} &
    \includegraphics[height=\datasetsheight,width=\datasetswidth, trim=120 0 120 0, clip]{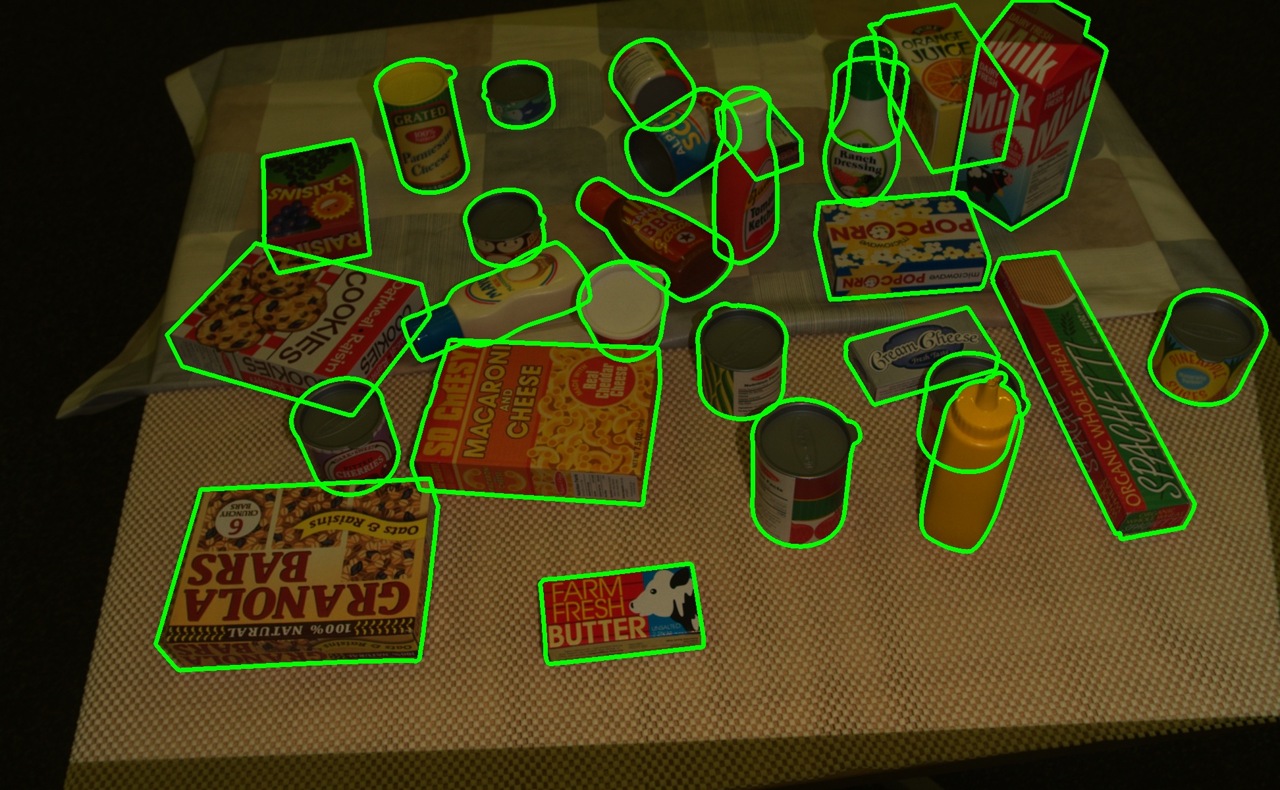} &
    \includegraphics[height=\datasetsheight,width=\datasetswidth]{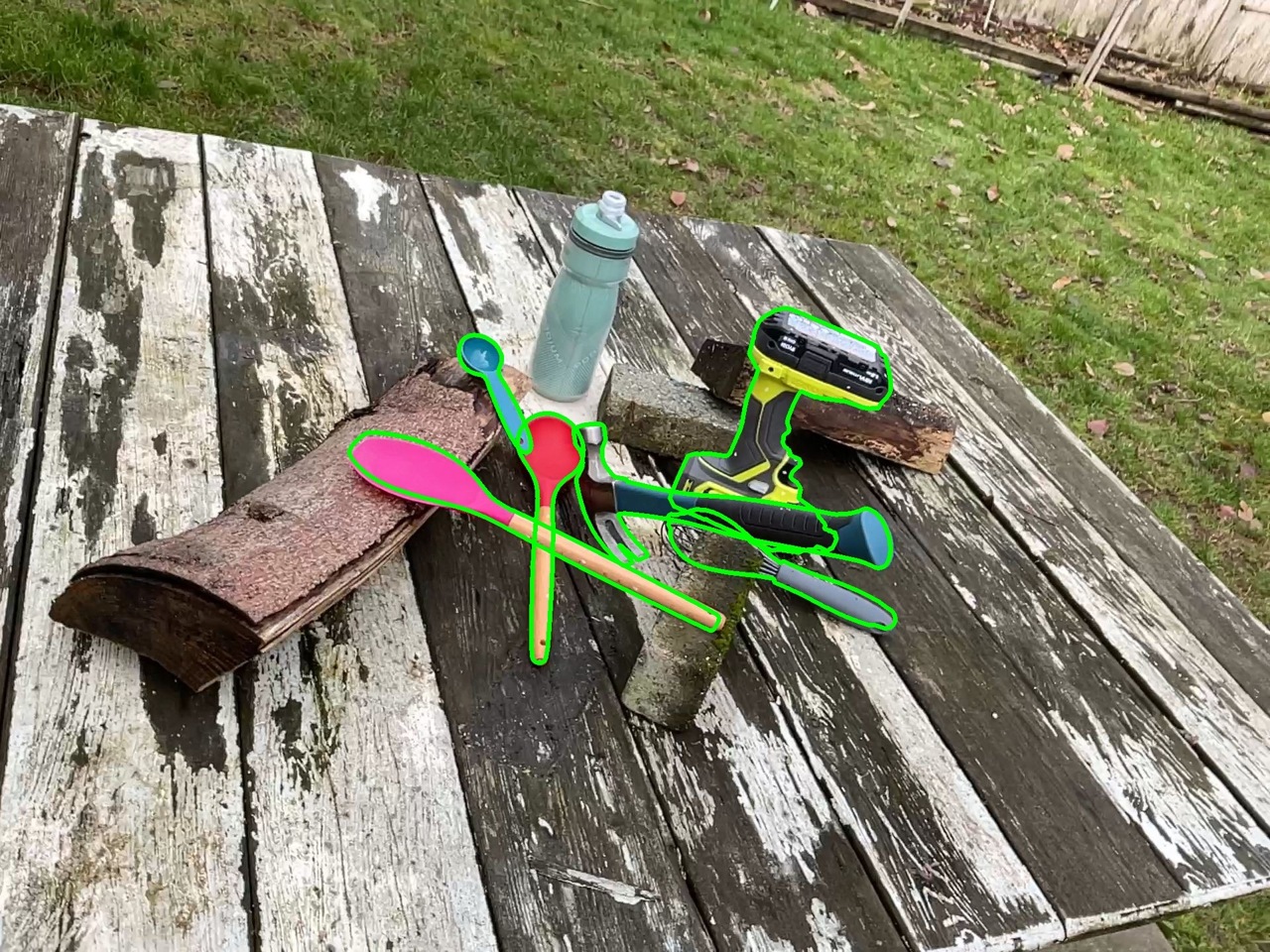} & &
    
    \raisebox{0.3cm}{\begin{overpic}[width=\onboardingwidth, height=\onboardingheight]{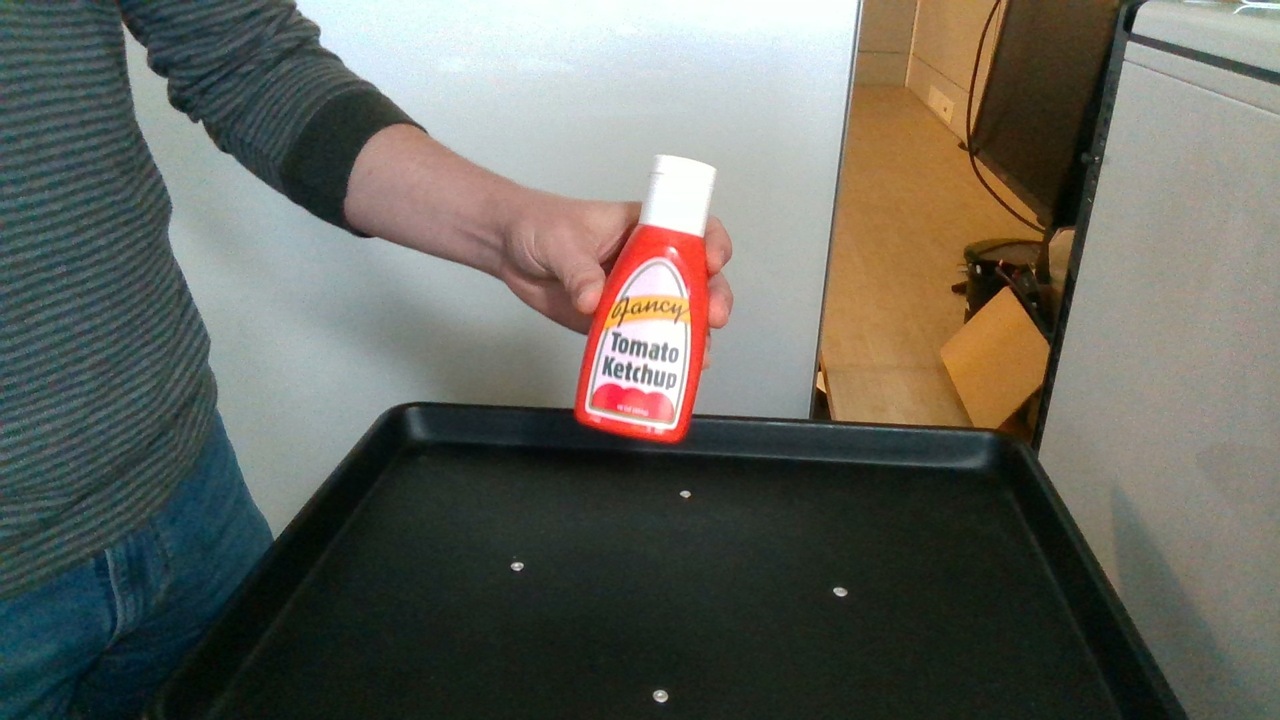}
        \put(0,0){\framedimage{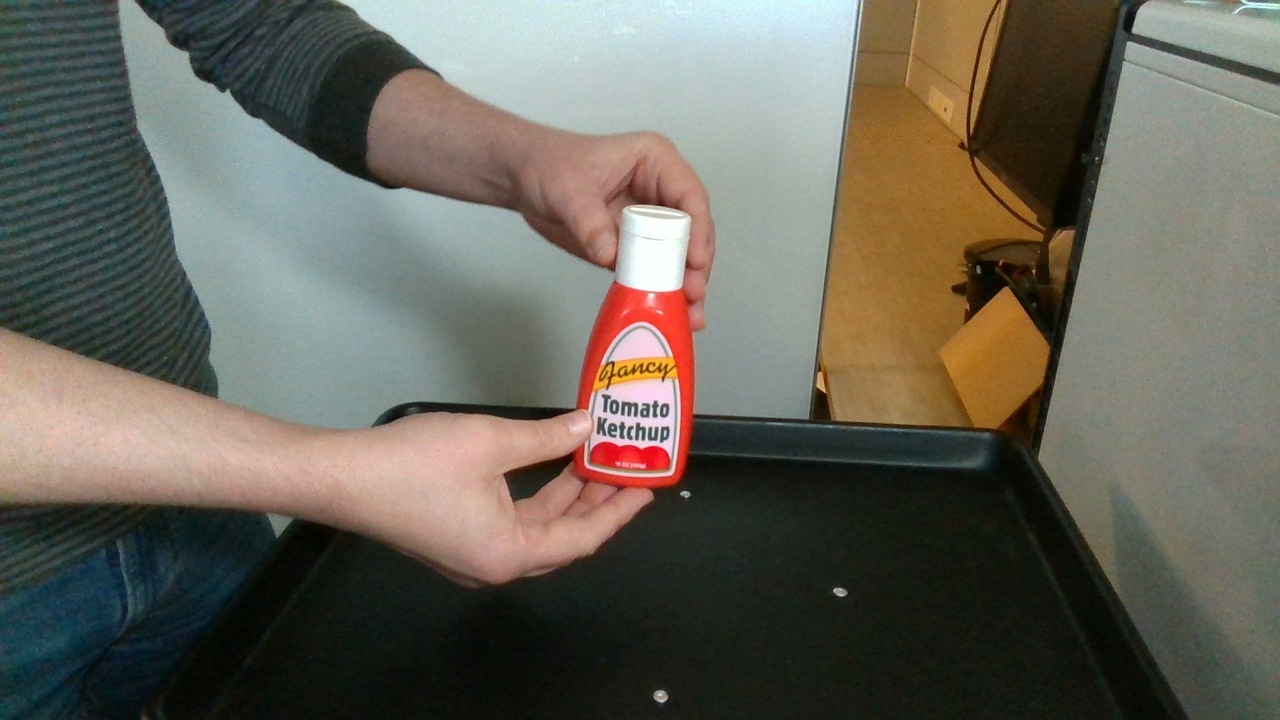}{200 0 200 0}}

        \put(3,-3){\framedimage{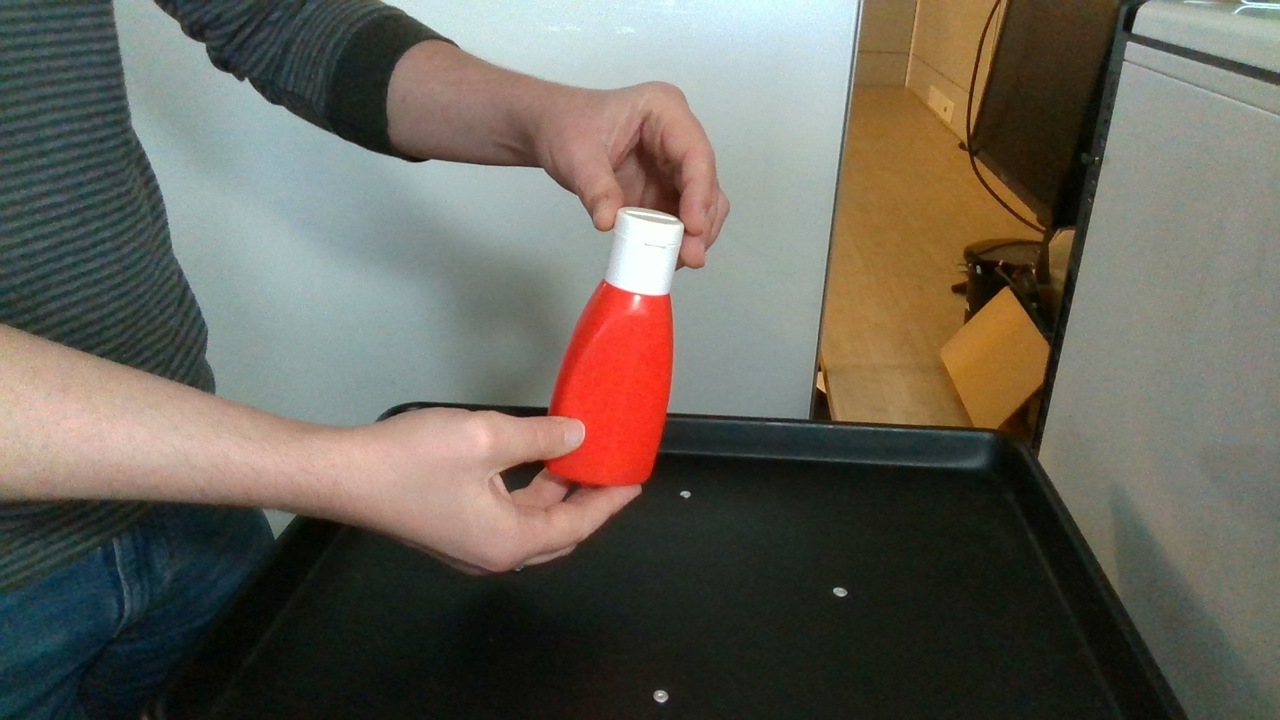}{200 0 200 0}}

        \put(6,-6){\framedimage{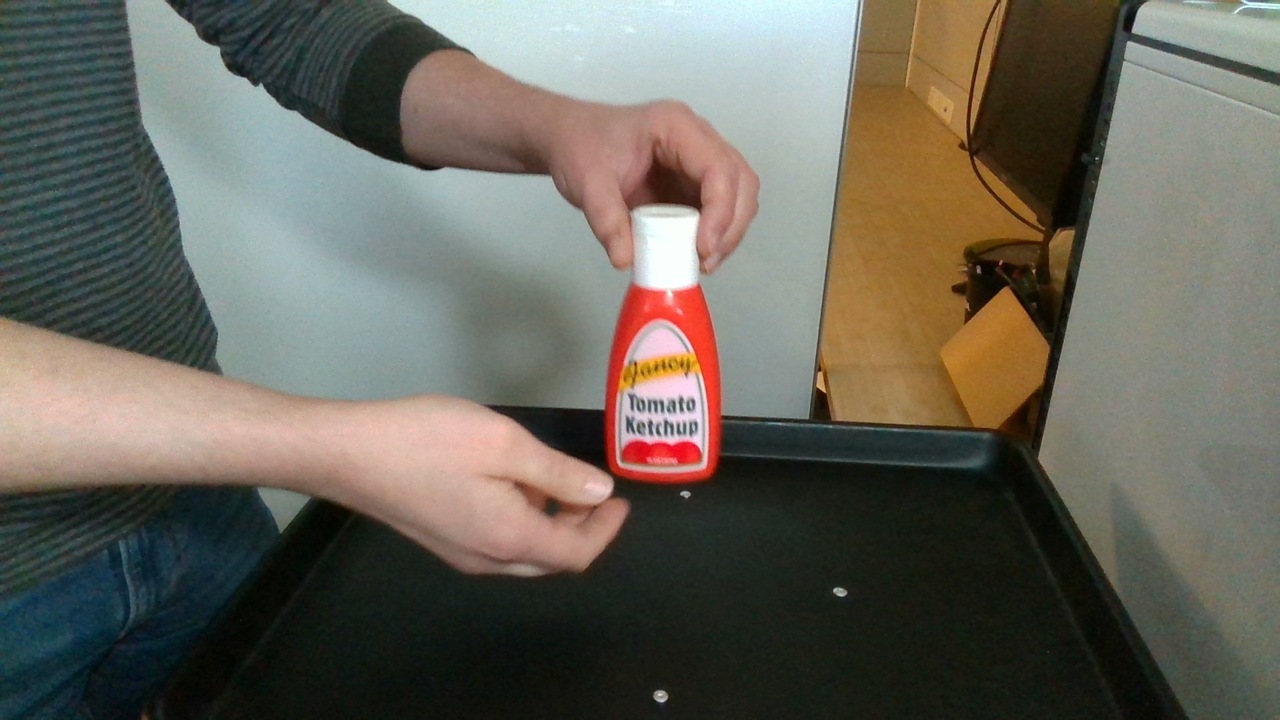}{200 0 200 0}}
    \end{overpic}} &
\end{tabular}
\vspace{-0.25cm}
\begin{center}
\begin{tabular}{m{3.5cm} m{4.4cm} m{4.4cm} m{0.4cm} m{3.5cm} m{1.5cm}} %
    \;\;\;\;\;\;\;\;\;\small{HOT3D~\cite{banerjee2024hot3d}} & \;\;\;\;\;\;\;\;\;\;\;\;\;\small{HOPEv2~\cite{tyree2022hope}} & \;\;\;\;\;\;\;\;\;\;\;\,\small{HANDAL~\cite{guo2023handal}} & & \;\;\;\small{Static/dynamic onboarding}
\end{tabular}
\end{center}
\vspace{-0.25cm}
\endgroup
\definecolor{darkgreen}{RGB}{34,139,34}
\captionsetup{type=figure}
\caption{\textbf{New BOP-H3 datasets with object onboarding sequences for model-free tasks.} The first three columns show sample images from the new datasets, with the contour of 3D object models in the ground-truth poses drawn in green. The fourth column shows a static (top) and dynamic (bottom) onboarding sequences, which are available in BOP-H3 and used for learning objects in the newly introduced model-free tasks.}
\label{fig:teaser}

\end{minipage}
\vspace{4pt}
\end{strip}

\begin{abstract}
\vspace{-7pt}
We present the evaluation methodology, datasets and results of the BOP Challenge 2024, the sixth in a series of public competitions organized to capture the state of the art in 6D object pose estimation and related tasks. In 2024, our goal was to transition BOP from lab-like setups to real-world scenarios.
First, we~introduced new model-free tasks, where no 3D object models are available and methods need to onboard objects just from provided reference videos. Second, we defined a new, more practical 6D object detection task where identities of objects visible in a test image are not provided as input (unlike in the classical 6D localization). Third, we introduced new BOP-H3 datasets recorded with high-resolution sensors and AR/VR headsets, closely resembling real-world scenarios. BOP-H3 include 3D models and onboarding videos to support both model-based and model-free tasks. Participants competed on seven challenge tracks, each defined by a task (6D localization, 6D detection, 2D detection), object onboarding setup (model-based, model-free), and dataset group (BOP-Classic-Core, BOP-H3). Notably, the best 2024 method for model-based 6D localization of unseen objects (FreeZeV2.1) achieves 22\% higher accuracy on BOP-Classic-Core than the best 2023 method (GenFlow), and is only 4\% behind the best 2023 method for seen objects (GPose2023) although being significantly slower (24.9 vs 2.7\,s per image). A more practical 2024 method for this task is Co-op which takes only 0.8\,s per image and is
13\% more accurate than GenFlow. Methods have similar rankings on 6D detection as on 6D localization but (as expected) higher run time. On model-based 2D detection of unseen objects, the best 2024 method (MUSE) achieves 21--29\% relative improvement compared to the best 2023 method (CNOS). However, the 2D detection accuracy for unseen objects is still -35\% behind the accuracy for seen objects (GDet2023), and the 2D detection stage is consequently the main bottleneck of existing pipelines for 6D localization/detection of unseen objects.
The online evaluation system stays open and is available at:~\texttt{\href{http://bop.felk.cvut.cz/}{bop.felk.cvut.cz}}.

\end{abstract}

\thispagestyle{plain}
\pagestyle{plain}

\vspace{-14pt}
\section{Introduction}

\customparagraph{2017--2023 summary.}
To measure the progress in 6D object pose estimation and related tasks, we created BOP (Benchmark Object Pose) in 2017 and have been organizing challenges on the benchmark datasets since then. Results of challenges from 2017, 2019, 2020, 2022, and 2023 are published in~\cite{hodan2018bop, hodan2019bop, hodan2020bop, sundermeyer2022bop, hodan2023bop}.
The field has come a long way, with the accuracy in model-based 6D localization of  \emph{seen objects} (target objects are seen during training) improving by more than 50\% (from 56.9 to 85.6 AR). In 2023, as the accuracy in this classical task had been saturating, we introduced a more practical yet more challenging task of model-based 6D localization of  \emph{unseen objects}, where new objects need to be onboarded just from their 3D models in under 5 minutes on a single GPU (the limit was chosen to be sufficient for operations like template rendering, but preventing exhaustive training and therefore ensuring relevance for practical applications). In addition to model-based 6D object localization, we have been evaluating model-based 2D object detection and 2D object segmentation.

\customparagraph{New model-free setup.}
While the model-based tasks are relevant for warehouse or factory settings where 3D models of target objects are typically available, their applicability is limited in open-world scenarios.
In 2024, we bridged this gap by introducing new model-free tasks, where 3D models are not available and methods need to learn new objects on the fly from onboarding
videos (Sec.~\ref{sec:onboarding_videos} and \ref{sec:onboarding}).
Methods that can operate in such a model-free setup will minimize the onboarding burden of new objects and unlock new types of applications, including augmented-reality systems capable of prompt object indexing and re-identification.

\customparagraph{New BOP-H3 datasets.}
To enable the model-free tasks and their comparison with the model-based variants, we introduced three new datasets referred jointly as BOP-H3: \textbf{H}OT3D~\cite{banerjee2024hot3d}, \textbf{H}OPEv2~\cite{tyree2022hope}, and \textbf{H}ANDAL~\cite{guo2023handal}. These datasets include texture-mapped 3D models and onboarding videos for 93 objects. To simulate different real-world scenarios, the datasets include two types of onboarding videos: \emph{static onboarding} where the object is static and the camera is moving around and capturing the object from different viewpoints, and \emph{dynamic onboarding} where the object is manipulated by hands and the camera is either static (on a tripod) or dynamic (on a head-mounted device). While methods were allowed to use all frames of the onboarding videos in 2024, we are planning to gradually limit the number of used frames to increase the practicality of the problem setup. See Fig.~\ref{fig:teaser} for sample images from BOP-H3 and Sec.~\ref{sec:bop-h3} for details.

\customparagraph{New 6D object detection task.}
In 2024, we also revisited the evaluation of object pose estimation. Since the beginning of BOP, we distinguish two object pose estimation tasks: \emph{6D object localization}, where identifiers of present object instances are provided for each test image, and \emph{6D object detection}, where no prior information is provided (see appendix A.1 in ~\cite{hodan2020bop} for a detailed comparison of these tasks). Up until 2024, we had been evaluating methods for object pose estimation only on 6D object localization because (1) pose accuracy on this simpler task had not been saturated, and (2) evaluating this task requires only calculating the recall rate which is noticeably less expensive than calculating the precision/recall curve required for evaluating 6D detection. While still supporting the 6D localization task, in 2024 we started evaluating also the 6D detection task.
This was possible thanks to new GPUs that we secured for the BOP evaluation server, run-time improvements of the evaluation scripts, and a simpler evaluation methodology -- only MSSD and MSPD pose-error functions are calculated when evaluating 6D detection, not VSD (Sec.~\ref{sec:evaluation_methodology}).
The VSD pose error function is more expensive to calculate and requires depth images which are not available in HOT3D and HANDAL. Besides speeding up the evaluation, omitting VSD therefore enables evaluating on RGB-only datasets.

\customparagraph{Summary of 2024 results.}
Participants
competed on seven challenge tracks, with each track defined by a task, object onboarding setup, and a group of datasets. Three of the tracks were on BOP-Classic-Core datasets and focused on model-based 6D localization, 6D detection, and 2D detection of unseen objects. The other four tracks were on BOP-H3 datasets and focused on model-based and model-free 6D detection and model-based and model-free 2D detection of unseen objects. In all tracks, methods had to onboard a new object within 5 minutes on a single GPU,
using provided 3D models in the model-based tasks and provided onboarding video sequences in the model-free tasks.

As detailed in Sec.~\ref{sec:evaluation}, the best 2024 method for model-based 6D localization of unseen objects (FreeZeV2.1~\cite{freeze}) achieves 22\% higher accuracy on BOP-Classic-Core than the best 2023 method (GenFlow~\cite{genflow}; 82.1 vs 67.4\,AR). FreeZeV2.1 is only 4\% behind the best 2023 method for seen objects (GPose2023~\cite{gpose2023}; 82.1 vs 85.6\,AR), although being significantly slower (24.9 vs 2.7\,s for estimating poses of all objects in an image on average). A more practical 2024 method for this task is Co-op~\cite{Moon2025Coop}, which takes only 0.8\,s per image and achieves a decent accuracy of 75.9\,AR (13\% higher than GenFlow).
Many methods for model-based 6D object pose estimation were evaluated on both the 6D object localization and the new 6D object detection task. Rankings on the two tasks are similar, with the main (expected) difference being a higher run time on the 6D object detection task. 

On model-based 2D detection of unseen objects, the best 2024 method (MUSE) achieves 21\% relative improvement compared to the best 2023 method for this task (CNOS~\cite{nguyen2023cnos}; 52.0 vs 42.8\,AP). However, the 2D detection accuracy for unseen objects is still noticeably behind the accuracy for seen objects (GDet2023~\cite{gpose2023} achieves 79.8\,AP). The 2D detection stage is consequently the primary bottleneck of recent pipelines for 6D detection/localization of unseen objects (all first detect target objects in 2D and then estimate the 6D pose per detection).

Participation in challenge tracks on the new BOP-H3 datasets and model-free tasks has been limited, supposedly due to the limited time and the non-negligible effort required to adopt the new datasets and develop new methods. However, the evaluation system stays open and we hope to see more submissions in the future.
\vspace{1ex}

\newcommand{\imgscale}{0.11} %

\begin{figure*}[t]
    \centering
    \renewcommand{\arraystretch}{0.5}
    
    \setlength\tabcolsep{0.5pt}

    \begin{tabular}{@{}ccccccccccccccccccccccccccccccccc@{}}
        \includegraphics[scale=\imgscale]{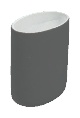} &
        \includegraphics[scale=\imgscale]{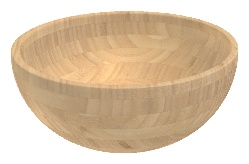} &
        \includegraphics[scale=\imgscale]{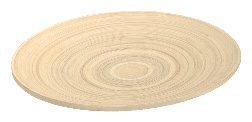} &
        \includegraphics[scale=\imgscale]{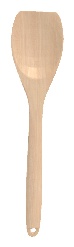} &
        \includegraphics[scale=\imgscale]{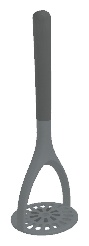} &
        \includegraphics[scale=\imgscale]{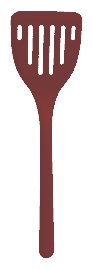} &
        \includegraphics[scale=\imgscale]{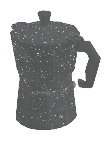} &
        \includegraphics[scale=\imgscale]{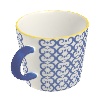} &
        \includegraphics[scale=\imgscale]{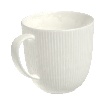} &
        \includegraphics[scale=\imgscale]{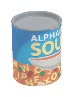} &
        \includegraphics[scale=\imgscale]{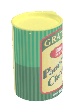} &
        \includegraphics[scale=\imgscale]{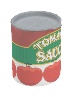} &
        \includegraphics[scale=\imgscale]{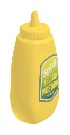} &
        \includegraphics[scale=\imgscale]{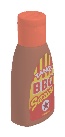} &
        \includegraphics[scale=\imgscale]{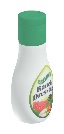} &
        \includegraphics[scale=\imgscale]{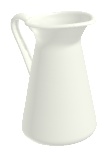} &
        \includegraphics[scale=\imgscale]{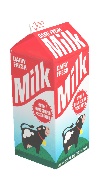} &
        \includegraphics[scale=\imgscale]{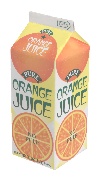} &
        \includegraphics[scale=\imgscale]{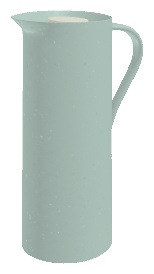} &
        \includegraphics[scale=\imgscale]{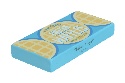} &
        \includegraphics[scale=\imgscale]{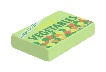} &
        \includegraphics[scale=\imgscale]{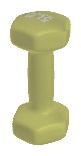} &
        \includegraphics[scale=\imgscale]{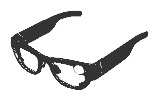} &
        \includegraphics[scale=\imgscale]{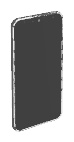} &
        \includegraphics[scale=\imgscale]{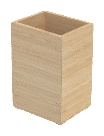} &
        \includegraphics[scale=\imgscale]{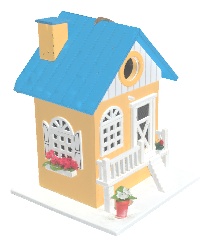} &
        \includegraphics[scale=\imgscale]{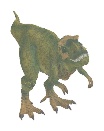} &
        \includegraphics[scale=\imgscale]{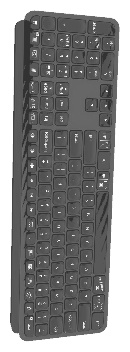} &
        \includegraphics[scale=\imgscale]{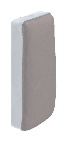} &
        \includegraphics[scale=\imgscale]{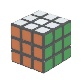} &
        \includegraphics[scale=\imgscale]{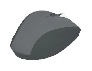} &
        \includegraphics[scale=\imgscale]{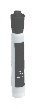} &
        \includegraphics[scale=\imgscale]{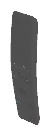} \\
        \scriptsize 1 & \scriptsize 2 & \scriptsize 3 & \scriptsize 4 & \scriptsize 5 & \scriptsize 6 & \scriptsize 7 & \scriptsize 8 & \scriptsize 9 & \scriptsize 10 & \scriptsize 11 & \scriptsize 12 & \scriptsize 13 & \scriptsize 14 & \scriptsize 15 & \scriptsize 16 & \scriptsize 17 & \scriptsize 18 & \scriptsize 19 & \scriptsize 20 & \scriptsize 21 & \scriptsize 22 & \scriptsize 23 & \scriptsize 24 & \scriptsize 25 & \scriptsize 26 & \scriptsize 27 & \scriptsize 28 & \scriptsize 29 & \scriptsize 30 & \scriptsize 31 & \scriptsize 32 & \scriptsize \,33 
    \end{tabular} \\
    \caption*{HOT3D objects (33 objects)}
    \vspace{10pt}

    \begin{tabular}{@{}cccccccccccccccccccccccccccc@{}}
        \includegraphics[scale=\imgscale]{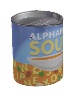} &
        \includegraphics[scale=\imgscale]{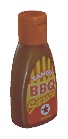} &
        \includegraphics[scale=\imgscale]{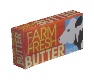} &
        \includegraphics[scale=\imgscale]{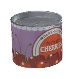} &
        \includegraphics[scale=\imgscale]{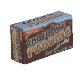} &
        \includegraphics[scale=\imgscale]{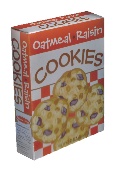} &
        \includegraphics[scale=\imgscale]{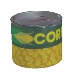} &
        \includegraphics[scale=\imgscale]{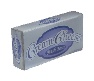} &
        \includegraphics[scale=\imgscale]{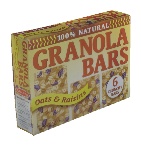} &
        \includegraphics[scale=\imgscale]{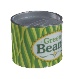} &
        \includegraphics[scale=\imgscale]{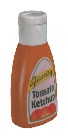} &
        \includegraphics[scale=\imgscale]{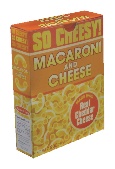} &
        \includegraphics[scale=\imgscale]{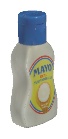} &
        \includegraphics[scale=\imgscale]{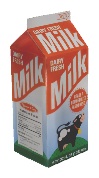} &
        \includegraphics[scale=\imgscale]{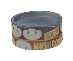} &
        \includegraphics[scale=\imgscale]{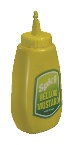} &
        \includegraphics[scale=\imgscale]{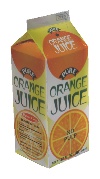} &
        \includegraphics[scale=\imgscale]{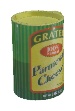} &
        \includegraphics[scale=\imgscale]{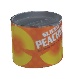} &
        \includegraphics[scale=\imgscale]{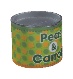} &
        \includegraphics[scale=\imgscale]{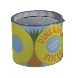} &
        \includegraphics[scale=\imgscale]{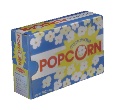} &
        \includegraphics[scale=\imgscale]{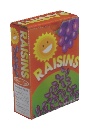} &
        \includegraphics[scale=\imgscale]{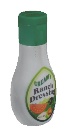} &
        \includegraphics[scale=\imgscale]{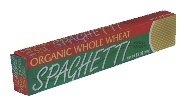} &
        \includegraphics[scale=\imgscale]{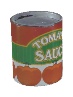} &
        \includegraphics[scale=\imgscale]{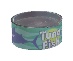} &
        \includegraphics[scale=\imgscale]{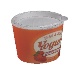} \\
        \scriptsize 1 & \scriptsize 2 & \scriptsize 3 & \scriptsize 4 & \scriptsize 5 & \scriptsize 6 & \scriptsize 7 & \scriptsize 8 & \scriptsize 9 & \scriptsize 10 & \scriptsize 11 & \scriptsize 12 & \scriptsize 13 & \scriptsize 14 & \scriptsize 15 & \scriptsize 16 & \scriptsize 17 & \scriptsize 18 & \scriptsize 19 & \scriptsize 20 & \scriptsize 21 & \scriptsize 22 & \scriptsize 23 & \scriptsize 24 & \scriptsize 25 & \scriptsize 26 & \scriptsize 27 & \scriptsize 28 
    \end{tabular} \\
    \caption*{HOPEv2 objects (28 objects)}

    \begin{tabular}{@{}ccccccccccccccccccccccccccccccccccccccccccccccc@{}}
        \includegraphics[scale=\imgscale]{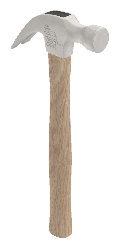} &
        \includegraphics[scale=\imgscale]{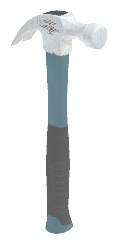} &
        \includegraphics[scale=\imgscale]{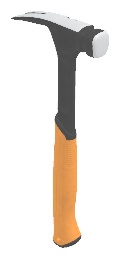} &
        \includegraphics[scale=\imgscale]{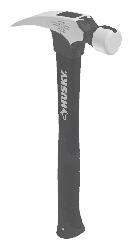} &
        \includegraphics[scale=\imgscale]{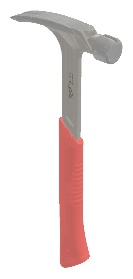} &
        \includegraphics[scale=\imgscale]{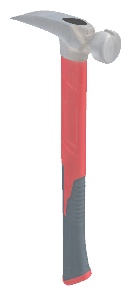} &
        \includegraphics[scale=\imgscale]{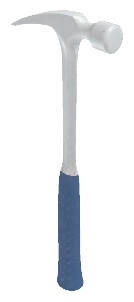} &
        \includegraphics[scale=\imgscale]{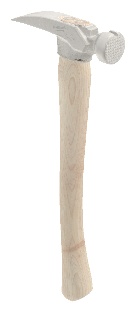} &
        \includegraphics[scale=\imgscale]{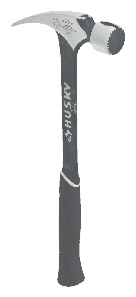} &
        \includegraphics[scale=\imgscale]{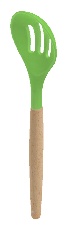} &
        \includegraphics[scale=\imgscale]{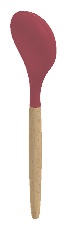} &
        \includegraphics[scale=\imgscale]{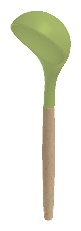} &
        \includegraphics[scale=\imgscale]{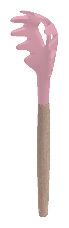} &
        \includegraphics[scale=\imgscale]{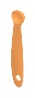} &
        \includegraphics[scale=\imgscale]{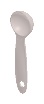} &
        \includegraphics[scale=\imgscale]{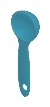} &
        \includegraphics[scale=\imgscale]{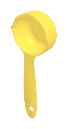} &
        \includegraphics[scale=\imgscale]{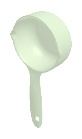} &
        \includegraphics[scale=\imgscale]{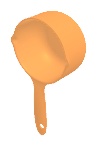} &
        \includegraphics[scale=\imgscale]{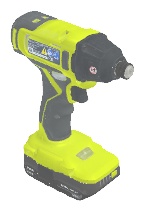} &
        \includegraphics[scale=\imgscale]{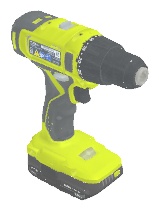} &
        \includegraphics[scale=\imgscale]{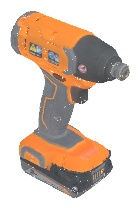} &
        \includegraphics[scale=\imgscale]{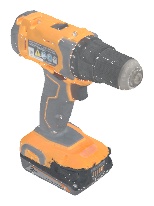} &
        \includegraphics[scale=\imgscale]{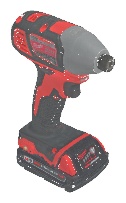} &
        \includegraphics[scale=\imgscale]{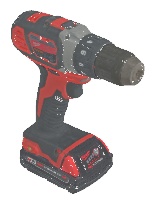} &
        \includegraphics[scale=\imgscale]{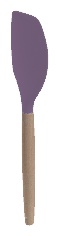} &
        \includegraphics[scale=\imgscale]{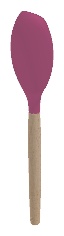} &
        \includegraphics[scale=\imgscale]{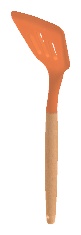} &
        \includegraphics[scale=\imgscale]{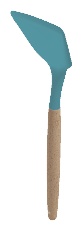} &
        \includegraphics[scale=\imgscale]{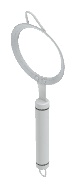} &
        \includegraphics[scale=\imgscale]{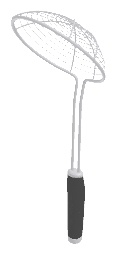} &
        \includegraphics[scale=\imgscale]{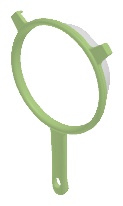} &
        \includegraphics[scale=\imgscale]{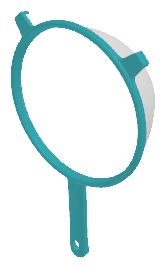} &
        \includegraphics[scale=\imgscale]{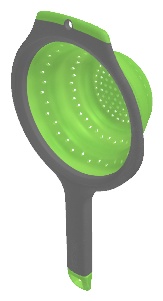} &
        \includegraphics[scale=\imgscale]{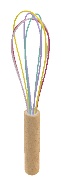} &
        \includegraphics[scale=\imgscale]{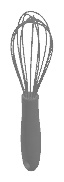} &
        \includegraphics[scale=\imgscale]{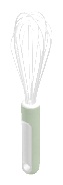} &
        \includegraphics[scale=\imgscale]{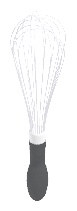} &
        \includegraphics[scale=\imgscale]{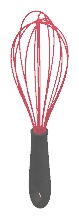} &
        \includegraphics[scale=\imgscale]{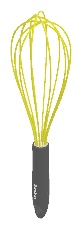} \\
        \scriptsize 1 & \scriptsize 2 & \scriptsize 3 & \scriptsize 4 & \scriptsize 5 & \scriptsize 6 & \scriptsize 7 & \scriptsize 8 & \scriptsize 9 & \scriptsize 10 & \scriptsize 11 & \scriptsize 12 & \scriptsize 13 & \scriptsize 14 & \scriptsize 15 & \scriptsize 16 & \scriptsize 17 & \scriptsize 18 & \scriptsize 19 & \scriptsize 20 & \scriptsize 21 & \scriptsize 22 & \scriptsize 23 & \scriptsize 24 & \scriptsize 25 & \scriptsize 26 & \scriptsize 27 & \scriptsize 28 & \scriptsize 29 & \scriptsize 30 & \scriptsize 31 & \scriptsize 32 & \scriptsize 33 & \scriptsize 34 & \scriptsize 35 & \scriptsize 36 & \scriptsize 37 & \scriptsize 38 & \scriptsize 39 & \scriptsize 40

    \end{tabular} \\
    \caption*{HANDAL objects (40 objects)}
    \vspace{7pt}

    \caption{\textbf{Objects from the BOP-H3 datasets introduced in 2024.} Objects are shown at the same scale. HOT3D and HOPEv2 share eight objects.
    \vspace{0.7ex}
    }
    \label{fig:bop_h3}
\end{figure*}

\section{Datasets} \label{sec:datasets}

In 2024, methods were evaluated on the BOP-Classic datasets (Sec.~\ref{sec:bop-classic}), which were used in all previous challenges, and the newly introduced BOP-H3 datasets (Sec.~\ref{sec:bop-h3}). 
Each dataset includes 3D object models and images of the objects annotated with amodal 2D bounding boxes, modal 2D segmentation masks, and 6D object poses.
The object models are provided in the form of 3D meshes (in most cases with a color texture) which were created manually or 3D reconstructed~\cite{newcombe2011kinectfusion}.
Depending on the dataset, the images are RGB-D, RGB, or monochrome.
All datasets include real test images showing the objects in scenes with various complexity, often with clutter and occlusion. Most datasets include training images which may be real and/or synthetic,
with all BOP-Classic-Core and BOP-H3 datasets offering
50K photorealistic physically-based rendered (PBR) training images generated and automatically annotated with BlenderProc~\cite{denninger2019blenderproc,denninger2020blenderproc,denninger2023blenderproc2}. Some datasets also include real validation images.
Ground-truth annotations are publicly available only for training and validation images, and also for test images from BOP-Classic datasets that do not have validation images. Private ground-truth annotations are only accessible by the BOP~evaluation server.
Tab.~\ref{tab:dataset_params} shows the dataset statistics.

\subsection{BOP-Classic datasets} \label{sec:bop-classic}
 
BOP-Classic is a group of twelve traditional 6D object pose estimation datasets.
As in previous years, authors were required to evaluate their methods on at least seven of the datasets, called BOP-Classic-Core, to be considered for the challenge awards. The seven datasets include test images showing 272K instances (19K used for the evaluation) of 132 objects (details in Sec.~7.2 of \cite{hodan2021phd}).

\begin{figure}[h!]
\begin{center}

\scriptsize

\renewcommand{\arraystretch}{0.9}

\begingroup
\setlength\tabcolsep{1.0pt}
\footnotesize
\begin{tabularx}{\columnwidth}{ l r R R R R R R R }
\toprule
&
&
\multicolumn{2}{c}{Train. im.} &
\multicolumn{1}{c}{Val im.} &
\multicolumn{2}{c}{Test im.} &
\multicolumn{2}{c}{Test inst.} \\
\cmidrule(l{2pt}r{2pt}){3-4} \cmidrule(l{2pt}r{2pt}){5-5} \cmidrule(l{2pt}r{2pt}){6-7} \cmidrule(l{2pt}r{2pt}){8-9}
\multicolumn{1}{l}{Dataset} &
\multicolumn{1}{c}{Obj.} &
\multicolumn{1}{c}{Real} &
\multicolumn{1}{c}{PBR} &
\multicolumn{1}{c}{Real} &
\multicolumn{1}{c}{All} &
\multicolumn{1}{c}{Used} &
\multicolumn{1}{c}{All} &
\multicolumn{1}{c}{Used} \\
\midrule
\multicolumn{8}{l}{\textbf{BOP-H3:}}\\
HOT3D \cite{banerjee2024hot3d} & 33 & 420600 & 50K & -- & 154200 & 5140 & 709715 & 23642 \\
HOPEv2 \cite{tyree2022hope} &  28 & -- & 50K & 50 & 457 & 457  & 9276 & 9276  \\
HANDAL \cite{guo2023handal} &  40 & -- & 50K & 2208 &  13261 & 1684 & 74771 & 9492 \\
\midrule
\multicolumn{8}{l}{\textbf{BOP-Classic-Core:}}\\
LM-O \cite{brachmann2014learning} & 8 & -- & 50K & -- & 1214 & 200 & 9038 & 1445 \\
T-LESS \cite{hodan2017tless} &  30 & 37584 & 50K & -- & 10080 & 1000 & 67308 & 6423 \\
ITODD \cite{drost2017introducing} &  28 & -- & 50K & 54 & 721 & 721 & 3041 & 3041 \\
HB \cite{kaskman2019homebreweddb} &  33 & -- & 50K & 4420 & 13000 & 300 & 67542 & 1630 \\
YCB-V \cite{xiang2017posecnn} &  21 & 113198 & 50K & -- & 20738 & 900 & 98547 & 4123 \\
TUD-L \cite{hodan2018bop} &  3 & 38288 & 50K & -- & 23914 & 600 & 23914 & 600 \\
IC-BIN \cite{doumanoglou2016recovering} & 2 & -- & 50K & -- & 177 & 150 & 2176 & 1786 \\
\midrule
\multicolumn{5}{l}{\textbf{BOP-Classic-Extra:}}\\
LM \cite{hinterstoisser2012accv} & 15 & -- & 50K & -- & 18273 & 3000 & 18273 & 3000 \\
RU-APC \cite{rennie2016dataset} & 14 & -- & -- & -- & 5964 & 1380 & 5964 & 1380\\
IC-MI \cite{tejani2014latent} & 6 & -- & -- & -- & 2067 & 300 & 5318 & 800 \\
TYO-L \cite{hodan2018bop} & 21 & -- & -- & -- & 1670 & 1670 & 1670 & 1670 \\
HOPEv1 \cite{tyree2022hope} & 28 & -- & 50K & 50 & 188 & 188 & 3472 & 2898 \\
\bottomrule
\end{tabularx}

\endgroup

\captionof{table}{\label{tab:dataset_params} \textbf{Parameters of BOP datasets.}
Column \emph{Test inst./All} shows the number of annotated object instances for which at least $10\%$ of the projected surface area is visible in test images. Columns \emph{Used} show the number of used test images and object instances. All datasets include 3D object models. Only BOP-H3 include videos for object onboarding.
\vspace{-0.7em}
}
\end{center}
\end{figure}

\subsection{BOP-H3 datasets} \label{sec:bop-h3}

BOP-H3 is a group of three new datasets (HOT3D, HOPEv2, HANDAL) that enable the evaluation of both model-based and model-free methods
by providing 3D models (Fig.~\ref{fig:bop_h3}) and onboarding videos for all objects.
These three datasets include test images showing 794K instances (42K used for the evaluation, which is 2 times more than in BOP-Classic-Core) of 93 unique objects.

\customparagraph{HOT3D~\cite{banerjee2024hot3d}}
is a dataset for egocentric hand and object tracking in 3D with multi-view RGB and monochrome image streams showing participants interacting with 33 diverse rigid objects. The dataset is recorded with two recent head-mounted devices from Meta: Aria, which is a research prototype of light-weight AI glasses, and Quest 3, a production VR headset that has been sold in millions of units. HOT3D also includes PBR materials for the 3D object models, real training images, 3D hand pose and shape annotations, and eye-tracking signal in recordings from Aria. In BOP, we use HOT3D-Clips, which is a curated subset of HOT3D. Each clip has 150 frames (5 seconds) that are all annotated with ground-truth poses of all modeled objects and hands.
There are 4117 clips in total (2969 training, 1148 test). HOT3D-Clips are also used in the Multiview Egocentric Hand Tracking Challenge~\cite{han2022umetrack}.

\customparagraph{HOPEv2~\cite{tyree2022hope}}
is a dataset for robotic manipulation featuring 28 toy grocery objects available from online retailers for about 50 USD. The original HOPEv1 dataset includes images of 50 static object arrangements in 10 household/office environments, with each arrangement captured under up to 5 lighting variations from multiple viewpoints.
For the 2024 challenge, we released an updated version with additional test images showing 7 static object arrangements from multiple viewpoints. HOPEv2 is the only BOP-H3 dataset that provides images with the depth channel.

\customparagraph{HANDAL~\cite{guo2023handal}}
is a dataset with graspable or manipulable objects (hammers, ladles, measuring cups, power drills, spatulas, strainers, whisks). The objects are arranged in indoor and outdoor scenes and captured from multiple viewpoints. The original dataset has 212 objects from 17 categories.
For the 2024 challenge, we captured additional test images and only consider 40 objects from 7 categories, with high-quality 3D models created by 3D artists.

\subsection{Object onboarding videos}
\label{sec:onboarding_videos}
\begin{figure}[t!]
\begin{center}

\scriptsize

\renewcommand{\arraystretch}{0.9}

\begin{tabular}{ @{}c@{ } @{}c@{ } @{}c@{ } @{}c@{ } }

\includegraphics[width=0.243\columnwidth]{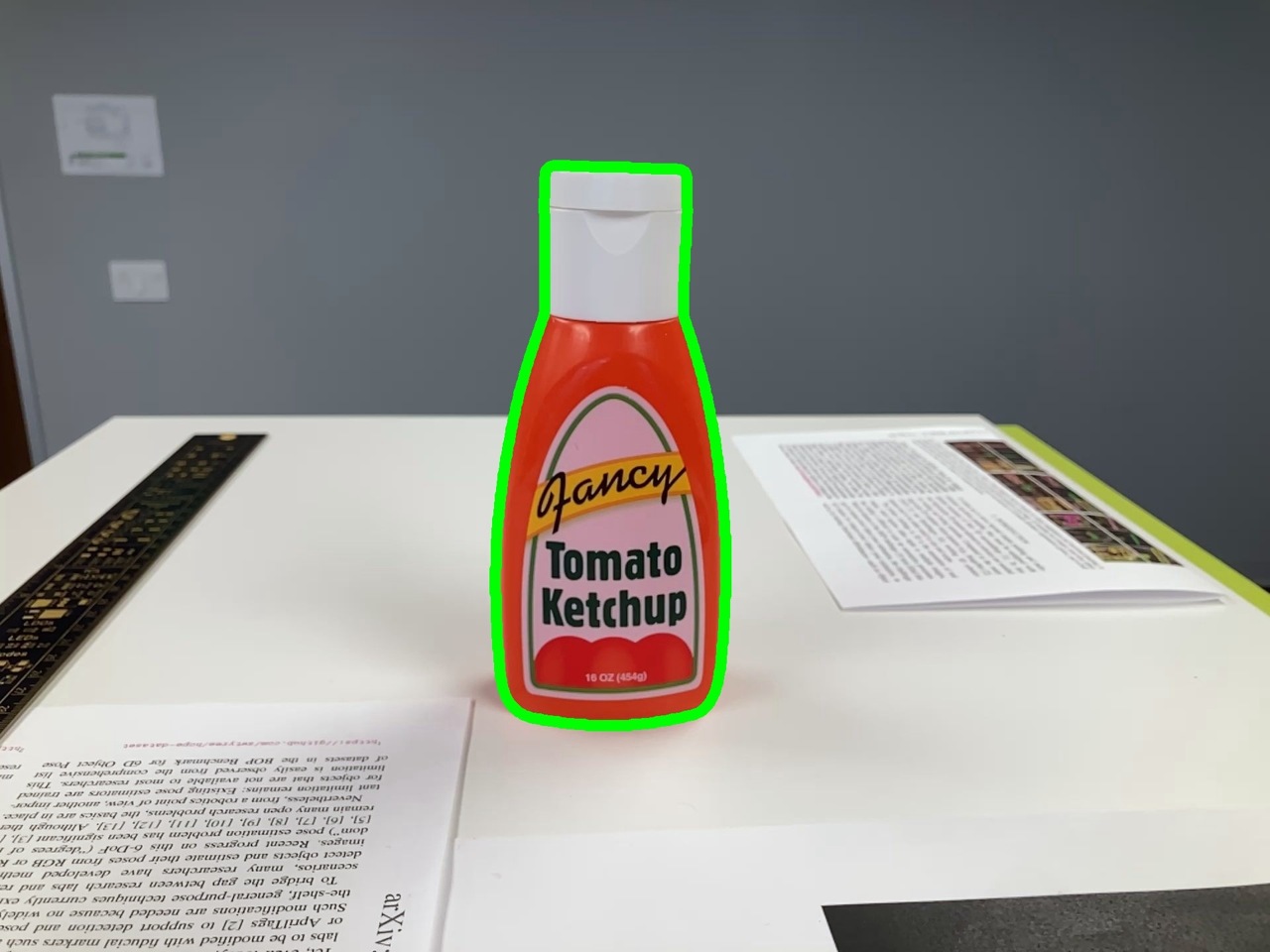} &
\includegraphics[width=0.243\columnwidth]{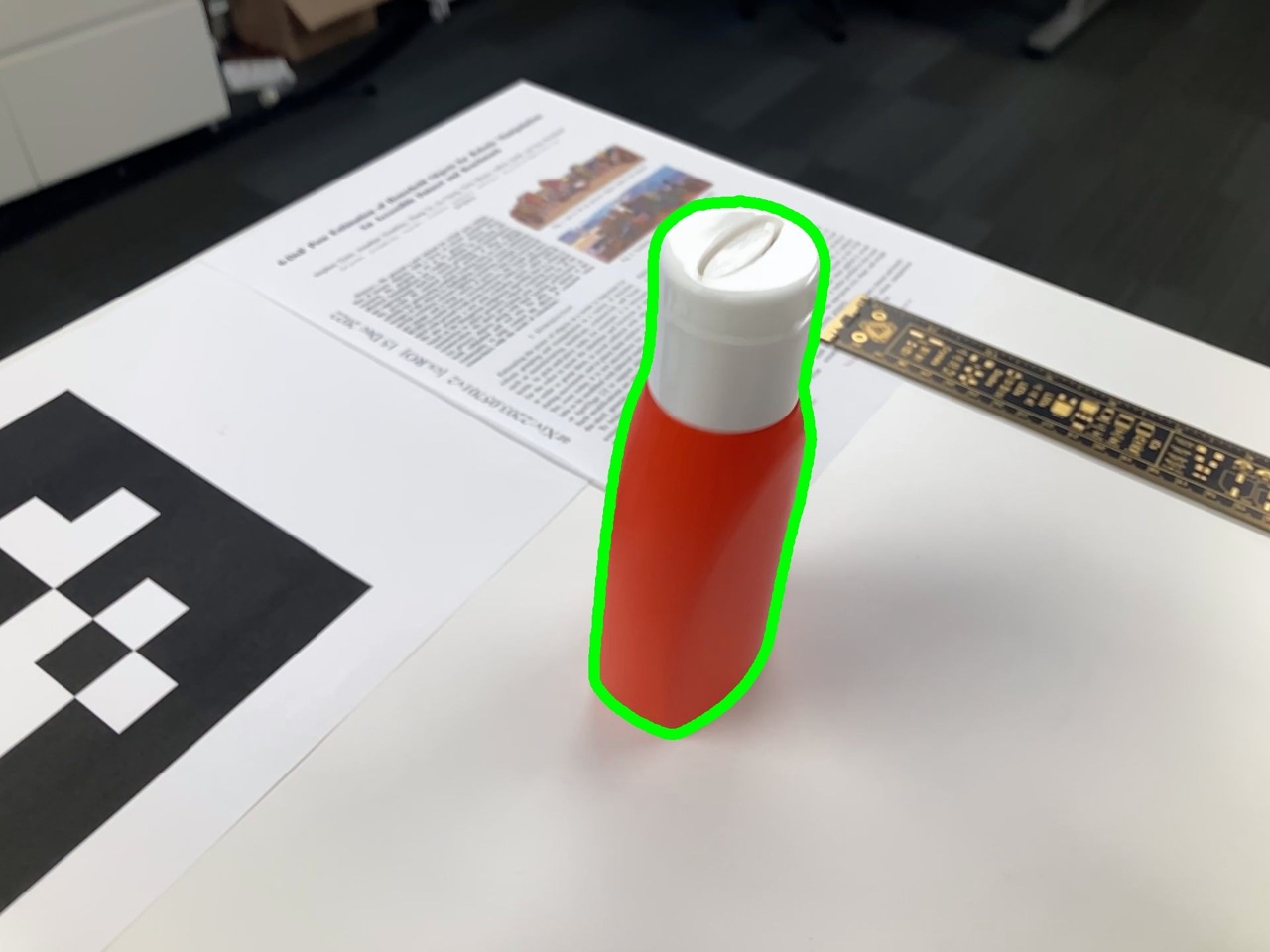} &
\includegraphics[width=0.243\columnwidth]{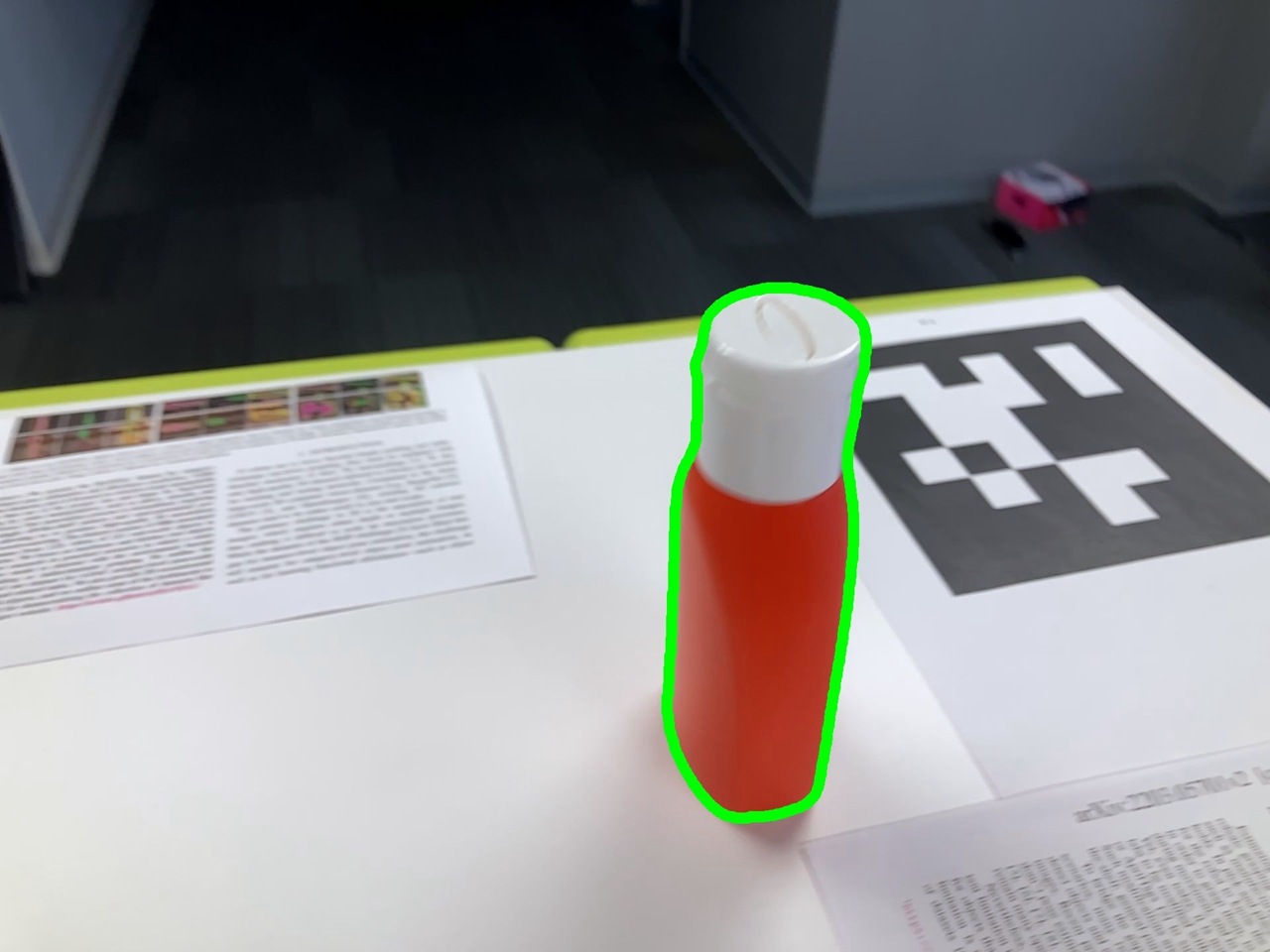} &
\includegraphics[width=0.243\columnwidth]{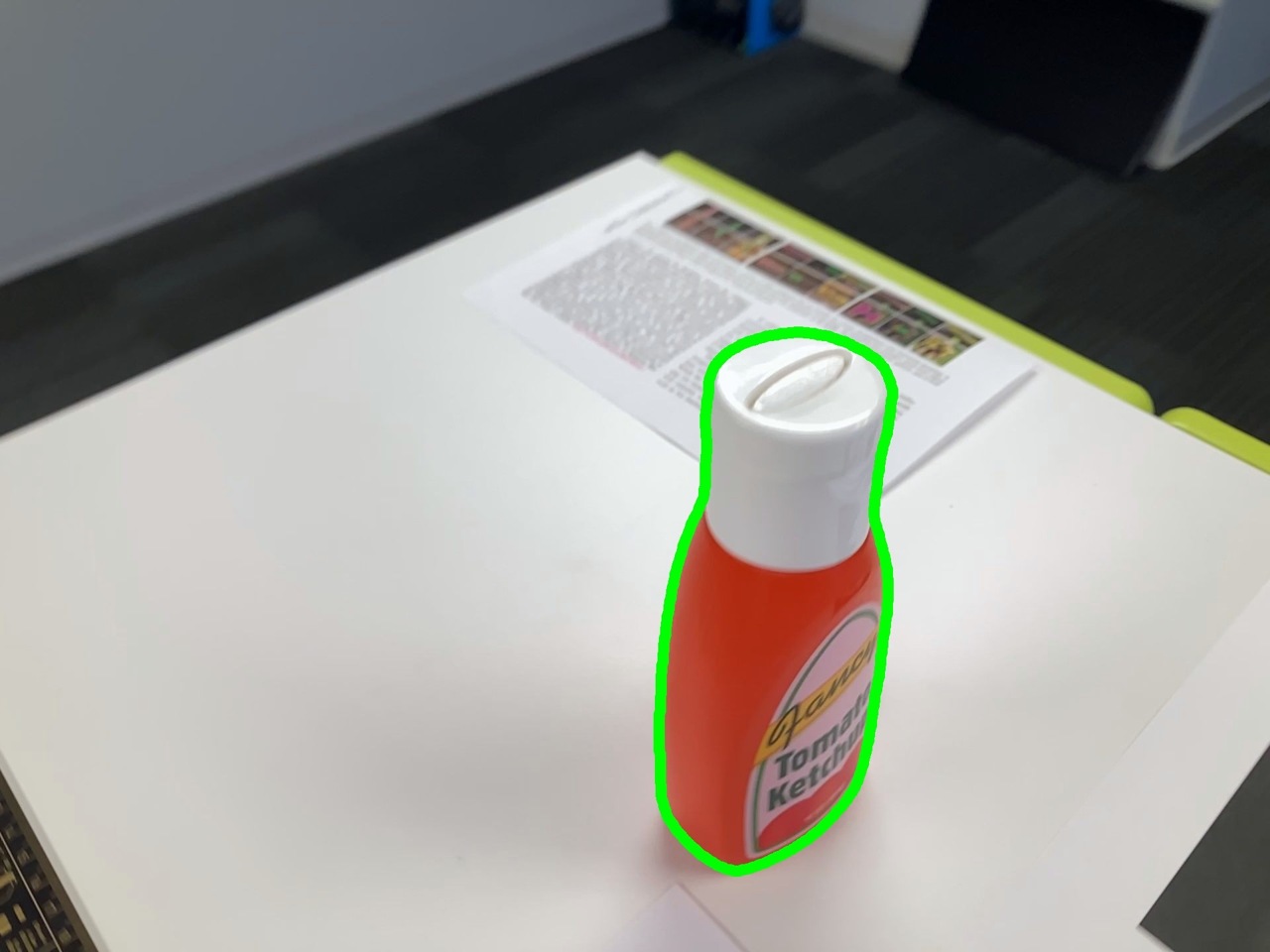} \\

\includegraphics[width=0.243\columnwidth]{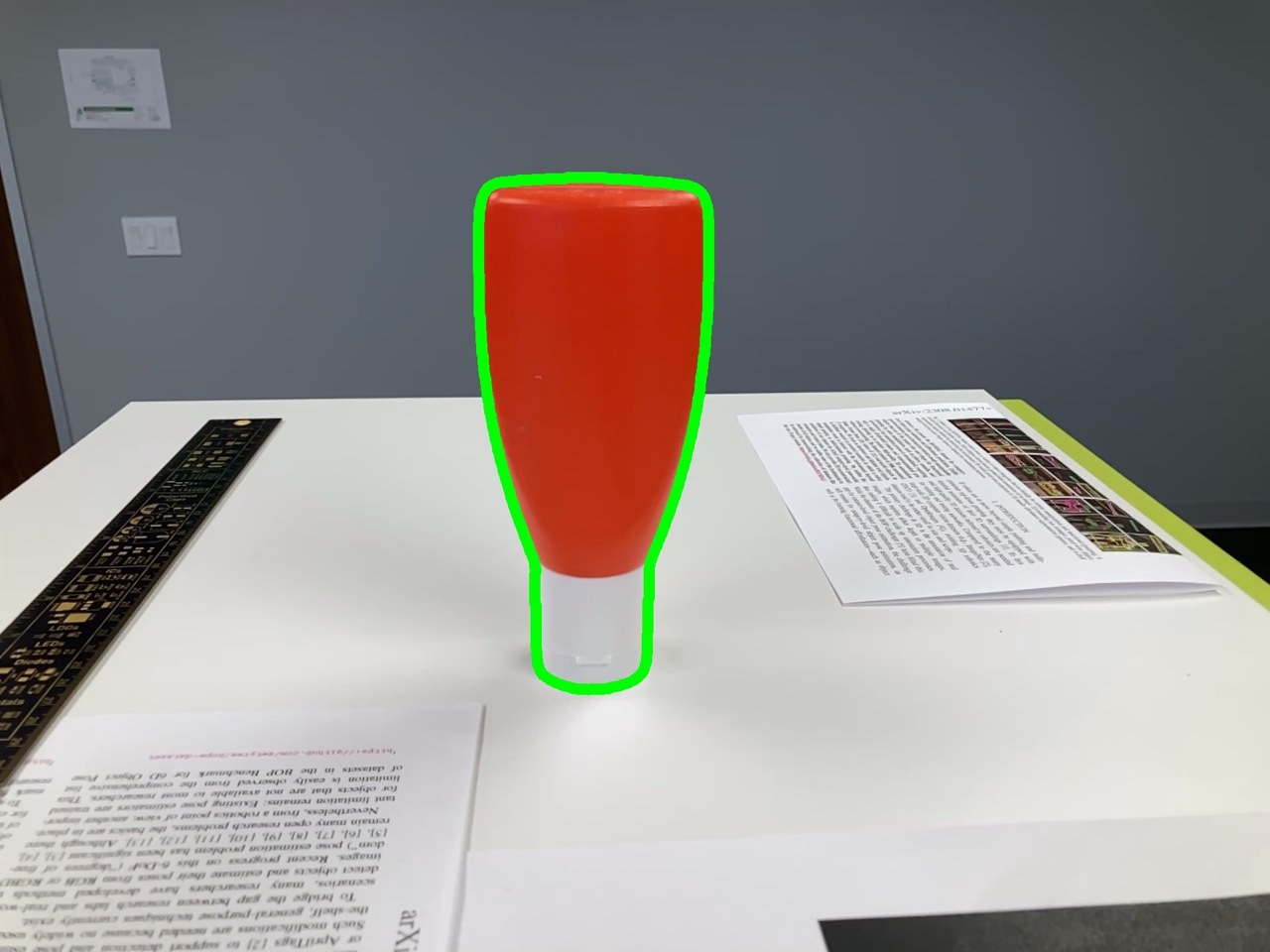} &
\includegraphics[width=0.243\columnwidth]{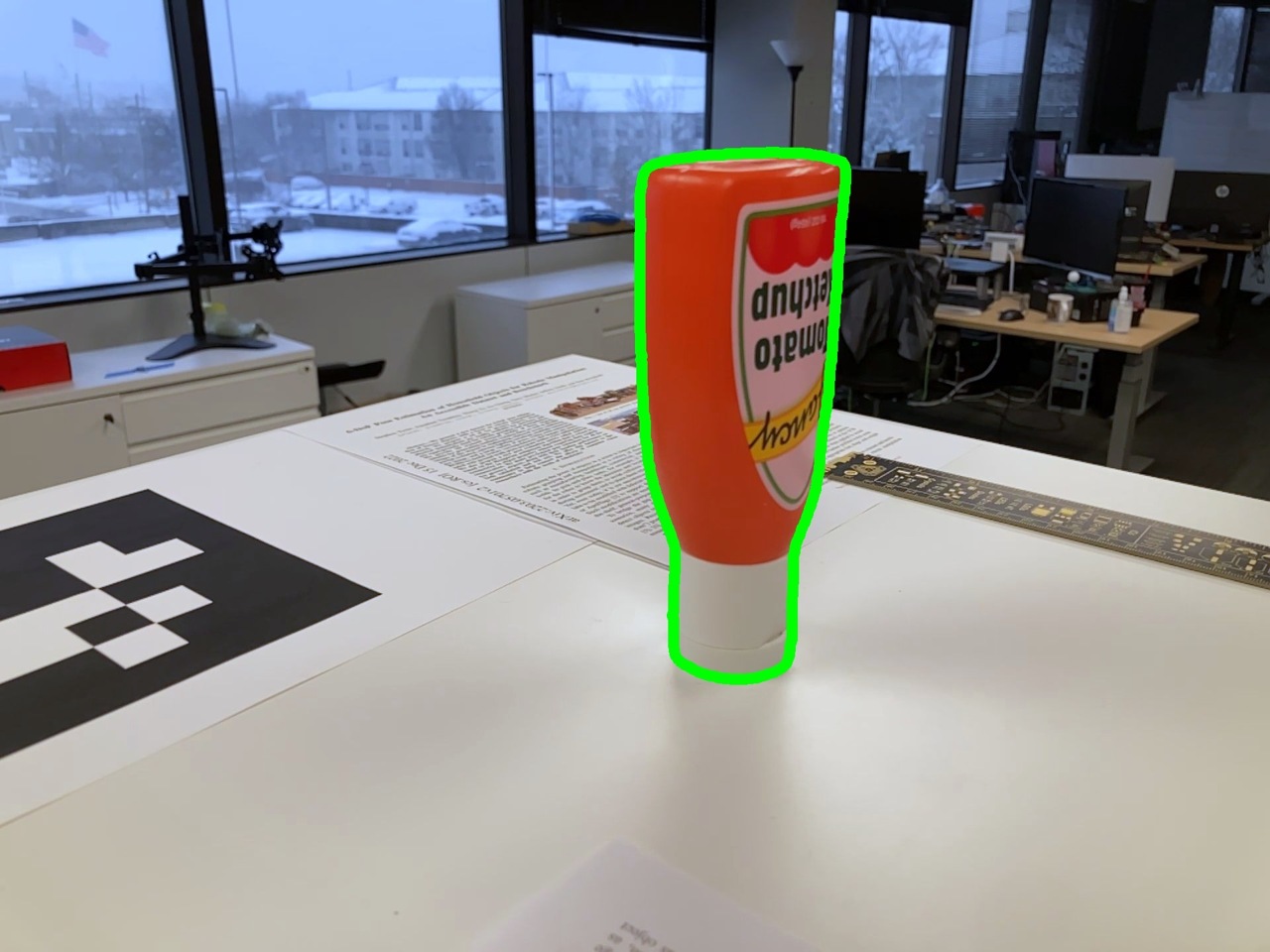} &
\includegraphics[width=0.243\columnwidth]{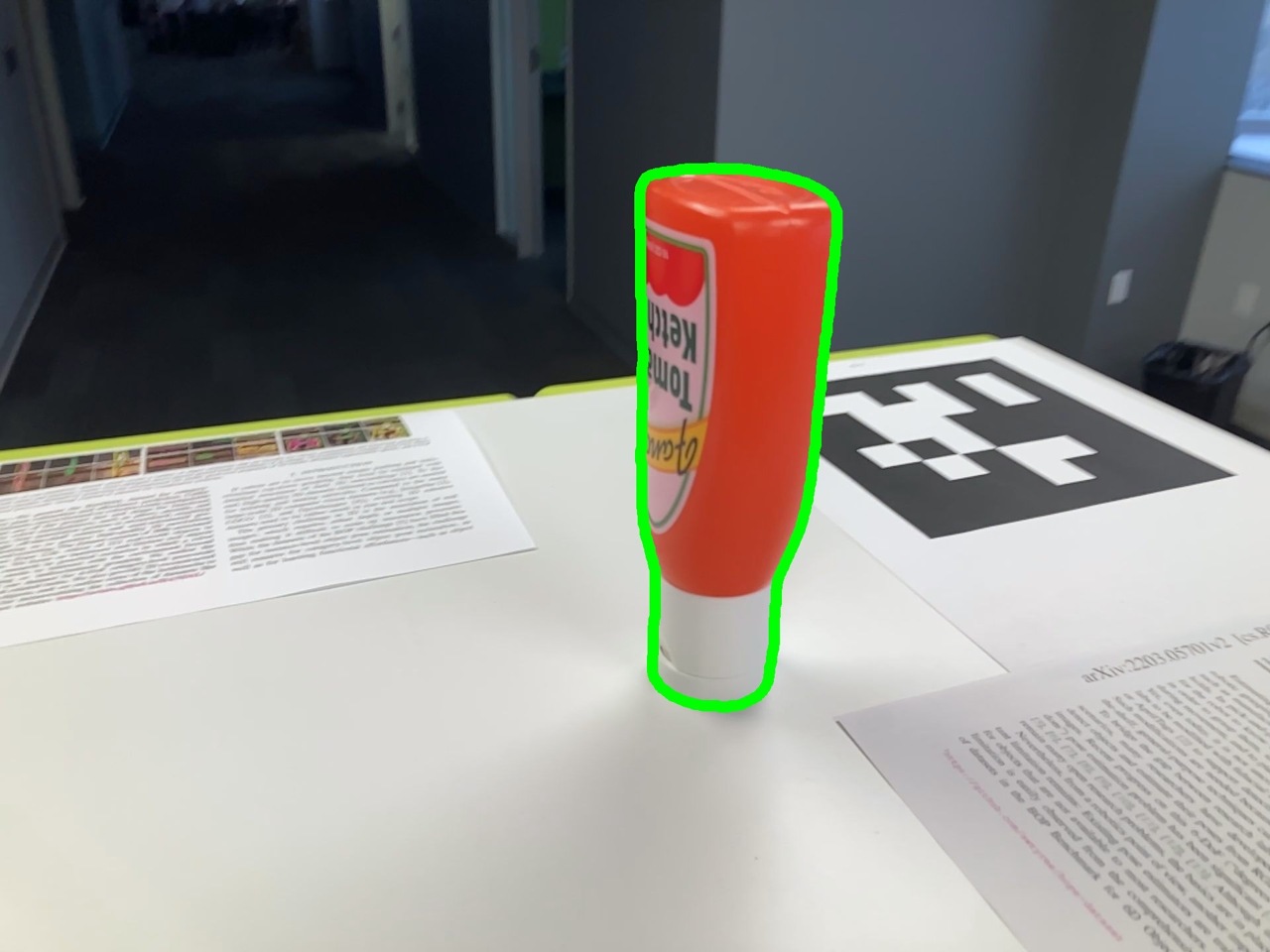} &
\includegraphics[width=0.243\columnwidth]{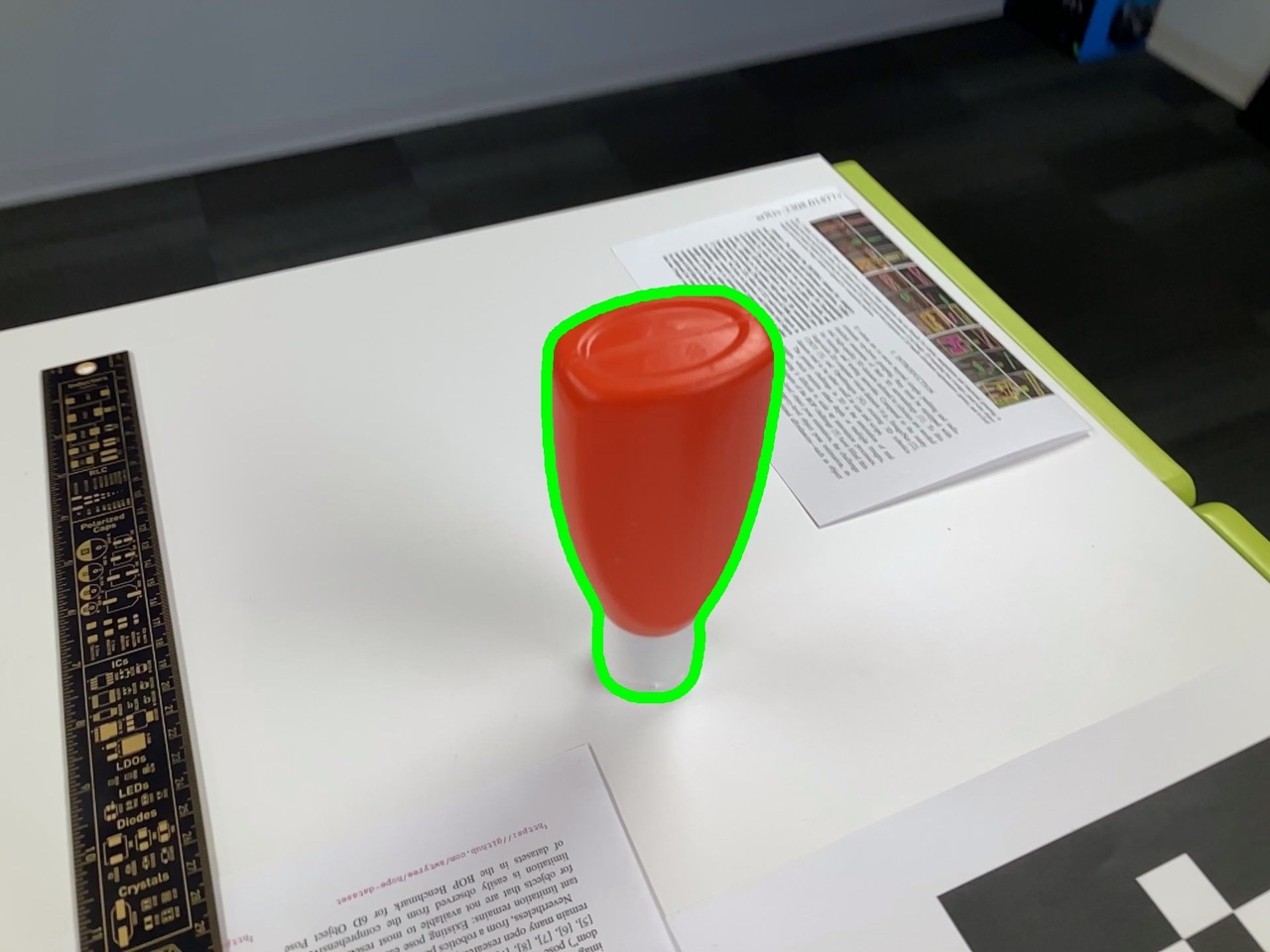} \\

\includegraphics[width=0.243\columnwidth, trim= 200 0 200 0, clip]{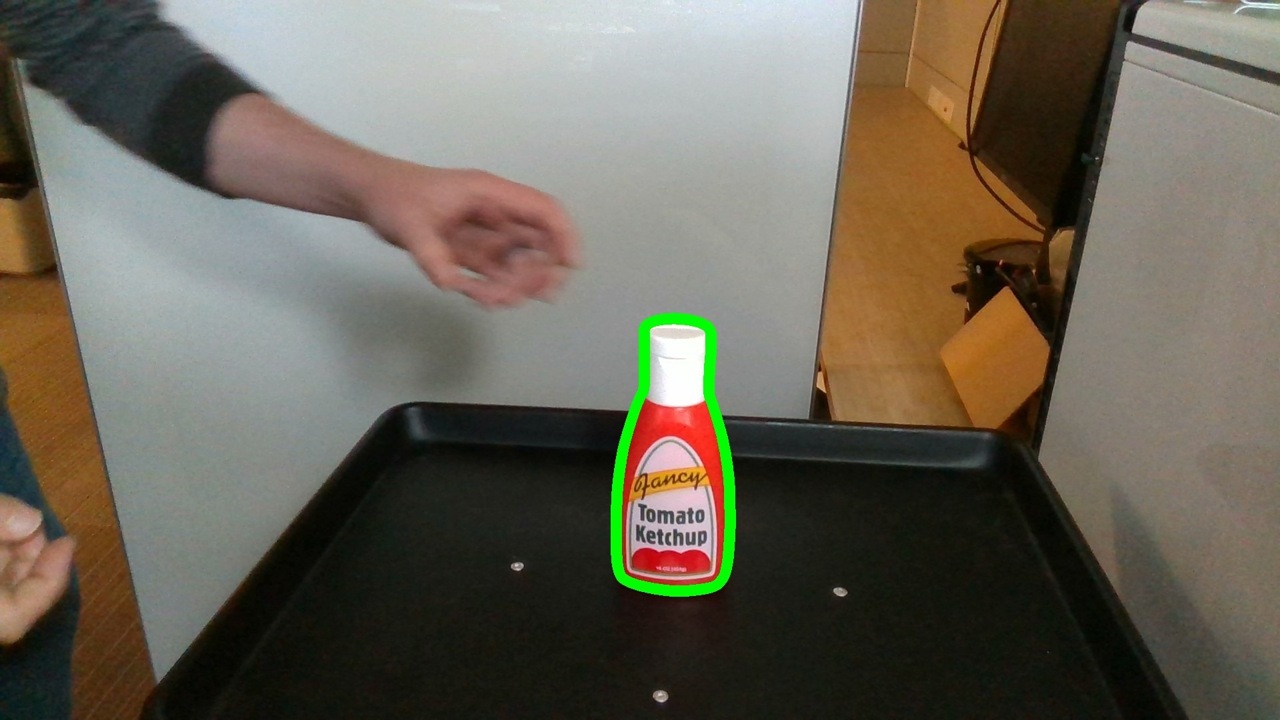} &
\includegraphics[width=0.243\columnwidth , trim= 200 0 200 0, clip]{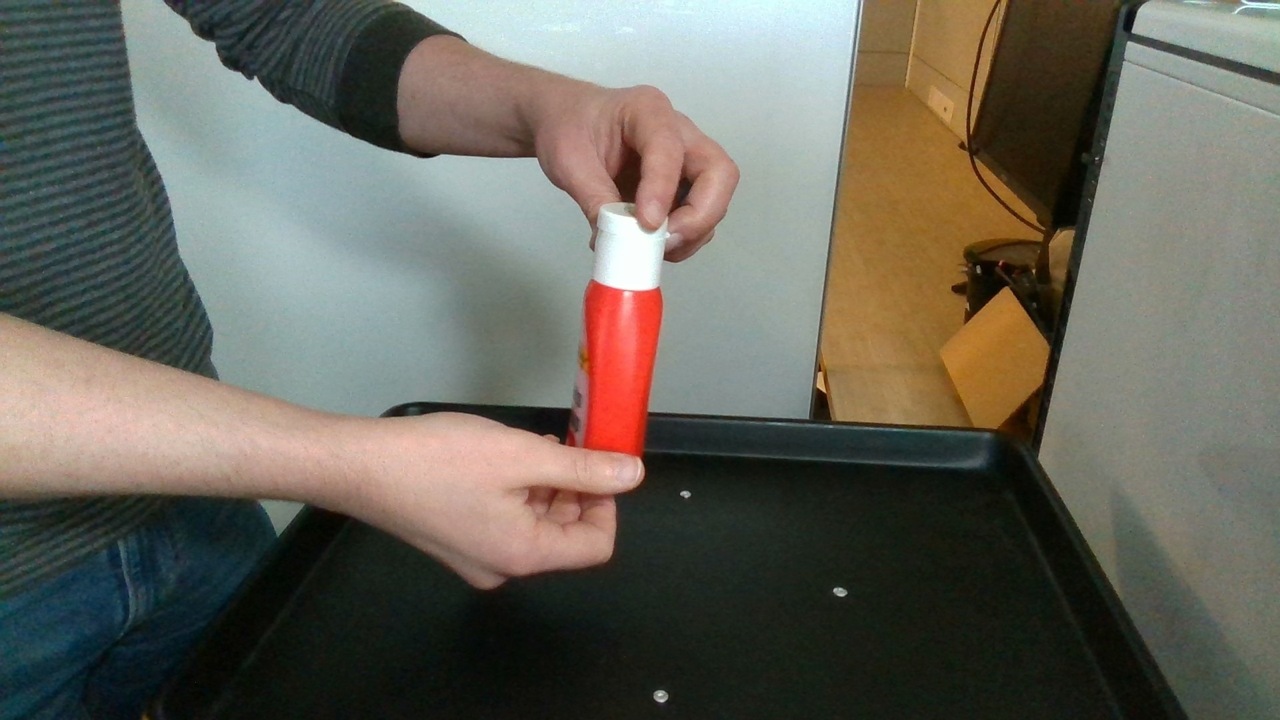} &
\includegraphics[width=0.243\columnwidth, trim= 200 0 200 0, clip]{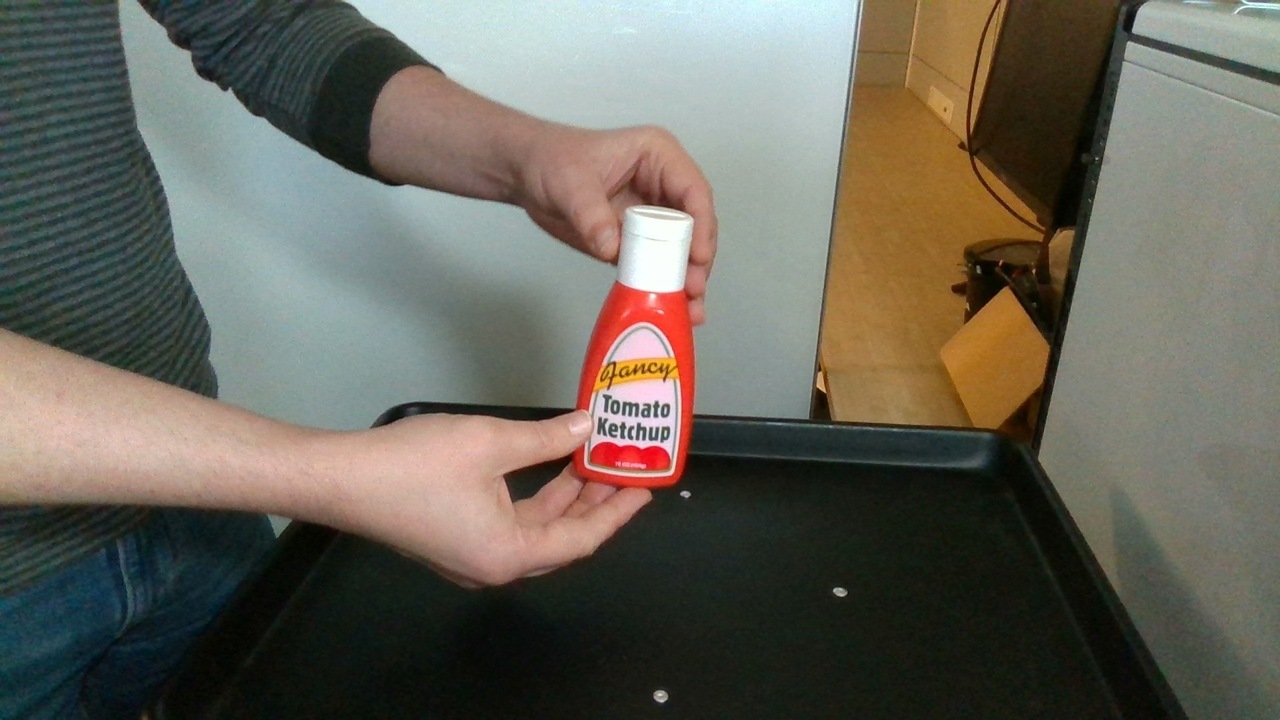} &
\includegraphics[width=0.243\columnwidth, trim= 200 0 200 0, clip]{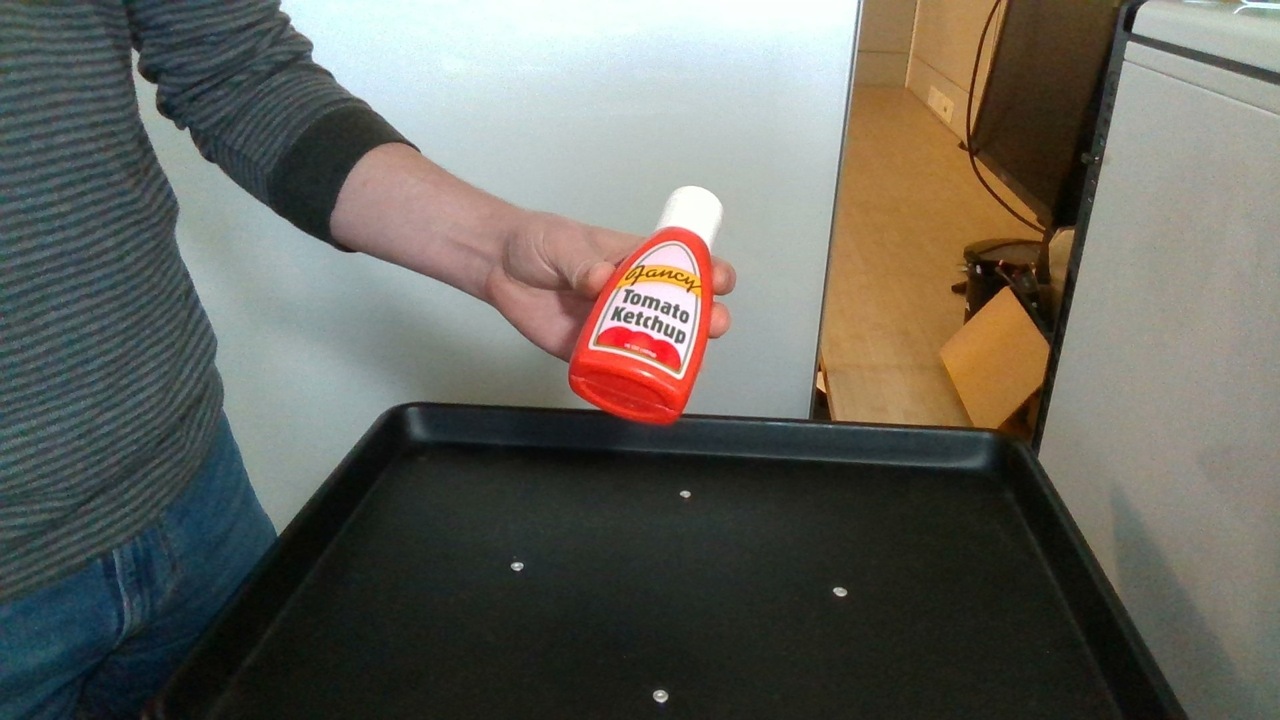} \\
\end{tabular}

\caption{\label{fig:onboarding} \textbf{Sample onboarding videos from HOPEv2~\cite{tyree2022hope}.} First two rows show sample frames from \emph{static onboarding} videos, one with the object standing upright and one with the object standing upside-down.
The third row shows sample frames from a \emph{dynamic onboarding} video where the object is manipulated by hands.
Ground-truth poses (shown with green contour) are provided for all frames of static but only the first frame of dynamic onboarding videos (see Sec.~\ref{sec:onboarding_videos} for details).
\vspace{-0.7em}
}

\end{center}
\end{figure}

In the new model-free tasks, 3D object models are not available and methods need to learn new objects from onboarding (reference) videos which are available in the BOP-H3 datasets. As shown in Fig.~\ref{fig:onboarding}, we define two types of video-based onboarding:

\customparagraph{Static onboarding:} The object is static and the camera is moving around the object, capturing all possible object views. Two videos are available, with the object standing upright in one and upside-down in the other. Such videos are useful for 3D object reconstruction by methods such as NeRF~\cite{mildenhall2021nerf} or Gaussian Splatting~\cite{kerbl20233d}. Object poses are available for all frames since these poses could be relatively easily obtained with tools like COLMAP~\cite{schonberger2016structure}.

\customparagraph{Dynamic onboarding:}
The object is manipulated by hands and the camera is either static (on a tripod) or dynamic (on a head-mounted device). This type of onboarding videos is useful for 3D object reconstruction methods such as BundleSDF~\cite{wen2023bundlesdf} or Hampali \etal~\cite{hampali2023inhand}. Ground-truth object poses are available only for the first frame to simulate a real-world setup (at least one ground-truth pose needs to be provided to define the object coordinate system necessary for the evaluation of object pose estimates). The dynamic onboarding setup is more challenging but more natural for AR/VR applications than the static setup.

\customparagraph{}In HOT3D, all onboarding videos are RGB (Aria) or monochrome (Quest 3).
In HOPEv2 and HANDAL, static onboarding videos are RGB and dynamic onboarding videos are RGB-D.

\subsection{Synthetic training dataset}

As in 2023, we provided over 2M synthetic training images showing 50K+ diverse objects from the
GSO~\cite{downs2022google} and ShapeNetCore~\cite{chang2015shapenet} datasets. These objects are not included in BOP-Classic nor BOP-H3 and the dataset can be therefore used for training methods for tasks on unseen objects.
The images were originally synthesized for MegaPose~\cite{megapose} using BlenderProc~\cite{denninger2019blenderproc,denninger2020blenderproc,denninger2023blenderproc2}.

\section{Challenge tracks} \label{sec:methodology}

In 2024, participants competed on seven challenge tracks on \emph{unseen objects} (target objects are not seen during training), with each track defined by a task (6D localization, 6D detection, 2D detection), type of object onboarding (model-based, model-free), and dataset group (BOP-Classic-Core, BOP-H3):
\vspace{1ex}
\begin{itemize}
    \item Track 1: Model-based 6D localization on BOP-Classic-Core
    \item Track 2: Model-based 6D detection on BOP-Classic-Core
    \item Track 3: Model-based 2D detection on BOP-Classic-Core
    \item Track 4: Model-based 6D detection on BOP-H3
    \item Track 5: Model-based 2D detection on BOP-H3
    \item Track 6: Model-free 6D detection on BOP-H3
    \item Track 7: Model-free 2D detection on BOP-H3
\end{itemize}
\vspace{1ex}
We primarily focused on the more practical detection tasks (Track 2-7) while keeping the classical model-based 6D localization on BOP-Classic-Core to enable direct comparison with previous years (Track 1).

On all tracks, methods go through three phases: (1)~\emph{training}, where methods can be trained on non-target objects, (2)~\emph{object onboarding}, where methods have to quickly onboard target objects, and (3)~\emph{inference}, where methods are expected to produce predictions about target objects seen in test images. The following sections describe these phases in more detail.

\subsection{Training phase}

At training time, methods can be trained on the provided dataset of 2M synthetic images, originally synthesized using BlenderProc~\cite{denninger2019blenderproc,denninger2020blenderproc} for MegaPose~\cite{megapose}. The images show training (non-target) objects annotated with ground-truth 6D poses, modal 2D segmentation masks (covering the visible object parts), and amodal 2D bounding boxes (covering the whole object silhouette, including the occluded parts). The dataset also includes color 3D mesh models of the objects. Methods are also free to use other datasets as long as the images did not show target objects.

\subsection{Object onboarding phase}
\label{sec:onboarding}
After training, methods can spend up to 5 minutes of the wall-clock time on a single computer with one GPU to onboard a target object.\footnote{The time is measured from the point right after the raw data (a 3D mesh model in the model-based setup and reference video(s) in the model-free setup) is loaded to the point when the object is onboarded.} The generated representation of onboarded objects (object templates, neural radiance fields, etc.) need to be fixed after onboarding (cannot be updated based on test images).

\customparagraph{Model-based onboarding:}
Methods are provided 3D models of target objects that are not seen at training. To onboard an object, methods can, \eg, fine-tune a neural network or render images of the 3D models, but cannot use any real images of the objects.

\customparagraph{Model-free onboarding:}
Methods are provided video(s) of target objects that are not seen at training. 3D models of the objects are not available. Methods can use either static or dynamic onboarding videos (Fig.~\ref{fig:onboarding}, Sec.~\ref{sec:bop-h3}), \eg, to reconstruct a 3D model, render novel views, or fine-tune a neural network.

\subsection{Inference phase}
\noindent\textbf{Input:} Methods are given a real-world test image showing an arbitrary number of instances of an arbitrary number of target objects, with all objects being from one specified dataset (\eg, YCB-V). Depending on the dataset, the test image may be RGB-D, RGB, or monochrome. In case of 6D localization, methods are additionally provided identities of objects visible in the test image in the form of a list $L = [(o_1, n_1),$ $\dots,$ $(o_m, n_m)]$, where $n_i$ is the number of instances of object $o_i$ for which at least $10\%$ of the projected surface area is visible in the test image.
No prior information about the visible object instances is provided in the detection tasks.

As most recent methods for 6D object localization and detection start with a 2D detection stage, participants in the 6D localization and detection tasks are also encouraged to evaluate their methods using default 2D detections/segmentations produced by CNOS~\cite{nguyen2023cnos}. Starting from the same 2D detections enables direct comparison of the object pose estimation stages.

\customparagraph{Output:} For each test image, methods produce predictions with confidence scores and run time.\footnote{Although the run time is measured on user machines, which may be different for each method, it still provides a general sense of how fast each method is. The time is measured from the point right after the raw data (the image, 3D models, \etc) is loaded to the point when predictions for all objects in the image are available (including the time needed to generate default 2D detections if used).}
Predictions are in the form of amodal 2D bounding boxes for 2D detection and in the form of 6D object poses for 6D localization and detection.
A 6D object pose is defined by a matrix $\textbf{P} = [\mathbf{R} \; | \; \mathbf{t}]$, where $\mathbf{R}$ is a 3D rotation matrix, and $\mathbf{t}$ is a 3D translation vector from the 3D object model space to the 3D camera space.
In all tasks, methods need to produce predictions only for object instances for which at least $10\%$ of the projected surface area is visible in the given test image.

\subsection{Evaluation methodology}
\label{sec:evaluation_methodology}
\noindent \textbf{Measuring error of 6D object poses.}
The error of an estimated pose \wrt the ground-truth pose can be calculated by the following pose-error functions (see Sec.~2.2 of~\cite{hodan2020bop} for details): (1) MSSD (Maximum Symmetry-Aware Surface Distance) which considers a set of pre-identified global object symmetries and measures the surface deviation in 3D, (2) MSPD (Maximum Symmetry-Aware Projection Distance) which considers the object symmetries and measures the perceivable deviation, (3) VSD (Visible Surface Discrepancy) which treats indistinguishable poses as equivalent by considering only the visible object part.
An estimated pose is considered correct \wrt a pose-error function~$e$, if $e < \theta_e$, where $e \in \{\text{MSSD},\;\text{MSPD},\;\text{VSD}\}$ and $\theta_e$ is a threshold of correctness.

\customparagraph{Evaluating 6D object localization.}
The fraction of
object instances for which a correct pose is estimated is called Recall.
The Average Recall \wrt a pose-error function~$e$ on a dataset $d$, denoted as $\text{AR}_{e,d}$, is defined as the average of the Recall rates calculated for multiple settings of $\theta_e$ and, in the case of $\text{VSD}$, also for multiple settings of the misalignment tolerance~$\tau$.
The accuracy on a dataset $d$ is measured by: $\text{AR}_d = (\text{AR}_{\text{VSD},d} + \text{AR}_{\text{MSSD},d} + \text{AR}_{\text{MSPD},d}) \, / \, 3$, and the overall accuracy, denoted as AR, is defined as the average of $\text{AR}_{d}$ scores over selected datasets.\footnote{When calculating AR, scores are not averaged over objects before averaging over datasets, which is done when calculating AP (in 2D/6D detection) to comply with the original COCO evaluation methodology~\cite{lin2014microsoft}.
}

\customparagraph{Evaluating 2D object detection.}
The detection accuracy is measured by the Average Precision (AP, also known as mAP), following the evaluation methodology from the COCO 2020 Object Detection Challenge~\cite{lin2014microsoft}. Specifically, a per-object $\text{AP}_o$ score is calculated by averaging the precision values at multiple thresholds on the Intersection over Union (IoU) of 2D bounding boxes: $[0.5,\;0.55,\;0.6,\;\dots ,\;0.95]$. The accuracy of a method on a dataset~$d$ is measured by $\text{AP}_d$ calculated by averaging per-object $\text{AP}_{o,d}$ scores over objects from the dataset. The overall accuracy, denoted as AP, is defined as the average of the per-dataset $\text{AP}_d$ scores.\footnote{\label{det_note}Up to $100$ most confident detections per image are considered. Correct detections for instances visible from less than $10\%$ are
not counted as false positives.}

\customparagraph{Evaluating 6D object detection.}
The detection accuracy is measured by the Average Precision (AP), similarly as for 2D object detection but using pose-error functions instead of the IoU of 2D bounding boxes.
Specifically, for each pose-error function $e \in \{\text{MSSD},\;\text{MSPD}\}$\footnote{The VSD pose-error function is not considered as it is more expensive to calculate (6D detection requires evaluating more pose estimates than 6D localization) and needs depth images which are not available in HOT3D and HANDAL.}, the~per-object accuracy $\text{AP}_{e,o}$ is defined by averaging the precision values calculated at multiple thresholds of correctness $\theta_e$ (defined in Sec.~2.4 of~\cite{hodan2020bop}).
The accuracy $\text{AP}_{e,d}$ on a dataset $d$ \wrt a pose-error function $e$ is calculated by averaging per-object scores $\text{AP}_{e,o}$ over objects from the dataset.
The accuracy on the dataset $d$ is then defined as $\text{AP}_{d}=(\text{AP}_{\text{MSSD},d}+\text{AP}_{\text{MSPD},d}) \, / \, 2$, and the overall accuracy, denoted as AP, is defined as the average of $\text{AP}_{d}$ scores over selected datasets.\footref{det_note}

\section{Results and discussion} 
\label{sec:evaluation}

\setlength{\tabcolsep}{3pt}
\begin{table*}[!t]
    \renewcommand{\arraystretch}{0.95}
    \tiny
    \centering
    \begin{tabularx}{\linewidth}{rlcllllllYYYYYYYYY}
    \toprule
    \# & Method & Awards & Year & Det./seg. & Refinement & Train im. & ...type & Test image & LM-O & T-LESS & TUD-L & IC-BIN & ITODD & HB & YCB-V & \mbox{$\text{AR}$} & Time  \\ 
        \midrule
        1 & FreeZeV2.1~\cite{freeze}  & \iconBest & \cellcolor{ccol!100}2024 & Custom & \cellcolor{ccol!100}ICP & - & - & RGB-D  & \cellcolor{arcol!77.1}77.1  & \cellcolor{arcol!75.5}75.5  & \cellcolor{arcol!97.6}97.6  & \cellcolor{arcol!69.7}69.7  & \cellcolor{arcol!74.2}74.2  & \cellcolor{arcol!89.2}89.2  & \cellcolor{arcol!91.5}91.5  & \cellcolor{avgcol!82.1}82.1  & \cellcolor{timecol!24.9}$\phantom{0}$24.9 \\

        2 & FRTPose.v1 (SAM6D-FastSAM) &  & \cellcolor{ccol!100}2024 & SAM6D-FastSAM & FoundationPose & RGB-D & \cellcolor{ccol!100}PBR & RGB-D  & \cellcolor{arcol!77.8}77.8 & \cellcolor{arcol!76.6}76.6 & \cellcolor{arcol!94.0}94.0 & \cellcolor{arcol!70.2}70.2 & \cellcolor{arcol!73.7}73.7 & \cellcolor{arcol!89.6}89.6 & \cellcolor{arcol!91.0}91.0 & \cellcolor{avgcol!81.8}81.8 & \cellcolor{timecol!40.1}$\phantom{0}$40.1 \\

        3 & FRTPose.v1 (Default Detections)  & \iconDefault & \cellcolor{ccol!100}2024 & \cellcolor{ccol!100}CNOS-FastSAM & FoundationPose & RGB-D & \cellcolor{ccol!100}PBR & RGB-D & \cellcolor{arcol!77.7}77.7 & \cellcolor{arcol!76.3}76.3 & \cellcolor{arcol!94.0}94.0 & \cellcolor{arcol!70.5}70.5 & \cellcolor{arcol!73.5}73.5 & \cellcolor{arcol!89.6}89.6 & \cellcolor{arcol!91.0}91.0 & \cellcolor{avgcol!81.8}81.8 & \cellcolor{timecol!46.5}$\phantom{0}$46.5 \\

        4 & FRTPose.v1 (MUSE) &  & \cellcolor{ccol!100}2024 & MUSE & FoundationPose & RGB-D & \cellcolor{ccol!100}PBR & RGB-D & \cellcolor{arcol!78.6}78.6 & \cellcolor{arcol!76.8}76.8 & \cellcolor{arcol!94.2}94.2 & \cellcolor{arcol!70.6}70.6 & \cellcolor{arcol!71.0}71.0 & \cellcolor{arcol!90.3}90.3 & \cellcolor{arcol!91.0}91.0 & \cellcolor{avgcol!81.8}81.8 & \cellcolor{timecol!27.6}$\phantom{0}$27.6\\

        5 & FreeZeV2~\cite{freeze} &  & \cellcolor{ccol!100}2024 & Custom & \cellcolor{ccol!100}ICP & - & - & RGB-D & \cellcolor{arcol!76.4}76.4 & \cellcolor{arcol!70.8}70.8 & \cellcolor{arcol!97.2}97.2 & \cellcolor{arcol!65.4}65.4 & \cellcolor{arcol!67.9}67.9 & \cellcolor{arcol!85.9}85.9 & \cellcolor{arcol!90.6}90.6 & \cellcolor{avgcol!79.2}79.2 & \cellcolor{timecol!17.2}$\phantom{0}$17.2 \\

        6 & FRTPose (SAM6D-FastSAM) &  & \cellcolor{ccol!100}2024 & SAM6D-FastSAM & FoundationPose & RGB-D & \cellcolor{ccol!100}PBR & RGB-D & \cellcolor{arcol!78.3}78.3 & \cellcolor{arcol!71.7}71.7 & \cellcolor{arcol!92.5}92.5 & \cellcolor{arcol!60.1}60.1 & \cellcolor{arcol!64.6}64.6 & \cellcolor{arcol!89.6}89.6 & \cellcolor{arcol!91.3}91.3 & \cellcolor{avgcol!78.3}78.3 & \cellcolor{timecol!20.7}$\phantom{0}$20.7\\

        7 & FRTPose (Default Detections) &  & \cellcolor{ccol!100}2024 & \cellcolor{ccol!100}CNOS-FastSAM & FoundationPose & RGB-D & \cellcolor{ccol!100}PBR & RGB-D & \cellcolor{arcol!78.3}78.3 & \cellcolor{arcol!71.4}71.4 & \cellcolor{arcol!92.2}92.2 & \cellcolor{arcol!59.0}59.0 & \cellcolor{arcol!61.8}61.8 & \cellcolor{arcol!89.6}89.6 & \cellcolor{arcol!91.3}91.3 & \cellcolor{avgcol!77.7}77.7 & \cellcolor{timecol!23.4}$\phantom{0}$23.4\\

        8 & Co-op (F3DT2D, 5 Hypo)~\cite{Moon2025Coop} &  & \cellcolor{ccol!100}2024 & F3DT2D & Co-op & RGB-D & \cellcolor{ccol!100}PBR & RGB-D & \cellcolor{arcol!73.8}73.8 & \cellcolor{arcol!69.5}69.5 & \cellcolor{arcol!92.9}92.9 & \cellcolor{arcol!63.5}63.5 & \cellcolor{arcol!62.9}62.9 & \cellcolor{arcol!87.8}87.8 & \cellcolor{arcol!89.8}89.8 & \cellcolor{avgcol!77.1}77.1 & \cellcolor{timecol!6.9}$\phantom{0}$$\phantom{0}$6.9\\

        9 & Co-op (F3DT2D, 1 Hypo)~\cite{Moon2025Coop}  & \iconFast & \cellcolor{ccol!100}2024 & F3DT2D & Co-op & RGB-D & \cellcolor{ccol!100}PBR & RGB-D & \cellcolor{arcol!73.0}73.0 & \cellcolor{arcol!68.0}68.0 & \cellcolor{arcol!92.9}92.9 & \cellcolor{arcol!62.4}62.4 & \cellcolor{arcol!60.0}60.0 & \cellcolor{arcol!86.3}86.3 & \cellcolor{arcol!88.6}88.6 & \cellcolor{avgcol!75.9}75.9 & \cellcolor{timecol!0.8}$\phantom{0}$$\phantom{0}$0.8\\

        10 & Co-op (CNOS, 5 Hypo)~\cite{Moon2025Coop} &  & \cellcolor{ccol!100}2024 & \cellcolor{ccol!100}CNOS-FastSAM & Co-op & RGB-D & \cellcolor{ccol!100}PBR & RGB-D & \cellcolor{arcol!73.0}73.0 & \cellcolor{arcol!66.4}66.4 & \cellcolor{arcol!90.5}90.5 & \cellcolor{arcol!59.7}59.7 & \cellcolor{arcol!61.3}61.3 & \cellcolor{arcol!87.1}87.1 & \cellcolor{arcol!88.7}88.7 & \cellcolor{avgcol!75.2}75.2 & \cellcolor{timecol!7.2}$\phantom{0}$$\phantom{0}$7.2\\

        11 & Co-op (CNOS, 1 Hypo)~\cite{Moon2025Coop} &  & \cellcolor{ccol!100}2024 & \cellcolor{ccol!100}CNOS-FastSAM & Co-op & RGB-D & \cellcolor{ccol!100}PBR & RGB-D & \cellcolor{arcol!71.5}71.5 & \cellcolor{arcol!64.6}64.6 & \cellcolor{arcol!90.5}90.5 & \cellcolor{arcol!57.5}57.5 & \cellcolor{arcol!58.2}58.2 & \cellcolor{arcol!85.7}85.7 & \cellcolor{arcol!87.4}87.4 & \cellcolor{avgcol!73.6}73.6 & \cellcolor{timecol!2.3}$\phantom{0}$$\phantom{0}$2.3\\

        12 & FoundationPose~\cite{foundationPose}  & \iconOpen & \cellcolor{ccol!100}2024 & SAM6D & FondationPose & RGB-D & \cellcolor{ccol!100}PBR & RGB-D & \cellcolor{arcol!75.6}75.6 & \cellcolor{arcol!64.6}64.6 & \cellcolor{arcol!92.3}92.3 & \cellcolor{arcol!50.8}50.8 & \cellcolor{arcol!58.0}58.0 & \cellcolor{arcol!83.5}83.5 & \cellcolor{arcol!88.9}88.9 & \cellcolor{avgcol!73.4}73.4 & \cellcolor{timecol!29.3}$\phantom{0}$29.3\\

        13 & FRTPose (SAM6D-FastSAM \& top k) &  & \cellcolor{ccol!100}2024 & SAM6D-FastSAM & FondationPose & RGB-D & \cellcolor{ccol!100}PBR & RGB-D & \cellcolor{arcol!70.3}70.3 & \cellcolor{arcol!58.1}58.1 & \cellcolor{arcol!87.1}87.1 & \cellcolor{arcol!59.9}59.9 & \cellcolor{arcol!64.4}64.4 & \cellcolor{arcol!80.4}80.4 & \cellcolor{arcol!86.9}86.9 & \cellcolor{avgcol!72.4}72.4 & \cellcolor{timecol!0.8}$\phantom{0}$$\phantom{0}$0.8\\

        14 & Co-op (CNOS, Coarse)~\cite{Moon2025Coop} &  & \cellcolor{ccol!100}2024 & \cellcolor{ccol!100}CNOS-FastSAM & - & RGB-D & \cellcolor{ccol!100}PBR & RGB-D & \cellcolor{arcol!70.0}70.0 & \cellcolor{arcol!64.2}64.2 & \cellcolor{arcol!87.9}87.9 & \cellcolor{arcol!56.4}56.4 & \cellcolor{arcol!56.6}56.6 & \cellcolor{arcol!84.2}84.2 & \cellcolor{arcol!85.3}85.3 & \cellcolor{avgcol!72.1}72.1 & \cellcolor{timecol!1.0}$\phantom{0}$$\phantom{0}$1.0\\

        15 & GZS6D-BP(coarse+refine+teaser) &  & \cellcolor{ccol!100}2024 & - & \cellcolor{ccol!100}Teaserpp & RGB-D & - & RGB-D & \cellcolor{arcol!67.8}67.8 & \cellcolor{arcol!69.4}69.4 & \cellcolor{arcol!92.2}92.2 & \cellcolor{arcol!55.0}55.0 & \cellcolor{arcol!59.7}59.7 & \cellcolor{arcol!80.3}80.3 & \cellcolor{arcol!77.2}77.2 & \cellcolor{avgcol!71.7}71.7 & \cellcolor{timecol!6.5}$\phantom{0}$$\phantom{0}$6.5\\

        16 & FreeZe (SAM6D)~\cite{freeze} &  & \cellcolor{ccol!100}2024 & SAM6D & \cellcolor{ccol!100}ICP & - & - & RGB-D & \cellcolor{arcol!71.6}71.6 & \cellcolor{arcol!53.1}53.1 & \cellcolor{arcol!94.9}94.9 & \cellcolor{arcol!54.5}54.5 & \cellcolor{arcol!58.6}58.6 & \cellcolor{arcol!79.6}79.6 & \cellcolor{arcol!84.0}84.0 & \cellcolor{avgcol!70.9}70.9 & \cellcolor{timecol!11.5}$\phantom{0}$11.5\\

        17 & SAM6D~\cite{lin2023sam} &  & \cellcolor{ccol!100}2024 & SAM6D-SAM & - & RGB-D & \cellcolor{ccol!100}PBR & RGB-D & \cellcolor{arcol!69.9}69.9 & \cellcolor{arcol!51.5}51.5 & \cellcolor{arcol!90.4}90.4 & \cellcolor{arcol!58.8}58.8 & \cellcolor{arcol!60.2}60.2 & \cellcolor{arcol!77.6}77.6 & \cellcolor{arcol!84.5}84.5 & \cellcolor{avgcol!70.4}70.4 & \cellcolor{timecol!4.4}$\phantom{0}$$\phantom{0}$4.4\\

        18 & FreeZe (CNOS)~\cite{freeze} &  & \cellcolor{ccol!100}2024 & \cellcolor{ccol!100}CNOS-FastSAM & \cellcolor{ccol!100}ICP & - & - & RGB-D & \cellcolor{arcol!68.9}68.9 & \cellcolor{arcol!52.0}52.0 & \cellcolor{arcol!93.6}93.6 & \cellcolor{arcol!49.9}49.9 & \cellcolor{arcol!56.1}56.1 & \cellcolor{arcol!79.0}79.0 & \cellcolor{arcol!85.3}85.3 & \cellcolor{avgcol!69.3}69.3 & \cellcolor{timecol!13.5}$\phantom{0}$13.5\\

        19 & GigaPose+GenFlow+kabsch (5 hypoth)~\cite{gigaPose,genflow} &  & \cellcolor{ccol!100}2024 & \cellcolor{ccol!100}CNOS-FastSAM & GenFlow & RGB-D & \cellcolor{ccol!100}PBR & RGB-D & \cellcolor{arcol!67.8}67.8 & \cellcolor{arcol!55.6}55.6 & \cellcolor{arcol!81.1}81.1 & \cellcolor{arcol!56.3}56.3 & \cellcolor{arcol!57.5}57.5 & \cellcolor{arcol!79.1}79.1 & \cellcolor{arcol!82.5}82.5 & \cellcolor{avgcol!68.6}68.6 & \cellcolor{timecol!11.1}$\phantom{0}$11.1 \\

        20 & Co-op (F3DT2D, 5 Hypo)~\cite{Moon2025Coop}  & \iconRGB & \cellcolor{ccol!100}2024 & F3DT2D & Co-op & RGB-D & \cellcolor{ccol!100}PBR & \cellcolor{ccol!100}RGB & \cellcolor{arcol!67.5}67.5 & \cellcolor{arcol!68.2}68.2 & \cellcolor{arcol!76.7}76.7 & \cellcolor{arcol!58.9}58.9 & \cellcolor{arcol!50.6}50.6 & \cellcolor{arcol!85.6}85.6 & \cellcolor{arcol!69.7}69.7 & \cellcolor{avgcol!68.2}68.2 & \cellcolor{timecol!3.9}$\phantom{0}$$\phantom{0}$3.9\\
                
        21 & GenFlow-MultiHypo16~\cite{genflow} &  & 2023 & \cellcolor{ccol!100}CNOS-FastSAM & GenFlow & RGB-D & \cellcolor{ccol!100}PBR & RGB-D  & \cellcolor{arcol!63.5}63.5 & \cellcolor{arcol!52.1}52.1 & \cellcolor{arcol!86.2}86.2 & \cellcolor{arcol!53.4}53.4 & \cellcolor{arcol!55.4}55.4 & \cellcolor{arcol!77.9}77.9 & \cellcolor{arcol!83.3}83.3 & 
        \cellcolor{avgcol!67.4}67.4 &
        \cellcolor{timecol!34.6}$\phantom{0}$34.6 \\
        
        22 & GenFlow-MultiHypo~\cite{genflow} &  & 2023 & \cellcolor{ccol!100}CNOS-FastSAM & GenFlow & \cellcolor{ccol!100}RGB & \cellcolor{ccol!100}PBR & RGB-D &  \cellcolor{arcol!62.2}62.2 & \cellcolor{arcol!50.9}50.9 & \cellcolor{arcol!84.9}84.9 & \cellcolor{arcol!52.4}52.4 & \cellcolor{arcol!54.4}54.4 & \cellcolor{arcol!77.0}77.0 & \cellcolor{arcol!81.8}81.8 & 
        \cellcolor{avgcol!66.2}66.2 & \cellcolor{timecol!21.5}$\phantom{0}$21.5 \\

        23 & SAM6D-FastSAM~\cite{lin2023sam} &  & \cellcolor{ccol!100}2024 & SAM6D-FastSAM & - & RGB-D & \cellcolor{ccol!100}PBR & RGB-D & \cellcolor{arcol!66.7}66.7 & \cellcolor{arcol!48.5}48.5 & \cellcolor{arcol!82.9}82.9 & \cellcolor{arcol!51.0}51.0 & \cellcolor{arcol!57.2}57.2 & \cellcolor{arcol!73.6}73.6 & \cellcolor{arcol!83.4}83.4 & \cellcolor{avgcol!66.2}66.2 & \cellcolor{timecol!1.4}$\phantom{0}$$\phantom{0}$1.4\\

        24 & Co-op (CNOS, 5 Hypo)~\cite{Moon2025Coop} &  & \cellcolor{ccol!100}2024 & \cellcolor{ccol!100}CNOS-FastSAM & Co-op & RGB-D & \cellcolor{ccol!100}PBR & \cellcolor{ccol!100}RGB & \cellcolor{arcol!65.5}65.5 & \cellcolor{arcol!64.8}64.8 & \cellcolor{arcol!72.9}72.9 & \cellcolor{arcol!54.4}54.4 & \cellcolor{arcol!49.1}49.1 & \cellcolor{arcol!85.0}85.0 & \cellcolor{arcol!68.9}68.9 & \cellcolor{avgcol!65.8}65.8 & \cellcolor{timecol!4.2}$\phantom{0}$$\phantom{0}$4.2\\

        25 & SAM6D-CNOSfastSAM~\cite{lin2023sam} &  & \cellcolor{ccol!100}2024 & \cellcolor{ccol!100}CNOS-FastSAM & - & RGB-D & \cellcolor{ccol!100}PBR & RGB-D & \cellcolor{arcol!65.1}65.1 & \cellcolor{arcol!47.9}47.9 & \cellcolor{arcol!82.5}82.5 & \cellcolor{arcol!49.7}49.7 & \cellcolor{arcol!56.2}56.2 & \cellcolor{arcol!73.8}73.8 & \cellcolor{arcol!81.5}81.5 & \cellcolor{avgcol!65.3}65.3 & \cellcolor{timecol!1.3}$\phantom{0}$$\phantom{0}$1.3\\

        26 & Co-op (CNOS, 1 Hypo)~\cite{Moon2025Coop} &  & \cellcolor{ccol!100}2024 & \cellcolor{ccol!100}CNOS-FastSAM & Co-op & RGB-D & \cellcolor{ccol!100}PBR & \cellcolor{ccol!100}RGB & \cellcolor{arcol!64.2}64.2 & \cellcolor{arcol!63.5}63.5 & \cellcolor{arcol!71.7}71.7 & \cellcolor{arcol!51.2}51.2 & \cellcolor{arcol!47.3}47.3 & \cellcolor{arcol!83.2}83.2 & \cellcolor{arcol!67.0}67.0 & \cellcolor{avgcol!64.0}64.0 & \cellcolor{timecol!1.7}$\phantom{0}$$\phantom{0}$1.7\\
        
        27 & Megapose-CNOS+Multih\_Teaserpp-10~\cite{megapose} &  & 2023 & \cellcolor{ccol!100}CNOS-FastSAM & \cellcolor{ccol!100}Teaserpp & \cellcolor{ccol!100}RGB & \cellcolor{ccol!100}PBR & RGB-D & \cellcolor{arcol!62.6}62.6 & \cellcolor{arcol!48.7}48.7 & \cellcolor{arcol!85.1}85.1 & \cellcolor{arcol!46.7}46.7 & \cellcolor{arcol!46.8}46.8 & \cellcolor{arcol!73.0}73.0 & \cellcolor{arcol!76.4}76.4 & 
        \cellcolor{avgcol!62.8}62.8 & \cellcolor{timecol!100}142.0 \\
        
        28 & Megapose-CNOS+Multih\_Teaserpp~\cite{megapose} &  & 2023 & \cellcolor{ccol!100}CNOS-FastSAM & \cellcolor{ccol!100}Teaserpp & \cellcolor{ccol!100}RGB & \cellcolor{ccol!100}PBR & RGB-D &  \cellcolor{arcol!62.0}62.0 & \cellcolor{arcol!48.5}48.5 & \cellcolor{arcol!84.6}84.6 & \cellcolor{arcol!46.2}46.2 & \cellcolor{arcol!46.0}46.0 & \cellcolor{arcol!72.5}72.5 & \cellcolor{arcol!76.4}76.4 & 
        \cellcolor{avgcol!62.3}62.3 & \cellcolor{timecol!100.}116.6 \\

        29 & SAM6D-ZeroPose~\cite{lin2023sam} &  & \cellcolor{ccol!100}2024 & SAM6D & - & RGB-D & \cellcolor{ccol!100}PBR & RGB-D & \cellcolor{arcol!63.5}63.5 & \cellcolor{arcol!43.0}43.0 & \cellcolor{arcol!80.2}80.2 & \cellcolor{arcol!51.8}51.8 & \cellcolor{arcol!48.4}48.4 & \cellcolor{arcol!69.1}69.1 & \cellcolor{arcol!79.2}79.2 & \cellcolor{avgcol!62.2}62.2 & \cellcolor{timecol!5.5}$\phantom{0}$$\phantom{0}$5.5\\
        
        30 & SAM6D-CNOSmask~\cite{lin2023sam} &  & 2023 & \cellcolor{ccol!100}CNOS-FastSAM & - & RGB-D & \cellcolor{ccol!100}PBR & RGB-D &   \cellcolor{arcol!64.8}64.8 & \cellcolor{arcol!48.3}48.3 & \cellcolor{arcol!79.4}79.4 & \cellcolor{arcol!50.4}50.4 & \cellcolor{arcol!35.1}35.1 & \cellcolor{arcol!72.7}72.7 & \cellcolor{arcol!80.4}80.4 & 
        \cellcolor{avgcol!61.6}61.6 & \cellcolor{timecol!3.872}$\phantom{0}\phantom{0}$3.9 \\
        
        31 & PoZe (CNOS)&  & 2023 & \cellcolor{ccol!100}CNOS-FastSAM & \cellcolor{ccol!100}ICP & RGB-D & Custom & RGB-D &  \cellcolor{arcol!64.4}64.4 & \cellcolor{arcol!49.4}49.4 & \cellcolor{arcol!92.4}92.4 & \cellcolor{arcol!40.9}40.9 & \cellcolor{arcol!51.6}51.6 & \cellcolor{arcol!71.2}71.2 & \cellcolor{arcol!61.1}61.1 & 
        \cellcolor{avgcol!61.6}61.6 &
        \cellcolor{timecol!100}159.4 \\

        32 & GigaPose+GenFlow (5 hypo)~\cite{gigaPose,genflow} &  & \cellcolor{ccol!100}2024 & \cellcolor{ccol!100}CNOS-FastSAM & GenFlow & RGB-D & \cellcolor{ccol!100}PBR & \cellcolor{ccol!100}RGB & \cellcolor{arcol!63.1}63.1 & \cellcolor{arcol!58.2}58.2 & \cellcolor{arcol!66.4}66.4 & \cellcolor{arcol!49.8}49.8 & \cellcolor{arcol!45.3}45.3 & \cellcolor{arcol!75.6}75.6 & \cellcolor{arcol!65.2}65.2 & \cellcolor{avgcol!60.5}60.5 & \cellcolor{timecol!10.6}$\phantom{0}$10.6\\

        33 & FoundPose+FeatRef+Megapose-5hyp~\cite{foundPose,megapose} &  & \cellcolor{ccol!100}2024 & \cellcolor{ccol!100}CNOS-FastSAM & MegaPose+FeatRef & \cellcolor{ccol!100}RGB & \cellcolor{ccol!100}PBR & \cellcolor{ccol!100}RGB &\cellcolor{arcol!61.0}61.0 & \cellcolor{arcol!57.0}57.0 & \cellcolor{arcol!69.3}69.3 & \cellcolor{arcol!47.9}47.9 & \cellcolor{arcol!40.7}40.7 & \cellcolor{arcol!72.3}72.3 & \cellcolor{arcol!69.0}69.0 & \cellcolor{avgcol!59.6}59.6 & \cellcolor{timecol!20.5}$\phantom{0}$20.5\\

        34 & OPFormer-Megapose refinement (CNOS) &  & \cellcolor{ccol!100}2024 & \cellcolor{ccol!100}CNOS-FastSAM & MegaPose & \cellcolor{ccol!100}RGB & \cellcolor{ccol!100}PBR & \cellcolor{ccol!100}RGB & \cellcolor{arcol!59.6}59.6 & \cellcolor{arcol!53.4}53.4 & \cellcolor{arcol!69.3}69.3 & \cellcolor{arcol!47.0}47.0 & \cellcolor{arcol!39.2}39.2 & \cellcolor{arcol!76.0}76.0 & \cellcolor{arcol!67.0}67.0 & \cellcolor{avgcol!58.8}58.8 & \cellcolor{timecol!1.5}$\phantom{0}$$\phantom{0}$1.5 \\

        35 & GigaPose (Add) + Megapose (5 hypo)~\cite{gigaPose,megapose}  &  & \cellcolor{ccol!100}2024 & \cellcolor{ccol!100}CNOS-FastSAM & MegaPose & \cellcolor{ccol!100}RGB & \cellcolor{ccol!100}PBR & \cellcolor{ccol!100}RGB & \cellcolor{arcol!60.4}60.4 & \cellcolor{arcol!57.6}57.6 & \cellcolor{arcol!64.8}64.8 & \cellcolor{arcol!48.2}48.2 & \cellcolor{arcol!39.8}39.8 & \cellcolor{arcol!72.4}72.4 & \cellcolor{arcol!66.6}66.6 & \cellcolor{avgcol!58.5}58.5 & \cellcolor{timecol!10.8}$\phantom{0}$10.8\\

        36 & Co-op (CNOS, Coarse)~\cite{Moon2025Coop} &  & \cellcolor{ccol!100}2024 & \cellcolor{ccol!100}CNOS-FastSAM & - & RGB-D & \cellcolor{ccol!100}PBR & \cellcolor{ccol!100}RGB & \cellcolor{arcol!59.7}59.7 & \cellcolor{arcol!59.2}59.2 & \cellcolor{arcol!64.2}64.2 & \cellcolor{arcol!45.8}45.8 & \cellcolor{arcol!39.1}39.1 & \cellcolor{arcol!78.1}78.1 & \cellcolor{arcol!62.6}62.6 & \cellcolor{avgcol!58.4}58.4 & \cellcolor{timecol!1.0}$\phantom{0}$$\phantom{0}$1.0\\

        37 & GigaPose+MegaPose (5 Hypo)~\cite{gigaPose,megapose} & &\cellcolor{ccol!100}2024 & \cellcolor{ccol!100}CNOS-FastSAM & MegaPose & \cellcolor{ccol!100}RGB & \cellcolor{ccol!100}PBR & \cellcolor{ccol!100}RGB & \cellcolor{arcol!59.8}59.8 & \cellcolor{arcol!56.5}56.5 & \cellcolor{arcol!63.1}63.1 & \cellcolor{arcol!47.3}47.3 & \cellcolor{arcol!39.7}39.7 & \cellcolor{arcol!72.2}72.2 & \cellcolor{arcol!66.1}66.1 & \cellcolor{avgcol!57.8}57.8 & \cellcolor{timecol!7.7}$\phantom{0}$$\phantom{0}$7.7\\

        38 & GenFlow-MultiHypo16~\cite{genflow} &  & 2023 & \cellcolor{ccol!100}CNOS-FastSAM & GenFlow & RGB-D & \cellcolor{ccol!100}PBR & \cellcolor{ccol!100}RGB & \cellcolor{arcol!57.2}57.2 & \cellcolor{arcol!52.8}52.8 & \cellcolor{arcol!68.8}68.8 & \cellcolor{arcol!45.8}45.8 & \cellcolor{arcol!39.8}39.8 & \cellcolor{arcol!74.6}74.6 & \cellcolor{arcol!64.2}64.2 & \cellcolor{avgcol!57.6}57.6 & \cellcolor{timecol!40.5}$\phantom{0}$40.5\\

        39 & TF6D (Default, CNOS) + Megapose~\cite{megapose} &  & \cellcolor{ccol!100}2024 & \cellcolor{ccol!100}CNOS-FastSAM & MegaPose & \cellcolor{ccol!100}RGB & \cellcolor{ccol!100}PBR & \cellcolor{ccol!100}RGB & \cellcolor{arcol!56.0}56.0 & \cellcolor{arcol!59.0}59.0 & \cellcolor{arcol!66.9}66.9 & \cellcolor{arcol!45.7}45.7 & \cellcolor{arcol!37.5}37.5 & \cellcolor{arcol!70.1}70.1 & \cellcolor{arcol!66.5}66.5 & \cellcolor{avgcol!57.4}57.4 & \cellcolor{timecol!4.1}$\phantom{0}$$\phantom{0}$4.1\\
        
        40 & ZeroPose-Multi-Hypo-Refinement~\cite{chen20233d,nguyen2023cnos} &  & 2023 & FastSAM+ImBind & MegaPose & RGB-D & \cellcolor{ccol!100}PBR & RGB-D &   \cellcolor{arcol!53.8}53.8 & \cellcolor{arcol!40.0}40.0 & \cellcolor{arcol!83.5}83.5 & \cellcolor{arcol!39.2}39.2 & \cellcolor{arcol!52.1}52.1 & \cellcolor{arcol!65.3}65.3 & \cellcolor{arcol!65.3}65.3 & 
        \cellcolor{avgcol!57.0}57.0 &
        \cellcolor{timecol!16.168}$\phantom{0}$16.2 \\
        
        41 & GenFlow-MultiHypo-RGB~\cite{genflow} &  & 2023 & \cellcolor{ccol!100}CNOS-FastSAM & GenFlow & RGB-D & \cellcolor{ccol!100}PBR & \cellcolor{ccol!100}RGB &   \cellcolor{arcol!56.3}56.3 & \cellcolor{arcol!52.3}52.3 & \cellcolor{arcol!68.4}68.4 & \cellcolor{arcol!45.3}45.3 & \cellcolor{arcol!39.5}39.5 & \cellcolor{arcol!73.9}73.9 & \cellcolor{arcol!63.3}63.3 & 
        \cellcolor{avgcol!57.0}57.0 & \cellcolor{timecol!20.890}$\phantom{0}$20.9 \\

        42 & GigaPose+GenFlow (1 hypo)~\cite{gigaPose, genflow} &  & \cellcolor{ccol!100}2024 & \cellcolor{ccol!100}CNOS-FastSAM & GenFlow & RGB-D & \cellcolor{ccol!100}PBR & \cellcolor{ccol!100}RGB & \cellcolor{arcol!59.5}59.5 & \cellcolor{arcol!55.0}55.0 & \cellcolor{arcol!60.7}60.7 & \cellcolor{arcol!47.8}47.8 & \cellcolor{arcol!41.3}41.3 & \cellcolor{arcol!72.2}72.2 & \cellcolor{arcol!60.8}60.8 & \cellcolor{avgcol!56.8}56.8 & \cellcolor{timecol!2.2}$\phantom{0}$$\phantom{0}$2.2\\

        43 & GenFlow~\cite{genflow} &  & \cellcolor{ccol!100}2024 & \cellcolor{ccol!100}CNOS-FastSAM & GenFlow & RGB-D & \cellcolor{ccol!100}PBR & \cellcolor{ccol!100}RGB & \cellcolor{arcol!54.7}54.7 & \cellcolor{arcol!51.4}51.4 & \cellcolor{arcol!67.0}67.0 & \cellcolor{arcol!43.7}43.7 & \cellcolor{arcol!38.4}38.4 & \cellcolor{arcol!73.0}73.0 & \cellcolor{arcol!61.9}61.9 & \cellcolor{avgcol!55.7}55.7 & \cellcolor{timecol!10.6}$\phantom{0}$10.6\\

        44 & FoundPose+FeatRef+Megapose~\cite{foundPose,megapose} &  & \cellcolor{ccol!100}2024 & \cellcolor{ccol!100}CNOS-FastSAM & MegaPose+FeatRef & \cellcolor{ccol!100}RGB & \cellcolor{ccol!100}PBR & \cellcolor{ccol!100}RGB & \cellcolor{arcol!55.6}55.6 & \cellcolor{arcol!51.1}51.1 & \cellcolor{arcol!63.3}63.3 & \cellcolor{arcol!40.0}40.0 & \cellcolor{arcol!35.7}35.7 & \cellcolor{arcol!69.7}69.7 & \cellcolor{arcol!66.1}66.1 & \cellcolor{avgcol!55.0}55.0 & \cellcolor{timecol!6.4}$\phantom{0}$6.4\\
        
        45 & Megapose-CNOS\_fastSAM+Multih-10~\cite{megapose} &  & 2023 & \cellcolor{ccol!100}CNOS-FastSAM & MegaPose & \cellcolor{ccol!100}RGB & \cellcolor{ccol!100}PBR & \cellcolor{ccol!100}RGB &    \cellcolor{arcol!56.0}56.0 & \cellcolor{arcol!50.8}50.8 & \cellcolor{arcol!68.7}68.7 & \cellcolor{arcol!41.9}41.9 & \cellcolor{arcol!34.6}34.6 & \cellcolor{arcol!70.6}70.6 & \cellcolor{arcol!62.0}62.0 & 
        \cellcolor{avgcol!54.9}54.9 &
        \cellcolor{timecol!53.878}$\phantom{0}$53.9 \\

        46 & FoundPose+MegaPose~\cite{foundPose,megapose}  &  & \cellcolor{ccol!100}2024 & \cellcolor{ccol!100}CNOS-FastSAM & MegaPose & \cellcolor{ccol!100}RGB & \cellcolor{ccol!100}PBR & \cellcolor{ccol!100}RGB & \cellcolor{arcol!55.4}55.4 & \cellcolor{arcol!51.0}51.0 & \cellcolor{arcol!63.3}63.3 & \cellcolor{arcol!43.0}43.0 & \cellcolor{arcol!34.6}34.6 & \cellcolor{arcol!69.5}69.5 & \cellcolor{arcol!66.1}66.1 & \cellcolor{avgcol!54.7}54.7 & \cellcolor{timecol!4.4}$\phantom{0}$$\phantom{0}$4.4\\

        47 & GigaPose+MegaPose~\cite{gigaPose, megapose} &  & \cellcolor{ccol!100}2024 & \cellcolor{ccol!100}CNOS-FastSAM & MegaPose & \cellcolor{ccol!100}RGB & \cellcolor{ccol!100}PBR & \cellcolor{ccol!100}RGB & \cellcolor{arcol!55.7}55.7 & \cellcolor{arcol!54.1}54.1 & \cellcolor{arcol!58.0}58.0 & \cellcolor{arcol!45.0}45.0 & \cellcolor{arcol!37.6}37.6 & \cellcolor{arcol!69.3}69.3 & \cellcolor{arcol!63.2}63.2 & \cellcolor{avgcol!54.7}54.7 & \cellcolor{timecol!2.3}$\phantom{0}$$\phantom{0}$2.3\\
        
        48 & Megapose-CNOS\_fastSAM+Multih~\cite{megapose} &  & 2023 & \cellcolor{ccol!100}CNOS-FastSAM & MegaPose & \cellcolor{ccol!100}RGB & \cellcolor{ccol!100}PBR & \cellcolor{ccol!100}RGB & \cellcolor{arcol!56.0}56.0 & \cellcolor{arcol!50.7}50.7 & \cellcolor{arcol!68.4}68.4 & \cellcolor{arcol!41.4}41.4 & \cellcolor{arcol!33.8}33.8 & \cellcolor{arcol!70.4}70.4 & \cellcolor{arcol!62.1}62.1 & 
        \cellcolor{avgcol!54.7}54.7 & 
        \cellcolor{timecol!47.386}$\phantom{0}$47.4 \\
        
        49 & ZeroPose-Multi-Hypo-Refinement~\cite{chen20233d} &  & 2023 & FastSAM+ImBind & MegaPose & RGB-D & PBR+Real & RGB-D  & \cellcolor{arcol!49.3}49.3 & \cellcolor{arcol!34.2}34.2 & \cellcolor{arcol!79.0}79.0 & \cellcolor{arcol!39.6}39.6 & \cellcolor{arcol!46.5}46.5 & \cellcolor{arcol!62.9}62.9 & \cellcolor{arcol!62.3}62.3 & 
        \cellcolor{avgcol!53.4}53.4 & 
        \cellcolor{timecol!18.971}$\phantom{0}$19.0 \\
        
        50 & MegaPose-CNOS\_fastSAM~\cite{megapose} &  & 2023 & \cellcolor{ccol!100}CNOS-FastSAM & MegaPose & \cellcolor{ccol!100}RGB & \cellcolor{ccol!100}PBR & \cellcolor{ccol!100}RGB & \cellcolor{arcol!49.9}49.9 & \cellcolor{arcol!47.7}47.7 & \cellcolor{arcol!65.3}65.3 & \cellcolor{arcol!36.7}36.7 & \cellcolor{arcol!31.5}31.5 & \cellcolor{arcol!65.4}65.4 & \cellcolor{arcol!60.1}60.1 & 
        \cellcolor{avgcol!50.9}50.9 & \cellcolor{timecol!31.724}$\phantom{0}$31.7 \\

        51 & OPFormer-Coarse (CNOS) &  & \cellcolor{ccol!100}2024 & \cellcolor{ccol!100}CNOS-FastSAM & - & - & - & \cellcolor{ccol!100}RGB & \cellcolor{arcol!52.5}52.5 & \cellcolor{arcol!41.8}41.8 & \cellcolor{arcol!61.5}61.5 & \cellcolor{arcol!34.2}34.2 & \cellcolor{arcol!27.8}27.8 & \cellcolor{arcol!67.3}67.3 & \cellcolor{arcol!60.6}60.6 & \cellcolor{avgcol!49.4}49.4 & \cellcolor{timecol!0.5}$\phantom{0}$$\phantom{0}$0.5\\

        52 & SMC-1.0s-CNOS &  & 2023 & \cellcolor{ccol!100}CNOS-FastSAM & - & - & - & D  & \cellcolor{arcol!55.8}55.8 & \cellcolor{arcol!42.3}42.3 & \cellcolor{arcol!59.9}59.9 & \cellcolor{arcol!31.6}31.6 & \cellcolor{arcol!38.9}38.9 & \cellcolor{arcol!58.5}58.5 & \cellcolor{arcol!45.4}45.4 & \cellcolor{avgcol!47.5}47.5 & \cellcolor{timecol!6.1}$\phantom{0}$$\phantom{0}$6.1\\

        53 & SMC-0.5s-CNOS &  & 2023 & \cellcolor{ccol!100}CNOS-FastSAM & - & - & - & D  & \cellcolor{arcol!51.2}51.2 & \cellcolor{arcol!41.5}41.5 & \cellcolor{arcol!51.1}51.1 & \cellcolor{arcol!29.0}29.0 & \cellcolor{arcol!35.8}35.8 & \cellcolor{arcol!53.8}53.8 & \cellcolor{arcol!40.3}40.3 & \cellcolor{avgcol!43.3}43.3 & \cellcolor{timecol!3.0}$\phantom{0}$$\phantom{0}$3.0\\

        54 & FoundPose+FeatRef~\cite{foundPose} & & \cellcolor{ccol!100}2024 & \cellcolor{ccol!100}CNOS-FastSAM & FeatRef & - & - & \cellcolor{ccol!100}RGB & \cellcolor{arcol!39.5}39.5 & \cellcolor{arcol!39.6}39.6 & \cellcolor{arcol!56.7}56.7 & \cellcolor{arcol!28.3}28.3 & \cellcolor{arcol!26.2}26.2 & \cellcolor{arcol!58.5}58.5 & \cellcolor{arcol!49.7}49.7 & \cellcolor{avgcol!42.6}42.6 & \cellcolor{timecol!2.6}$\phantom{0}$$\phantom{0}$2.6\\

        55 & TF6D (Default, CNOS) &  & \cellcolor{ccol!100}2024 & \cellcolor{ccol!100}CNOS-FastSAM & - & - & - & \cellcolor{ccol!100}RGB  & \cellcolor{arcol!32.3}32.3 & \cellcolor{arcol!35.0}35.0 & \cellcolor{arcol!47.3}47.3 & \cellcolor{arcol!33.2}33.2 & \cellcolor{arcol!25.1}25.1 & \cellcolor{arcol!53.7}53.7 & \cellcolor{arcol!54.1}54.1 & \cellcolor{avgcol!40.1}40.1 & \cellcolor{timecol!1.6}$\phantom{0}$$\phantom{0}$1.6\\

        56 & FoundPose-Coarse~\cite{foundPose} &  & \cellcolor{ccol!100}2024 & \cellcolor{ccol!100}CNOS-FastSAM & - & - & - & \cellcolor{ccol!100}RGB & \cellcolor{arcol!39.7}39.7 & \cellcolor{arcol!33.8}33.8 & \cellcolor{arcol!46.9}46.9 & \cellcolor{arcol!23.9}23.9 & \cellcolor{arcol!20.4}20.4 & \cellcolor{arcol!50.8}50.8 & \cellcolor{arcol!45.2}45.2 & \cellcolor{avgcol!37.3}37.3 & \cellcolor{timecol!1.7}$\phantom{0}$$\phantom{0}$1.7\\
        
        57 & ZeroPose-One-Hypo~\cite{chen20233d}&  & 2023 & FastSAM+ImBind & - & RGB-D & PBR+Real & RGB-D & \cellcolor{arcol!27.2}27.2 & \cellcolor{arcol!15.6}15.6 & \cellcolor{arcol!53.6}53.6 & \cellcolor{arcol!30.7}30.7 & \cellcolor{arcol!36.2}36.2 & \cellcolor{arcol!46.2}46.2 & \cellcolor{arcol!34.1}34.1 & 
        \cellcolor{avgcol!34.8}34.8 & \cellcolor{timecol!9.756}$\phantom{0}\phantom{0}$9.8 \\
        
        58 & GigaPose~\cite{gigaPose} &  & \cellcolor{ccol!100}2024 & \cellcolor{ccol!100}CNOS-FastSAM & - & \cellcolor{ccol!100}RGB & \cellcolor{ccol!100}PBR & \cellcolor{ccol!100}RGB & \cellcolor{arcol!29.6}29.6 & \cellcolor{arcol!26.4}26.4 & \cellcolor{arcol!30.0}30.0 & \cellcolor{arcol!22.3}22.3 & \cellcolor{arcol!17.5}17.5 & \cellcolor{arcol!34.1}34.1 & \cellcolor{arcol!27.8}27.8 & \cellcolor{avgcol!26.8}26.8 & \cellcolor{timecol!0.4}$\phantom{0}\phantom{0}$0.4\\
        
        59 & GenFlow-coarse~\cite{genflow}  &  & 2023 & \cellcolor{ccol!100}CNOS-FastSAM & - & RGB-D & \cellcolor{ccol!100}PBR & \cellcolor{ccol!100}RGB &  \cellcolor{arcol!25.0}25.0 & \cellcolor{arcol!21.5}21.5 & \cellcolor{arcol!30.0}30.0 & \cellcolor{arcol!16.8}16.8 & \cellcolor{arcol!15.4}15.4 & \cellcolor{arcol!28.3}28.3 & \cellcolor{arcol!27.7}27.7 & 
        \cellcolor{avgcol!23.5}23.5 & \cellcolor{timecol!3.839}$\phantom{0}\phantom{0}$3.8  \\

        60 & MegaPose-CNOS\_fastSAM+CoarseBest~\cite{megapose} &  & 2023 & \cellcolor{ccol!100}CNOS-FastSAM & - & \cellcolor{ccol!100}RGB & \cellcolor{ccol!100}PBR & \cellcolor{ccol!100}RGB &  \cellcolor{arcol!22.9}22.9 & \cellcolor{arcol!17.7}17.7 & \cellcolor{arcol!25.8}25.8 & \cellcolor{arcol!15.2}15.2 & \cellcolor{arcol!10.8}10.8 & \cellcolor{arcol!25.1}25.1 & \cellcolor{arcol!28.1}28.1 & \cellcolor{avgcol!20.8}20.8 & \cellcolor{timecol!15.5}$\phantom{0}$15.5\\
        
        \bottomrule
  
    \end{tabularx}
    \vspace{-3pt}
    \caption{\textbf{Track 1: Model-based 6D localization of unseen objects on BOP-Classic-Core.}
    Methods are ranked by the $\text{AR}$ score, which is the average of per-dataset $\text{AR}_d$ scores 
    (Sec.~\ref{sec:evaluation_methodology}).
    The last column shows the average time to generate predictions for all objects in a single image, averaged over the datasets (measured on different computers by the participants).
    Column \emph{Year} is the year of submission, \emph{Det./seg.}~is the object detection/segmentation method, \emph{Refinement} is the pose refinement method,
    \emph{Train im.}~and \emph{Test im.}~show image channels used at training and test time respectively, and \emph{Train im.~type} is the domain of training images. All test images are real. See Sec.~\ref{sec:awards} for description of the awards.
    }
    \label{tab:track1}
\end{table*}

\setlength{\tabcolsep}{3pt}
\begin{table*}%
    \renewcommand{\arraystretch}{0.95}
    \tiny
    \centering
    \begin{tabularx}{\linewidth}{rlcllllllYYYYYYYYY}
        \toprule
        \# & Method & Awards & Year & Det./seg. & Refinement & Train im. & ...type & Test image & LM-O & T-LESS & TUD-L & IC-BIN & ITODD & HB & YCB-V & \mbox{$\text{AP}$} & Time  \\ 
        \midrule
        1 & FreeZeV2.1~\cite{freeze}  & \iconBest & \cellcolor{ccol!100}2024& Custom & \cellcolor{ccol!100}ICP & - & - & RGB-D  &\cellcolor{arcol!79.7}79.7 & \cellcolor{arcol!75.1}75.1 & \cellcolor{arcol!99.1}99.1 & \cellcolor{arcol!69.6}69.6 & \cellcolor{arcol!76.9}76.9 & \cellcolor{arcol!85.3}85.3 & \cellcolor{arcol!90.5}90.5 & \cellcolor{avgcol!82.3}82.3 & \cellcolor{timecol!37.3}37.3 \\

        2 & FreeZeV2 (SAM6D)~\cite{freeze} & & \cellcolor{ccol!100}2024& SAM6D-FastSAM & \cellcolor{ccol!100}ICP & - & - & RGB-D & \cellcolor{arcol!77.5}77.5 & \cellcolor{arcol!61.0}61.0 & \cellcolor{arcol!97.5}97.5 & \cellcolor{arcol!62.0}62.0 & \cellcolor{arcol!61.7}61.7 & \cellcolor{arcol!78.2}78.2 & \cellcolor{arcol!86.9}86.9 & \cellcolor{avgcol!75.0}75.0 & \cellcolor{timecol!55.4}55.4 \\

        3 & FreeZeV2 (SAM6D, Coarse-to-Fine)~\cite{freeze} & & \cellcolor{ccol!100}2024& SAM6D-FastSAM & \cellcolor{ccol!100}ICP & - & - & RGB-D & \cellcolor{arcol!74.3}74.3 & \cellcolor{arcol!60.1}60.1 & \cellcolor{arcol!90.2}90.2 & \cellcolor{arcol!53.1}53.1 & \cellcolor{arcol!57.3}57.3 & \cellcolor{arcol!74.1}74.1 & \cellcolor{arcol!85.8}85.8 & \cellcolor{avgcol!70.7}70.7 & \cellcolor{timecol!12.9}12.9 \\

        4 & Co-op (F3DT2D, Coarse, RGBD)~\cite{Moon2025Coop}  & \iconFast & \cellcolor{ccol!100}2024& F3DT2D & - & RGB-D & \cellcolor{ccol!100}PBR & RGB-D & \cellcolor{arcol!69.8}69.8 & \cellcolor{arcol!62.0}62.0 & \cellcolor{arcol!84.1}84.1 & \cellcolor{arcol!56.4}56.4 & \cellcolor{arcol!57.6}57.6 & \cellcolor{arcol!74.6}74.6 & \cellcolor{arcol!80.8}80.8 & \cellcolor{avgcol!69.3}69.3 & \cellcolor{timecol!0.9}$\phantom{0}$0.9 \\

        5 & Co-op (F3DT2D, 5 Hypo, RGBD)~\cite{Moon2025Coop} & & \cellcolor{ccol!100}2024& F3DT2D & Co-op & RGB-D & \cellcolor{ccol!100}PBR & RGB-D & \cellcolor{arcol!70.1}70.1 & \cellcolor{arcol!61.3}61.3 & \cellcolor{arcol!76.6}76.6 & \cellcolor{arcol!42.6}42.6 & \cellcolor{arcol!62.7}62.7 & \cellcolor{arcol!73.4}73.4 & \cellcolor{arcol!81.2}81.2 & \cellcolor{avgcol!66.9}66.9 & \cellcolor{timecol!12.2}12.2 \\

        6 & Co-op (CNOS, Coarse, RGBD)~\cite{Moon2025Coop}  & \iconDefault & \cellcolor{ccol!100}2024& \cellcolor{ccol!100}CNOS-FastSAM & - & RGB-D & \cellcolor{ccol!100}PBR & RGB-D & \cellcolor{arcol!68.3}68.3 & \cellcolor{arcol!59.6}59.6 & \cellcolor{arcol!80.8}80.8 & \cellcolor{arcol!46.9}46.9 & \cellcolor{arcol!56.0}56.0 & \cellcolor{arcol!74.3}74.3 & \cellcolor{arcol!78.2}78.2 & \cellcolor{avgcol!66.3}66.3 & \cellcolor{timecol!2.2}$\phantom{0}$2.2 \\

        7 & Co-op (CNOS, 1 Hypo, RGBD)~\cite{Moon2025Coop} & & \cellcolor{ccol!100}2024& \cellcolor{ccol!100}CNOS-FastSAM & Co-op & RGB-D & \cellcolor{ccol!100}PBR & RGB-D & \cellcolor{arcol!67.0}67.0 & \cellcolor{arcol!58.3}58.3 & \cellcolor{arcol!76.9}76.9 & \cellcolor{arcol!45.8}45.8 & \cellcolor{arcol!57.5}57.5 & \cellcolor{arcol!73.7}73.7 & \cellcolor{arcol!76.6}76.6 & \cellcolor{avgcol!65.1}65.1 & \cellcolor{timecol!6.9}$\phantom{0}$6.9 \\

        8 & Co-op (CNOS, 5 Hypo, RGBD)~\cite{Moon2025Coop} & & \cellcolor{ccol!100}2024& \cellcolor{ccol!100}CNOS-FastSAM & Co-op & RGB-D & \cellcolor{ccol!100}PBR & RGB-D & \cellcolor{arcol!68.2}68.2 & \cellcolor{arcol!58.9}58.9 & \cellcolor{arcol!73.8}73.8 & \cellcolor{arcol!39.9}39.9 & \cellcolor{arcol!60.4}60.4 & \cellcolor{arcol!73.1}73.1 & \cellcolor{arcol!76.6}76.6 & \cellcolor{avgcol!64.4}64.4 & \cellcolor{timecol!14.3}14.3 \\

        9 & Co-op (F3DT2D, 5 Hypo)~\cite{Moon2025Coop}  & \iconRGB & \cellcolor{ccol!100}2024& F3DT2D & Co-op & RGB-D & \cellcolor{ccol!100}PBR & \cellcolor{ccol!100}RGB &\cellcolor{arcol!63.7}63.7 & \cellcolor{arcol!61.6}61.6 & \cellcolor{arcol!65.8}65.8 & \cellcolor{arcol!46.7}46.7 & \cellcolor{arcol!50.4}50.4 & \cellcolor{arcol!73.2}73.2 & \cellcolor{arcol!62.6}62.6 & \cellcolor{avgcol!60.6}60.6 & \cellcolor{timecol!8.7}$\phantom{0}$8.7 \\

        10 & Co-op (CNOS, 1 Hypo)~\cite{Moon2025Coop} & & \cellcolor{ccol!100}2024& \cellcolor{ccol!100}CNOS-FastSAM & Co-op & RGB-D & \cellcolor{ccol!100}PBR & \cellcolor{ccol!100}RGB &\cellcolor{arcol!61.5}61.5 & \cellcolor{arcol!58.8}58.8 & \cellcolor{arcol!64.1}64.1 & \cellcolor{arcol!40.9}40.9 & \cellcolor{arcol!46.5}46.5 & \cellcolor{arcol!72.7}72.7 & \cellcolor{arcol!59.2}59.2 & \cellcolor{avgcol!57.7}57.7 & \cellcolor{timecol!6.4}$\phantom{0}$6.4 \\

        11 & GigaPose+GenFlow (5 hypothesis)~\cite{gigaPose,genflow} & & \cellcolor{ccol!100}2024& \cellcolor{ccol!100}CNOS-FastSAM & GenFlow & RGB-D & \cellcolor{ccol!100}PBR & \cellcolor{ccol!100}RGB &\cellcolor{arcol!59.7}59.7 & \cellcolor{arcol!56.5}56.5 & \cellcolor{arcol!68.8}68.8 & \cellcolor{arcol!43.9}43.9 & \cellcolor{arcol!42.5}42.5 & \cellcolor{arcol!70.7}70.7 & \cellcolor{arcol!60.7}60.7 & \cellcolor{avgcol!57.5}57.5 & \cellcolor{timecol!15.5}15.5 \\

        12 & Co-op (CNOS, 5 Hypo)~\cite{Moon2025Coop} & & \cellcolor{ccol!100}2024& \cellcolor{ccol!100}CNOS-FastSAM & Co-op & RGB-D & \cellcolor{ccol!100}PBR & \cellcolor{ccol!100}RGB &\cellcolor{arcol!61.2}61.2 & \cellcolor{arcol!59.0}59.0 & \cellcolor{arcol!61.9}61.9 & \cellcolor{arcol!39.1}39.1 & \cellcolor{arcol!48.3}48.3 & \cellcolor{arcol!72.3}72.3 & \cellcolor{arcol!59.1}59.1 & \cellcolor{avgcol!57.3}57.3 & \cellcolor{timecol!11.5}11.5 \\

        13 & GigaPose+GenFlow (RGBD)~\cite{gigaPose,genflow} & & \cellcolor{ccol!100}2024& \cellcolor{ccol!100}CNOS-FastSAM & GenFlow & RGB-D & \cellcolor{ccol!100}PBR & RGB-D & \cellcolor{arcol!57.8}57.8 & \cellcolor{arcol!46.7}46.7 & \cellcolor{arcol!71.5}71.5 & \cellcolor{arcol!40.2}40.2 & \cellcolor{arcol!44.7}44.7 & \cellcolor{arcol!67.6}67.6 & \cellcolor{arcol!68.2}68.2 & \cellcolor{avgcol!56.7}56.7 & \cellcolor{timecol!4.5}$\phantom{0}$4.5 \\

        14 & Co-op (CNOS, Coarse)~\cite{Moon2025Coop} & & \cellcolor{ccol!100}2024& \cellcolor{ccol!100}CNOS-FastSAM & - & RGB-D & \cellcolor{ccol!100}PBR & \cellcolor{ccol!100}RGB &\cellcolor{arcol!58.9}58.9 & \cellcolor{arcol!55.8}55.8 & \cellcolor{arcol!64.0}64.0 & \cellcolor{arcol!40.3}40.3 & \cellcolor{arcol!38.8}38.8 & \cellcolor{arcol!71.1}71.1 & \cellcolor{arcol!57.6}57.6 & \cellcolor{avgcol!55.2}55.2 & \cellcolor{timecol!2.2}$\phantom{0}$2.2 \\

        15 & GigaPose+GenFlow ~\cite{gigaPose,genflow} & & \cellcolor{ccol!100}2024& \cellcolor{ccol!100}CNOS-FastSAM & GenFlow & RGB-D & \cellcolor{ccol!100}PBR & \cellcolor{ccol!100}RGB &\cellcolor{arcol!55.4}55.4 & \cellcolor{arcol!43.6}43.6 & \cellcolor{arcol!60.8}60.8 & \cellcolor{arcol!35.3}35.3 & \cellcolor{arcol!36.7}36.7 & \cellcolor{arcol!66.8}66.8 & \cellcolor{arcol!54.6}54.6 & \cellcolor{avgcol!50.4}50.4 & \cellcolor{timecol!4.7}$\phantom{0}$4.7 \\

        16 & GigaPose+GenFlow+kabsch (5 hypothesis)~\cite{gigaPose,genflow} & & \cellcolor{ccol!100}2024& \cellcolor{ccol!100}CNOS-FastSAM & GenFlow & RGB-D & \cellcolor{ccol!100}PBR & RGB-D & \cellcolor{arcol!23.7}23.7 & \cellcolor{arcol!52.6}52.6 & \cellcolor{arcol!24.6}24.6 & \cellcolor{arcol!29.2}29.2 & \cellcolor{arcol!52.5}52.5 & \cellcolor{arcol!54.7}54.7 & \cellcolor{arcol!52.4}52.4 & \cellcolor{avgcol!41.4}41.4 & \cellcolor{timecol!16.6}16.6 \\

        17 & GigaPose-CVPR24~\cite{gigaPose}  & \iconOpen & \cellcolor{ccol!100}2024& \cellcolor{ccol!100}CNOS-FastSAM & - & \cellcolor{ccol!100}RGB &\cellcolor{ccol!100}PBR & \cellcolor{ccol!100}RGB &\cellcolor{arcol!6.2}$\phantom{0}$6.2 & \cellcolor{arcol!25.5}25.5 & \cellcolor{arcol!3.5}$\phantom{0}$3.5 & \cellcolor{arcol!5.7}$\phantom{0}$5.7 & \cellcolor{arcol!14.9}14.9 & \cellcolor{arcol!18.2}18.2 & \cellcolor{arcol!12.5}12.5 & \cellcolor{avgcol!12.3}12.3 & \cellcolor{timecol!0.7}$\phantom{0}$0.7 \\

        \bottomrule
  
    \end{tabularx}
    \vspace{-3pt}
    \caption{\textbf{Track 2: Model-based 6D detection of unseen objects on BOP-Classic-Core.} Methods are ranked by $\text{AP}$ (Sec.~\ref{sec:evaluation_methodology}).
    Columns as in Tab.~\ref{tab:track1}. 
    }
    \label{tab:track2}
    
\end{table*}

\setlength{\tabcolsep}{2pt}
\begin{table*}[t!]
    \renewcommand{\arraystretch}{0.95}
    \tiny
    \centering
    \begin{tabularx}{\linewidth}{rlcllllYYYYYYYYY}
        \toprule
        \# & Method & Awards & Year & Onboarding im. & ...type & Test image & LM-O & T-LESS & TUD-L & IC-BIN & ITODD & HB & YCB-V & \mbox{$\text{AP}$} & Time  \\
        \midrule
        1 & MUSE  & \iconBest \; \iconFast \; \iconRGB & \cellcolor{ccol!100}2024 & - & - & \cellcolor{ccol!100}RGB & \cellcolor{arcol!51.2}51.2 & \cellcolor{arcol!46.7}46.7 & \cellcolor{arcol!59.5}59.5 & \cellcolor{arcol!29.8}29.8 & \cellcolor{arcol!50.2}50.2 & \cellcolor{arcol!58.9}58.9 & \cellcolor{arcol!67.4}67.4 & \cellcolor{avgcol!52.0}52.0 & \cellcolor{timecol!0.56}0.56 \\

        2 & F3DT2D &  & \cellcolor{ccol!100}2024 & - & -& \cellcolor{ccol!100}RGB & \cellcolor{arcol!50.4}50.4 & \cellcolor{arcol!48.2}48.2 & \cellcolor{arcol!57.3}57.3 & \cellcolor{arcol!28.4}28.4 & \cellcolor{arcol!48.0}48.0 & \cellcolor{arcol!57.7}57.7 & \cellcolor{arcol!66.6}66.6 & \cellcolor{avgcol!50.9}50.9 & \cellcolor{timecol!0.43}0.43 \\

        3 & SAM6D-FastSAM~\cite{lin2023sam}  &  \iconOpen & 2023 & RGB-D & \cellcolor{ccol!100}PBR & RGB-D & \cellcolor{arcol!46.3}46.3 & \cellcolor{arcol!45.8}45.8 & \cellcolor{arcol!57.3}57.3 & \cellcolor{arcol!24.5}24.5 & \cellcolor{arcol!41.9}41.9 & \cellcolor{arcol!55.1}55.1 & \cellcolor{arcol!58.9}58.9 & \cellcolor{avgcol!47.1}47.1 & \cellcolor{timecol!0.45}0.45 \\
        
        4 & NIDS-Net\_WA\_Sappe~\cite{lu2024adapting} &  & \cellcolor{ccol!100}2024 & \cellcolor{ccol!100}RGB &  Custom & \cellcolor{ccol!100}RGB & \cellcolor{arcol!45.7}45.7 & \cellcolor{arcol!49.3}49.3 & \cellcolor{arcol!48.6}48.6 & \cellcolor{arcol!25.7}25.7 & \cellcolor{arcol!37.9}37.9 & \cellcolor{arcol!58.7}58.7 & \cellcolor{arcol!62.1}62.1 & \cellcolor{avgcol!46.9}46.9 & \cellcolor{timecol!0.49}0.49 \\
        
        5 & NIDS-Net\_WA~\cite{lu2024adapting} &  & \cellcolor{ccol!100}2024 & \cellcolor{ccol!100}RGB &  Custom &  \cellcolor{ccol!100}RGB & \cellcolor{arcol!44.9}44.9 & \cellcolor{arcol!48.9}48.9 & \cellcolor{arcol!46.0}46.0 & \cellcolor{arcol!24.5}24.5 & \cellcolor{arcol!36.0}36.0 & \cellcolor{arcol!59.4}59.4 & \cellcolor{arcol!62.4}62.4 & \cellcolor{avgcol!46.0}46.0 & \cellcolor{timecol!0.49}0.49 \\

        6 & SAM6D~\cite{lin2023sam} &  & \cellcolor{ccol!100}2024 & RGB-D & \cellcolor{ccol!100}PBR & RGB-D & \cellcolor{arcol!46.5}46.5 & \cellcolor{arcol!43.7}43.7 & \cellcolor{arcol!53.7}53.7 & \cellcolor{arcol!26.1}26.1 & \cellcolor{arcol!39.4}39.4 & \cellcolor{arcol!53.0}53.0 & \cellcolor{arcol!51.8}51.8 & \cellcolor{avgcol!44.9}44.9 & \cellcolor{timecol!2.80}2.80 \\
        
        7 & SAM6D-FastSAM~\cite{lin2023sam}  &  & \cellcolor{ccol!100}2024 & RGB-D & \cellcolor{ccol!100}PBR & \cellcolor{ccol!100}RGB & \cellcolor{arcol!43.8}43.8 & \cellcolor{arcol!41.7}41.7 & \cellcolor{arcol!54.6}54.6 & \cellcolor{arcol!23.4}23.4 & \cellcolor{arcol!37.4}37.4 & \cellcolor{arcol!52.3}52.3 & \cellcolor{arcol!57.3}57.3 & \cellcolor{avgcol!44.4}44.4 & \cellcolor{timecol!0.25}0.25 \\
        
        8 & ViewInvDet & & \cellcolor{ccol!100}2024 & - & -& \cellcolor{ccol!100}RGB & \cellcolor{arcol!44.9}44.9 & \cellcolor{arcol!40.3}40.3 & \cellcolor{arcol!50.8}50.8 & \cellcolor{arcol!26.8}26.8 & \cellcolor{arcol!32.8}32.8 & \cellcolor{arcol!55.4}55.4 & \cellcolor{arcol!58.1}58.1 & \cellcolor{avgcol!44.2}44.2 & \cellcolor{timecol!1.70}1.70 \\
        
        9 & NIDS-Net\_basic~\cite{lu2024adapting}  & & \cellcolor{ccol!100}2024 & RGB-D & \cellcolor{ccol!100}PBR & \cellcolor{ccol!100}RGB & \cellcolor{arcol!44.9}44.9 & \cellcolor{arcol!42.8}42.8 & \cellcolor{arcol!43.4}43.4 & \cellcolor{arcol!24.4}24.4 & \cellcolor{arcol!34.9}34.9 & \cellcolor{arcol!54.8}54.8 & \cellcolor{arcol!56.5}56.5 & \cellcolor{avgcol!43.1}43.1 & \cellcolor{timecol!0.49}0.49 \\
        
        10 & CNOS (FastSAM)~\cite{nguyen2023cnos} & & 2023 & - & - & \cellcolor{ccol!100}RGB & \cellcolor{arcol!43.3}43.3 & \cellcolor{arcol!39.5}39.5 & \cellcolor{arcol!53.4}53.4 & \cellcolor{arcol!22.6}22.6 & \cellcolor{arcol!32.5}32.5 & \cellcolor{arcol!51.7}51.7 & \cellcolor{arcol!56.8}56.8 & \cellcolor{avgcol!42.8}42.8 & \cellcolor{timecol!0.22}0.22 \\
        
        11 & CNOS (SAM)~\cite{nguyen2023cnos} & & 2023 & - & - & \cellcolor{ccol!100}RGB & \cellcolor{arcol!39.5}39.5 & \cellcolor{arcol!33.0}33.0 & \cellcolor{arcol!36.8}36.8 & \cellcolor{arcol!20.7}20.7 & \cellcolor{arcol!31.3}31.3 & \cellcolor{arcol!42.3}42.3 & \cellcolor{arcol!49.0}49.0 & \cellcolor{avgcol!36.1}36.1 & \cellcolor{timecol!1.85}1.85 \\
        
        12 & ZeroPose~\cite{chen20233d} &  &2023 & - & - & \cellcolor{ccol!100}RGB & \cellcolor{arcol!36.7}36.7 & \cellcolor{arcol!30.0}30.0 & \cellcolor{arcol!43.1}43.1 & \cellcolor{arcol!22.8}22.8 & \cellcolor{arcol!25.0}25.0 & \cellcolor{arcol!39.8}39.8 & \cellcolor{arcol!41.6}41.6 & \cellcolor{avgcol!34.1}34.1 & \cellcolor{timecol!3.82}3.82 \\
        \bottomrule
    \end{tabularx}
    \vspace{-3pt}
    \caption{\textbf{Track 3: Model-based 2D detection of unseen objects on BOP-Classic-Core.} The methods are ranked by $\text{AP}$ (Sec.~\ref{sec:evaluation_methodology}). Columns as in Tab.~\ref{tab:track1}.
    } 
    \label{tab:track3}
\end{table*}

Participants were submitting results to the online evaluation system
from May 29th until November 29th, 2024. We received submissions for all but Track 6, presumably due to the limited time (BOP-H3 datasets were released later) and the extra effort required to prepare model-free methods.
In this section we aim to provide a high-level summary of the results and refer the reader to the online system for more per-submission details. Evaluation scripts used by the system are publicly available in the BOP toolkit.

\begin{figure}
    \vspace{-1.0ex}
    \centering
    \footnotesize
    \setlength{\tabcolsep}{1pt} %
    \renewcommand{\arraystretch}{0.6} %
    \begin{tabular}{cccc}
        \raisebox{16pt}{\rotatebox[origin=l]{90}{LM-O}}\hspace{1pt} &
        \includegraphics[width=0.309\linewidth]{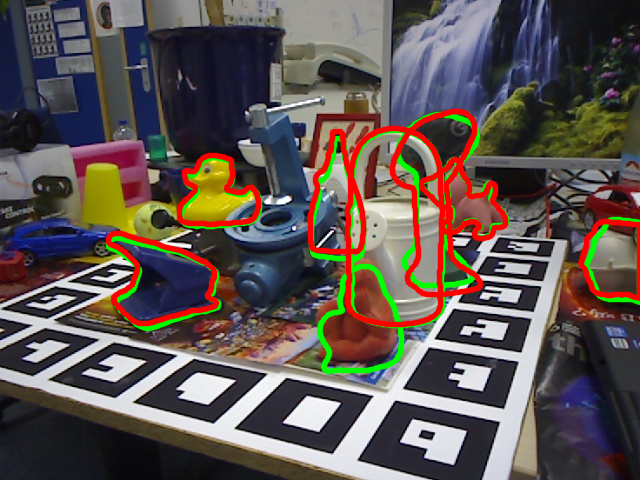} &
        \includegraphics[width=0.309\linewidth]{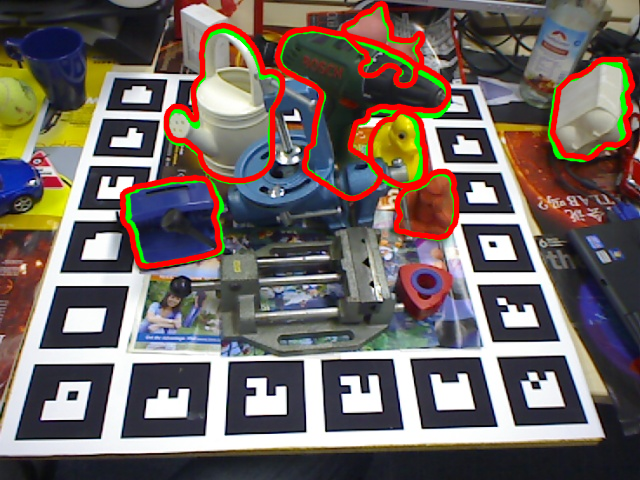} &
        \includegraphics[width=0.309\linewidth]{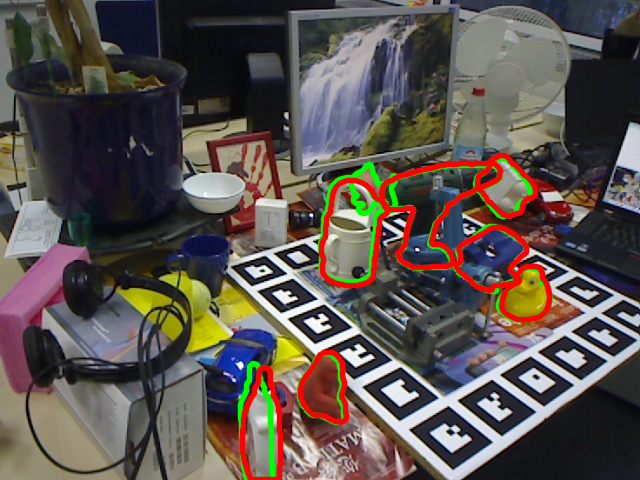} \\

        \raisebox{14pt}{\rotatebox[origin=l]{90}{T-LESS}}\hspace{1pt} &
        \includegraphics[width=0.309\linewidth]{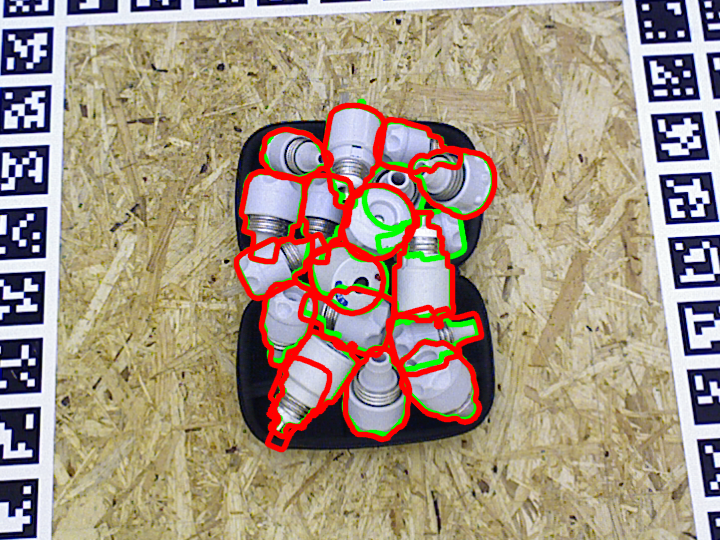} &
        \includegraphics[width=0.309\linewidth]{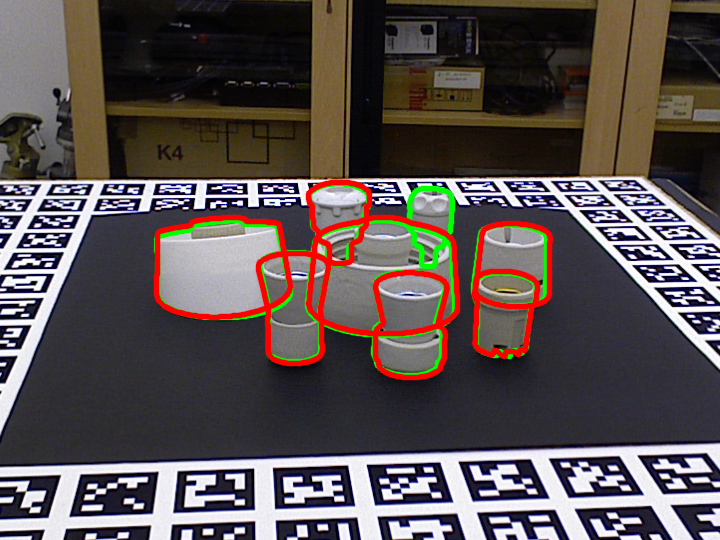} &
        \includegraphics[width=0.309\linewidth]{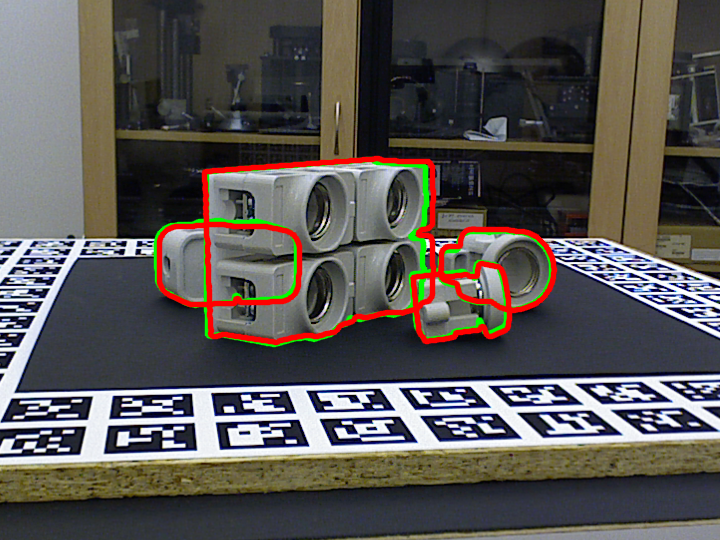} \\

        \raisebox{15pt}{\rotatebox[origin=l]{90}{TUD-L}}\hspace{1pt} &
        \includegraphics[width=0.309\linewidth]{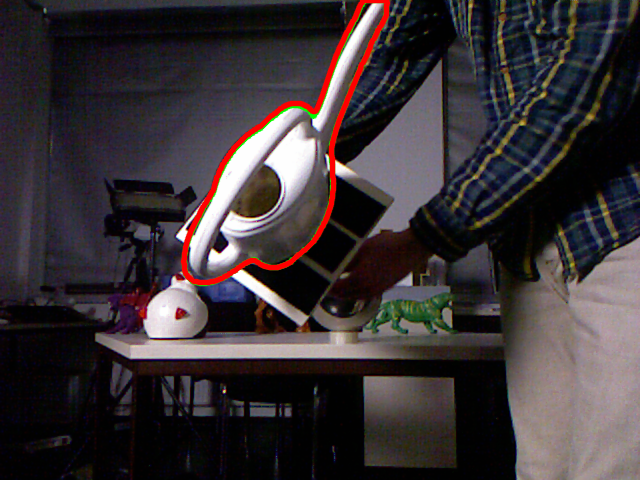} &
        \includegraphics[width=0.309\linewidth]{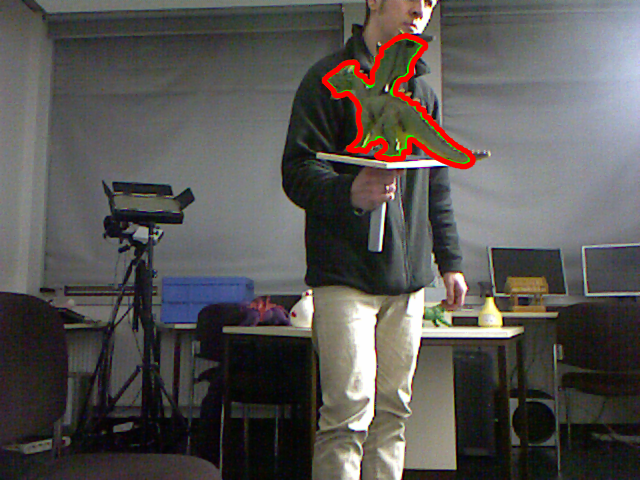} &
        \includegraphics[width=0.309\linewidth]{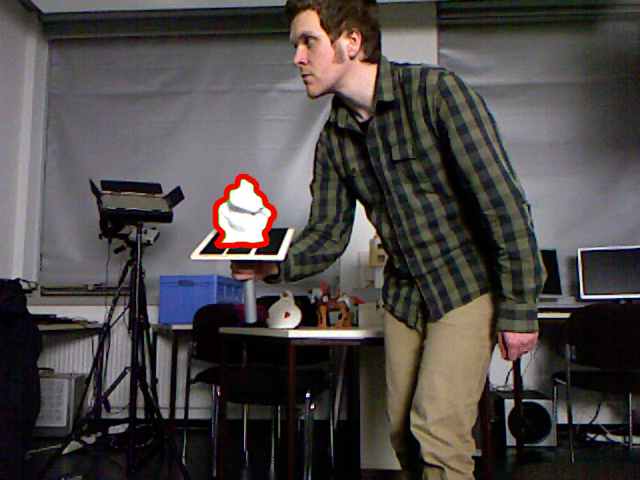} \\

        \raisebox{15pt}{\rotatebox[origin=l]{90}{IC-BIN}}\hspace{1pt} &
        \includegraphics[width=0.309\linewidth]{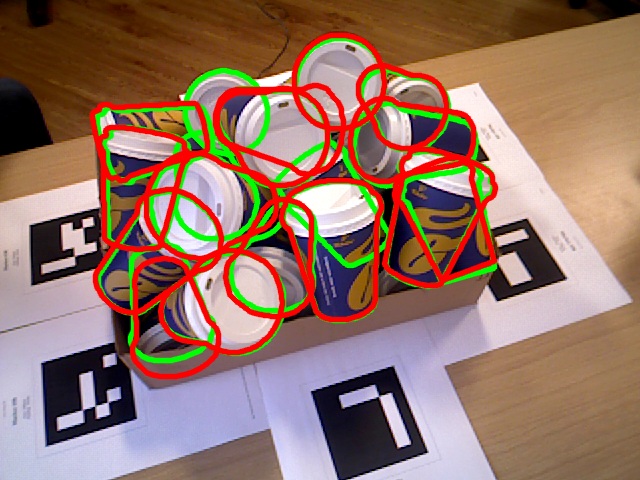} &
        \includegraphics[width=0.309\linewidth]{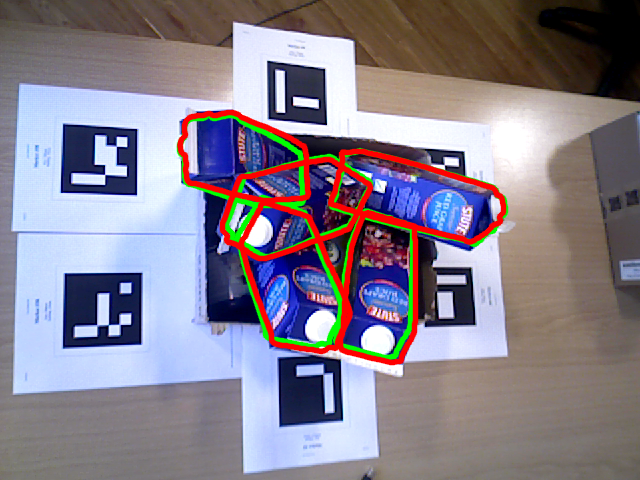} &
        \includegraphics[width=0.309\linewidth]{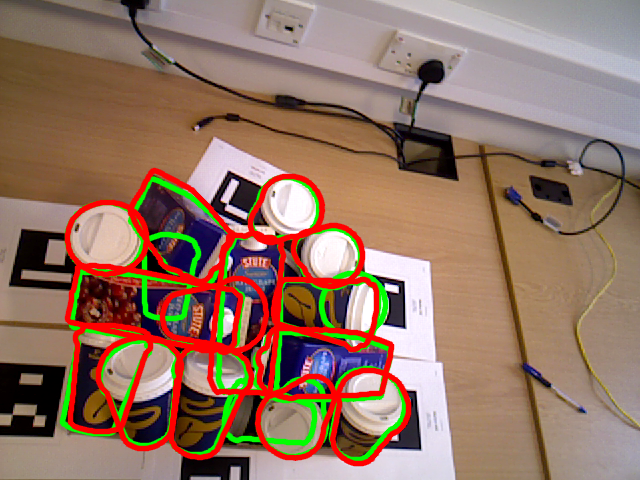} \\

        \raisebox{14pt}{\rotatebox[origin=l]{90}{ITODD}}\hspace{1pt} &
        \includegraphics[width=0.309\linewidth]{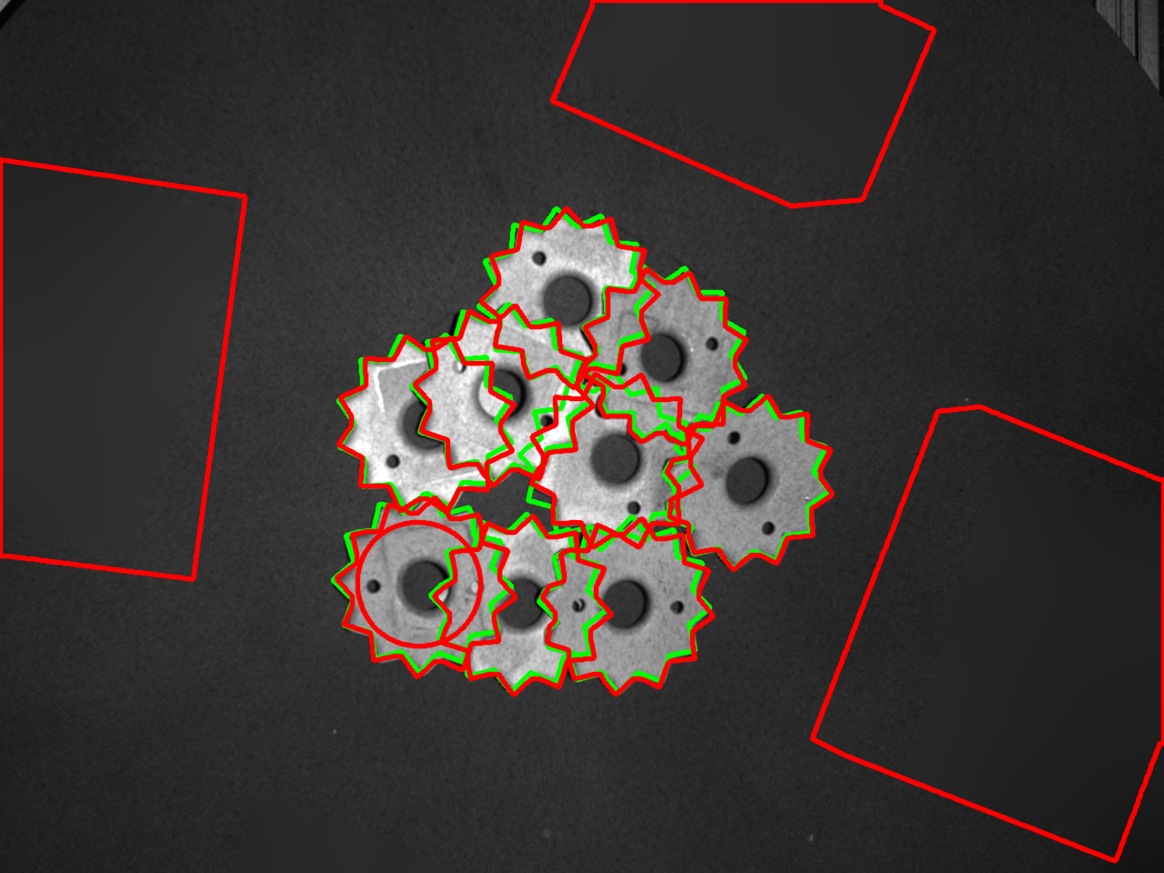} &
        \includegraphics[width=0.309\linewidth]{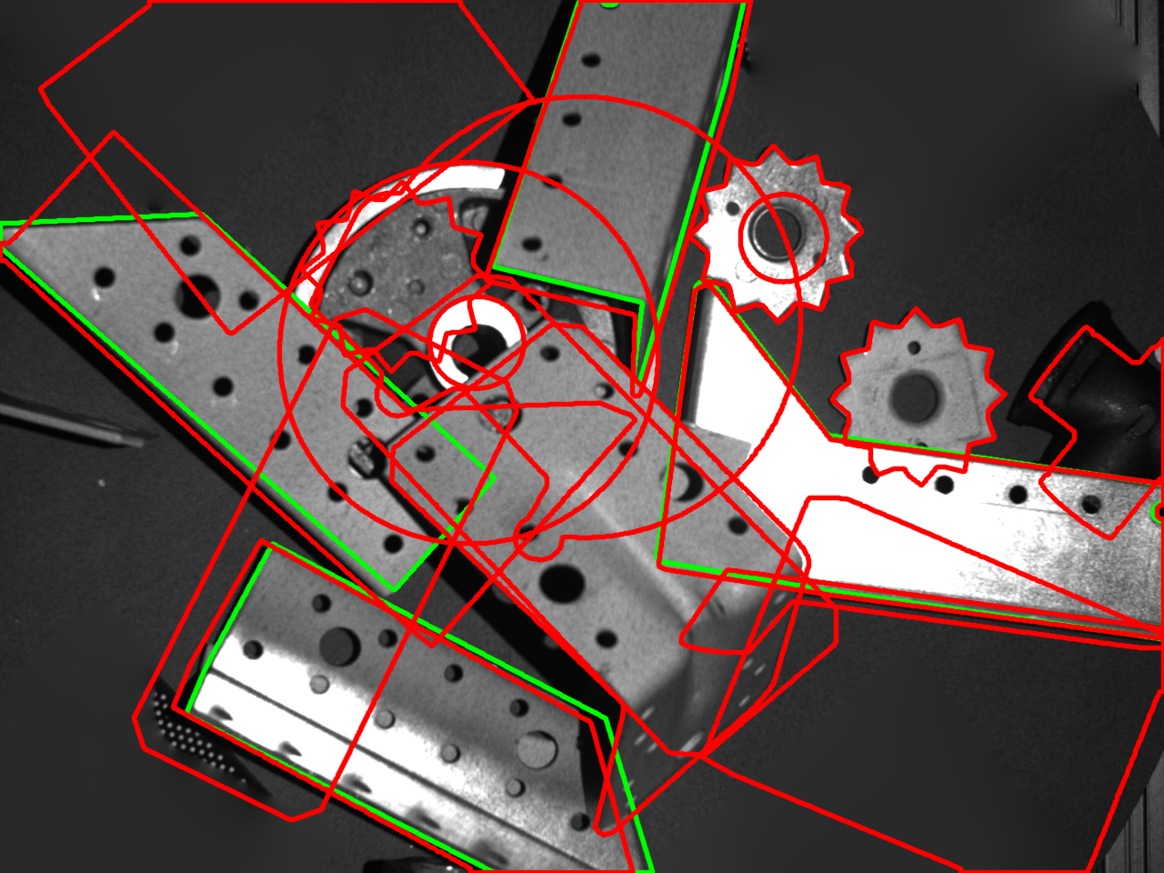} &
        \includegraphics[width=0.309\linewidth]{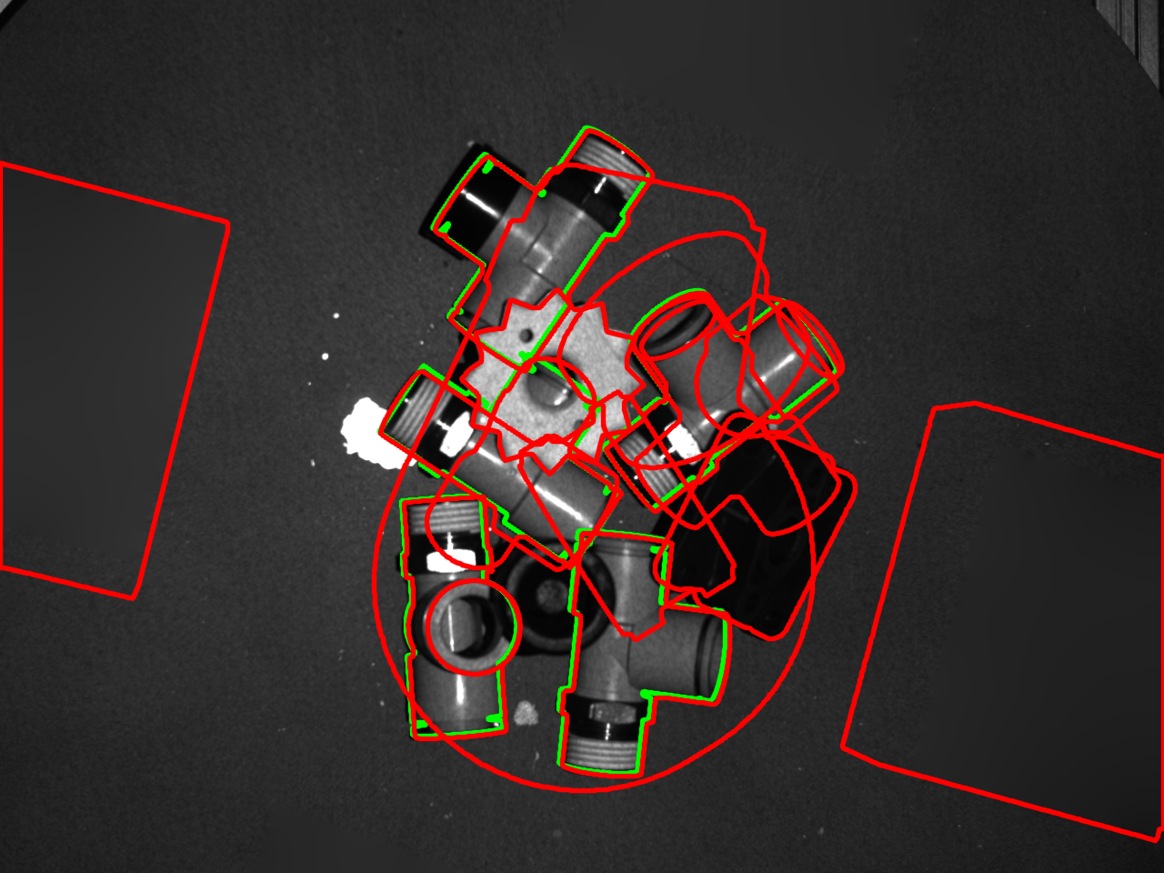} \\

        \raisebox{22pt}{\rotatebox[origin=l]{90}{HB}}\hspace{1pt} &
        \includegraphics[width=0.309\linewidth]{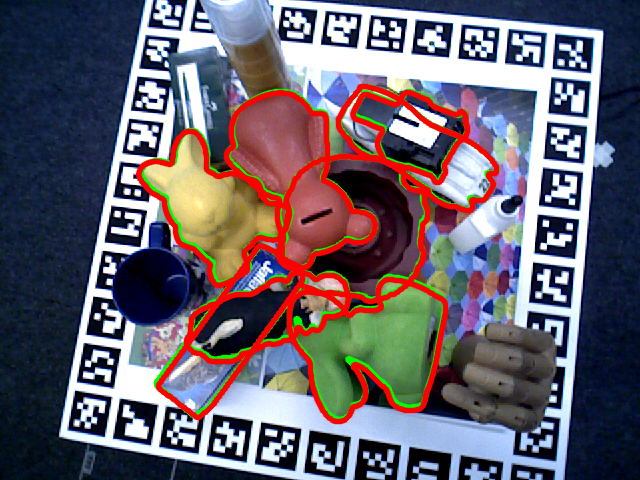} &
        \includegraphics[width=0.309\linewidth]{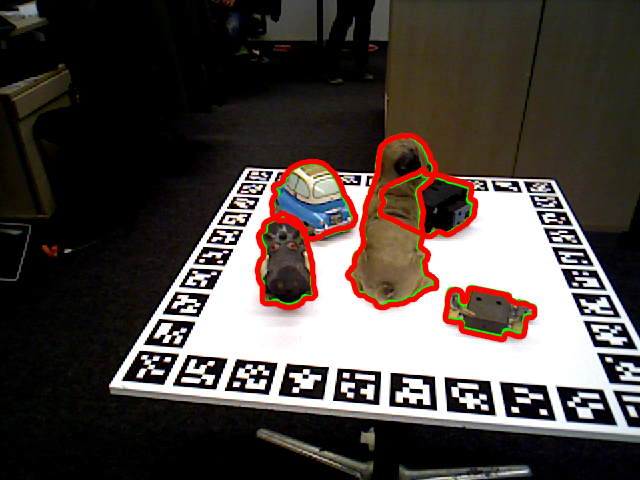} &
        \includegraphics[width=0.309\linewidth]{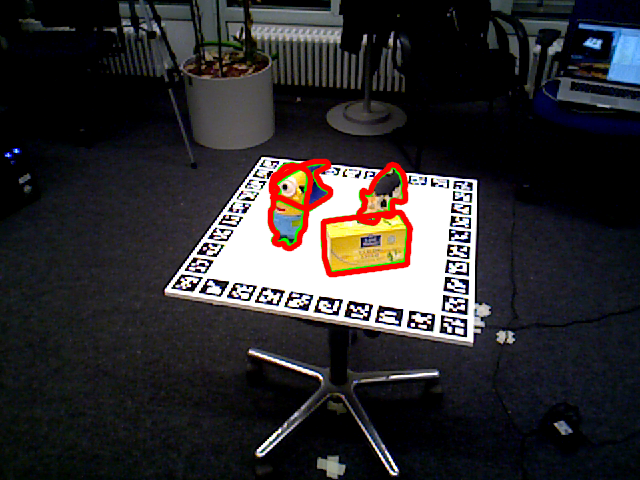} \\

        \raisebox{14pt}{\rotatebox[origin=l]{90}{YCB-V}}\hspace{1pt} &
        \includegraphics[width=0.309\linewidth]{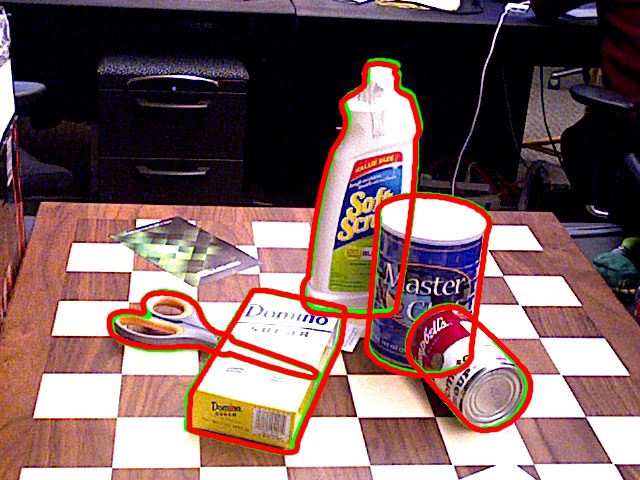} &
        \includegraphics[width=0.309\linewidth]{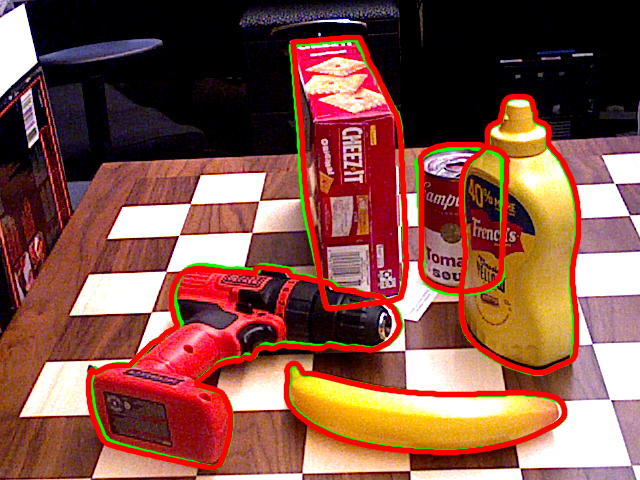} &
        \includegraphics[width=0.309\linewidth]{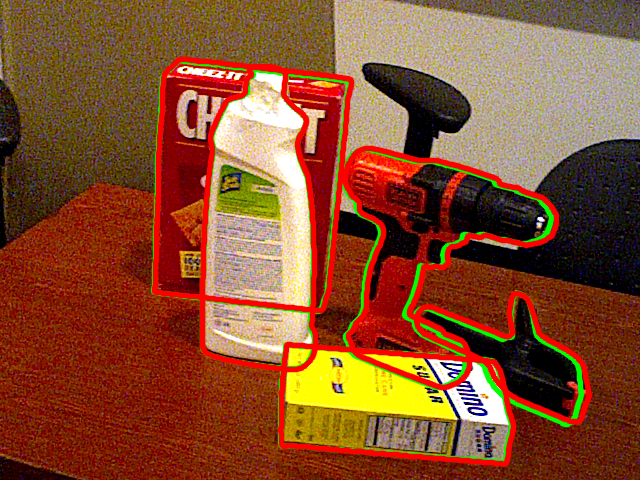} \\

        \raisebox{25pt}{\rotatebox[origin=l]{90}{HOT3D}}\hspace{1pt} &
        \includegraphics[width=0.309\linewidth]{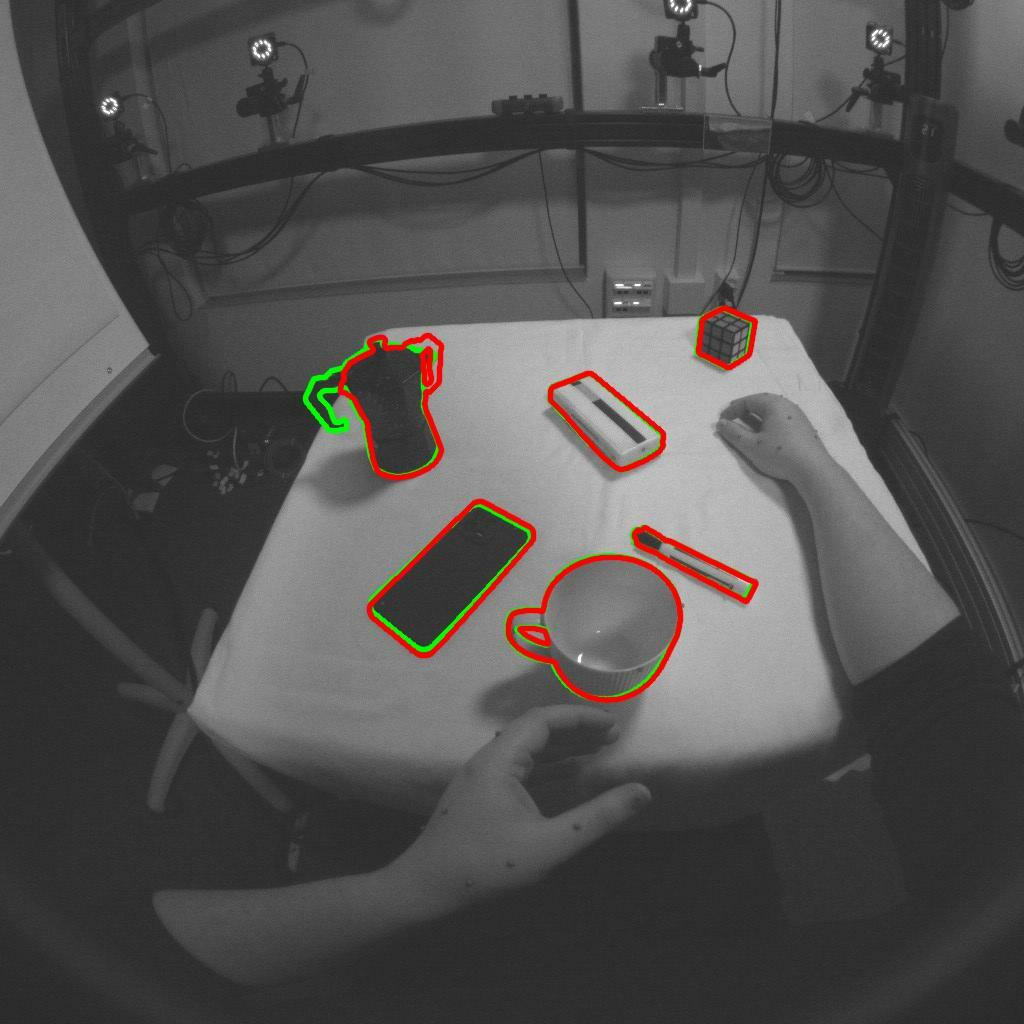} &
        \includegraphics[width=0.309\linewidth]{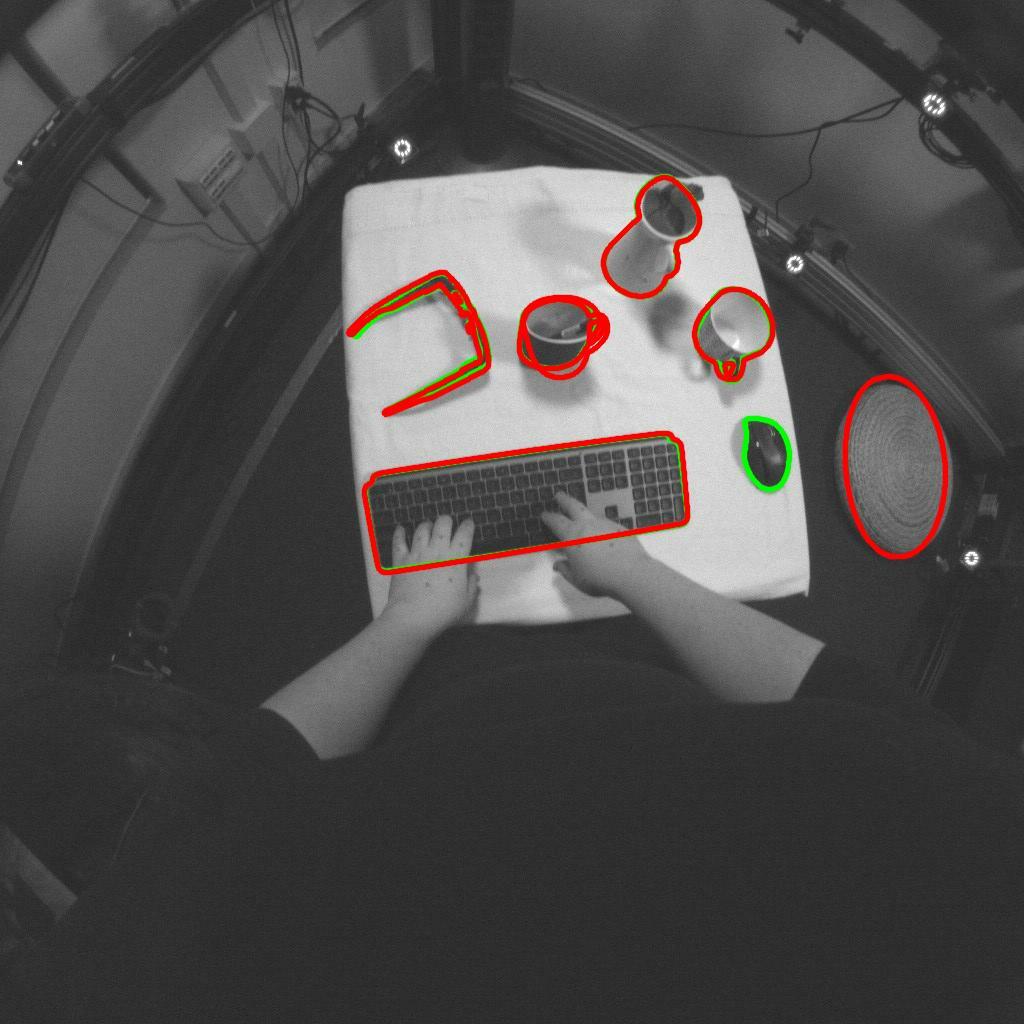} &
        \includegraphics[width=0.309\linewidth]{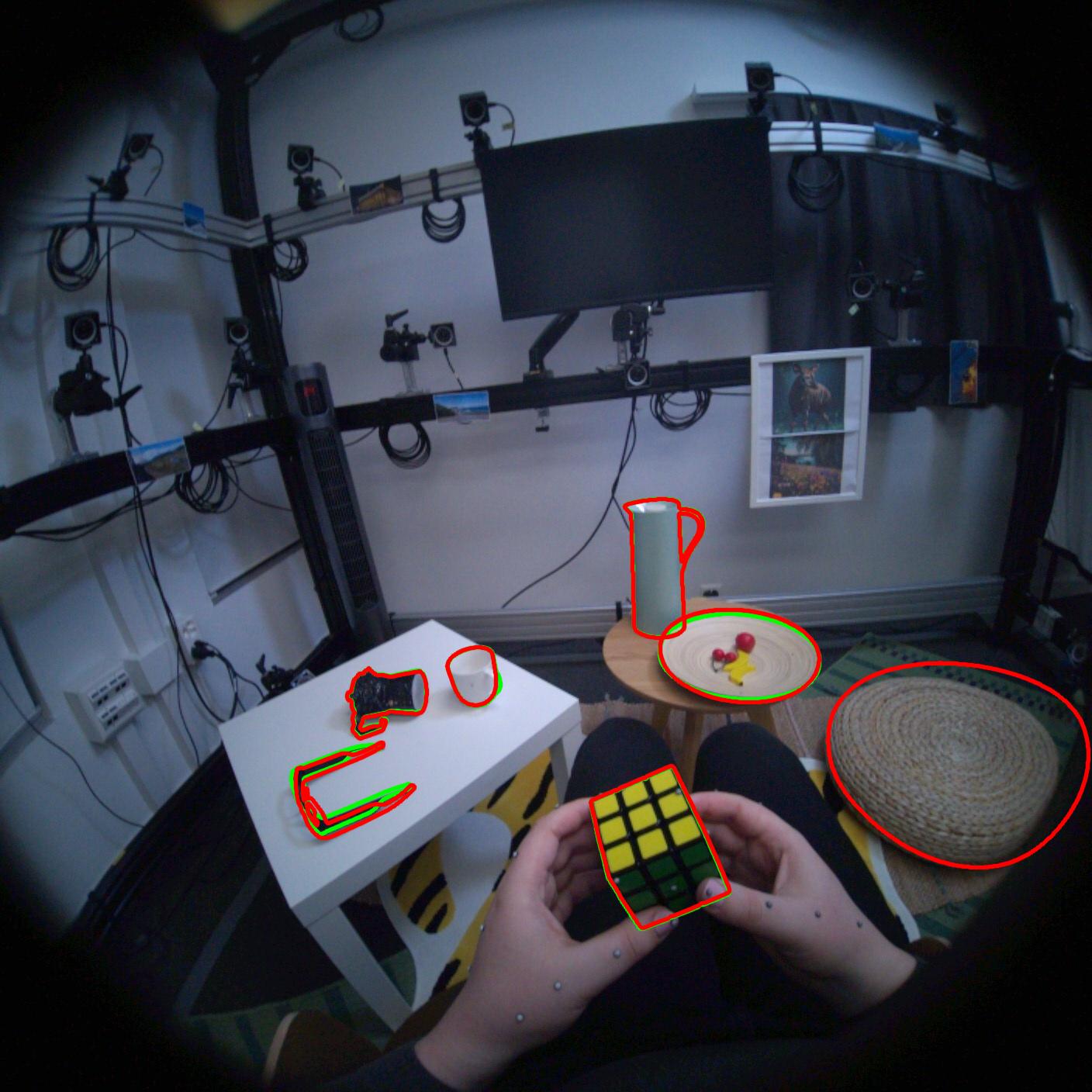} \\

        \raisebox{11pt}{\rotatebox[origin=l]{90}{HOPEv2}}\hspace{1pt} &
        \includegraphics[width=0.309\linewidth, trim= 300 0 0 0, clip]{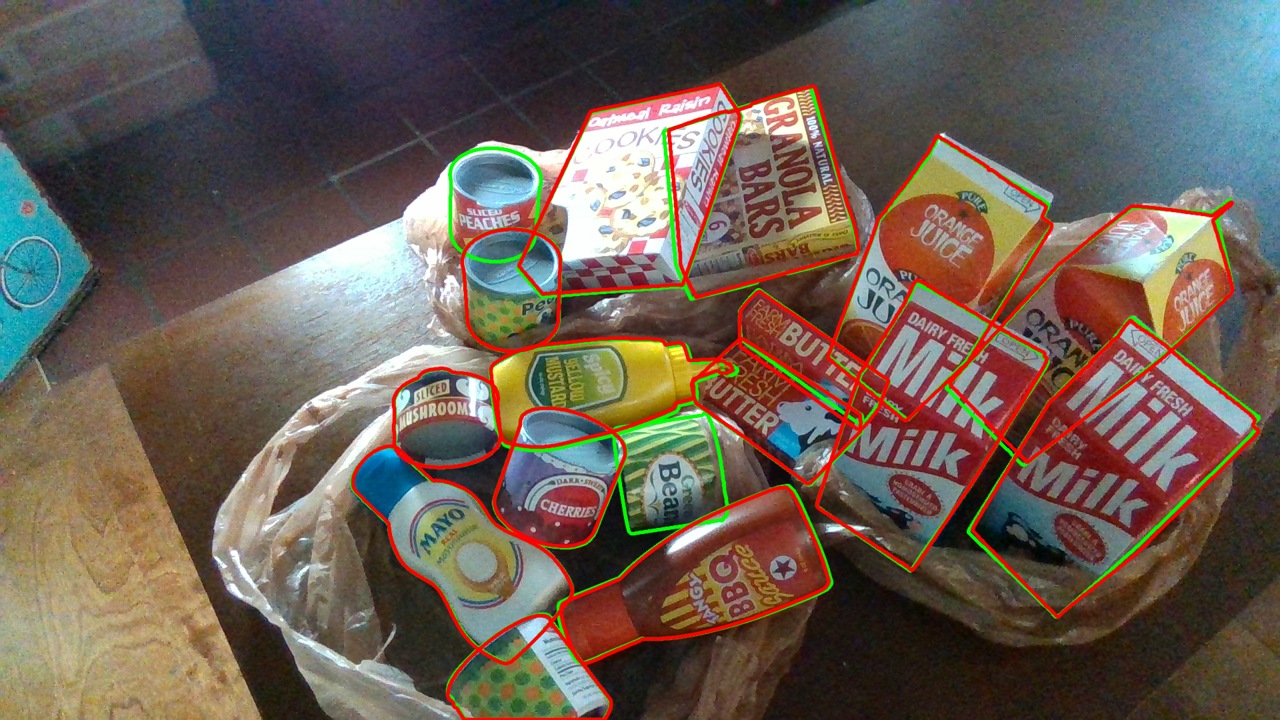} &
        \includegraphics[width=0.309\linewidth, trim= 300 0 0 0, clip]{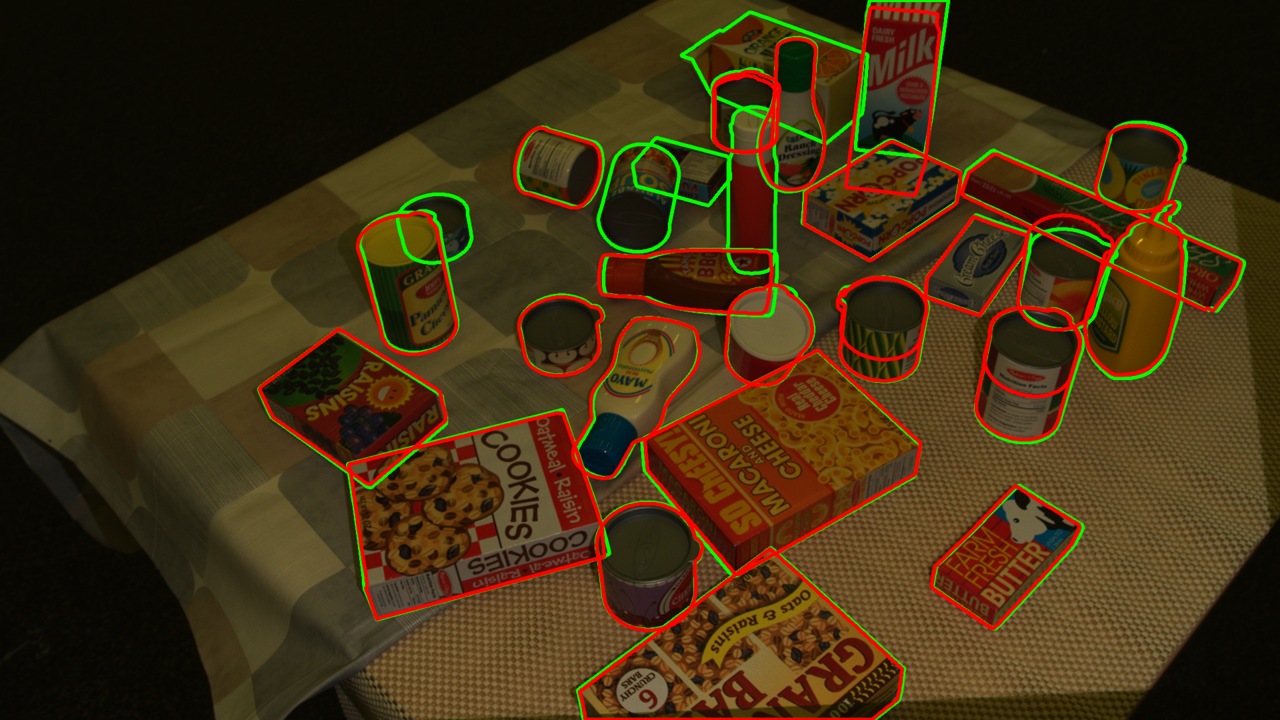} &
        \includegraphics[width=0.309\linewidth, trim= 0 0 300 0, clip]{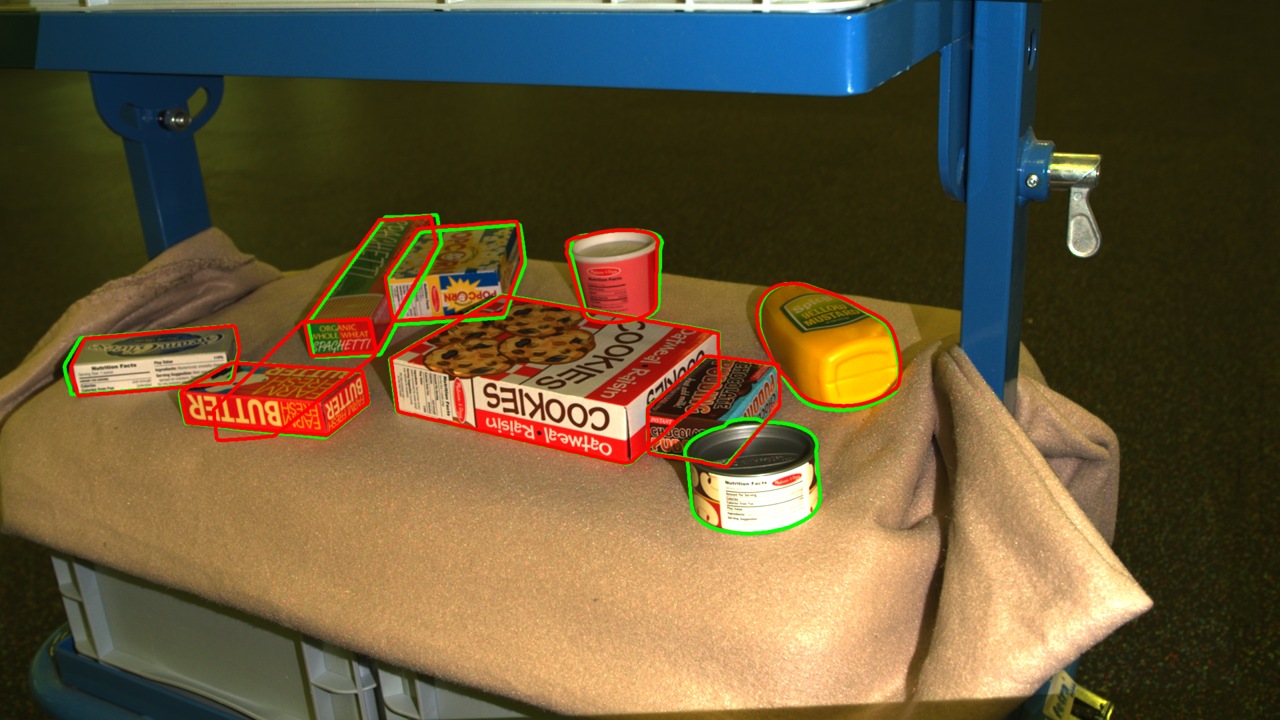} \\

        \raisebox{11pt}{\rotatebox[origin=l]{90}{HANDAL}}\hspace{1pt} &
        \includegraphics[width=0.309\linewidth]{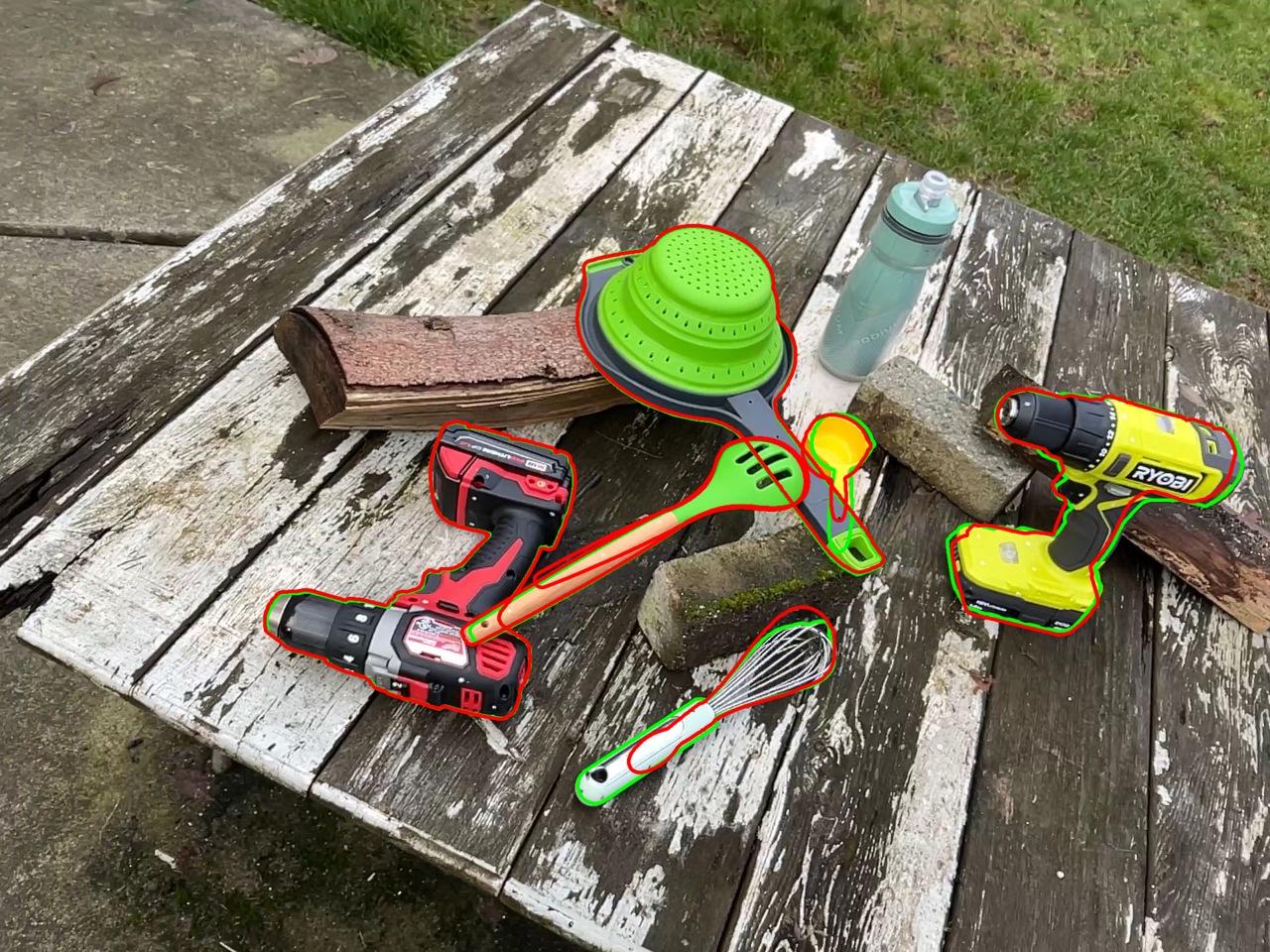} &
        \includegraphics[width=0.309\linewidth]{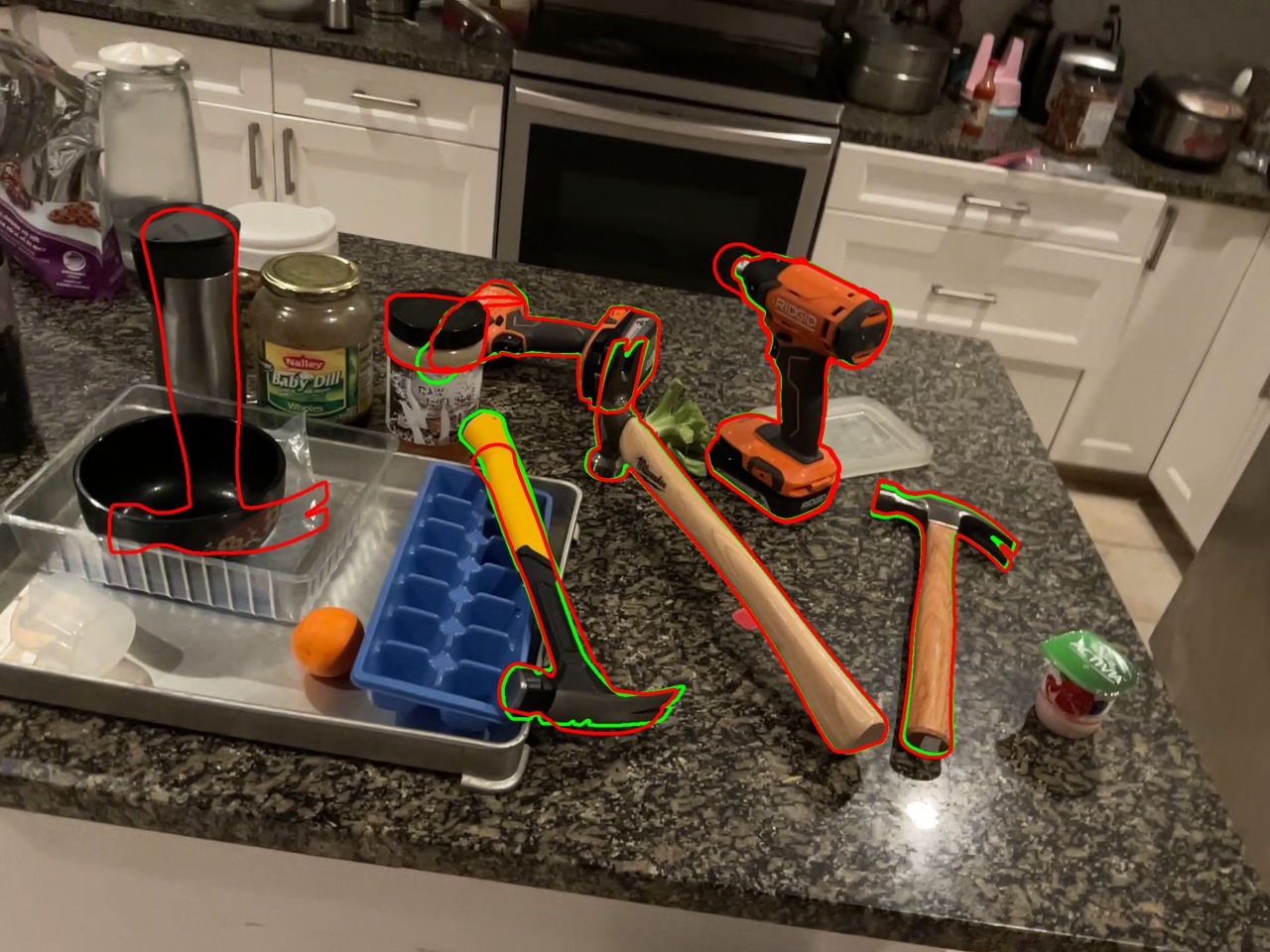} &
        \includegraphics[width=0.309\linewidth]{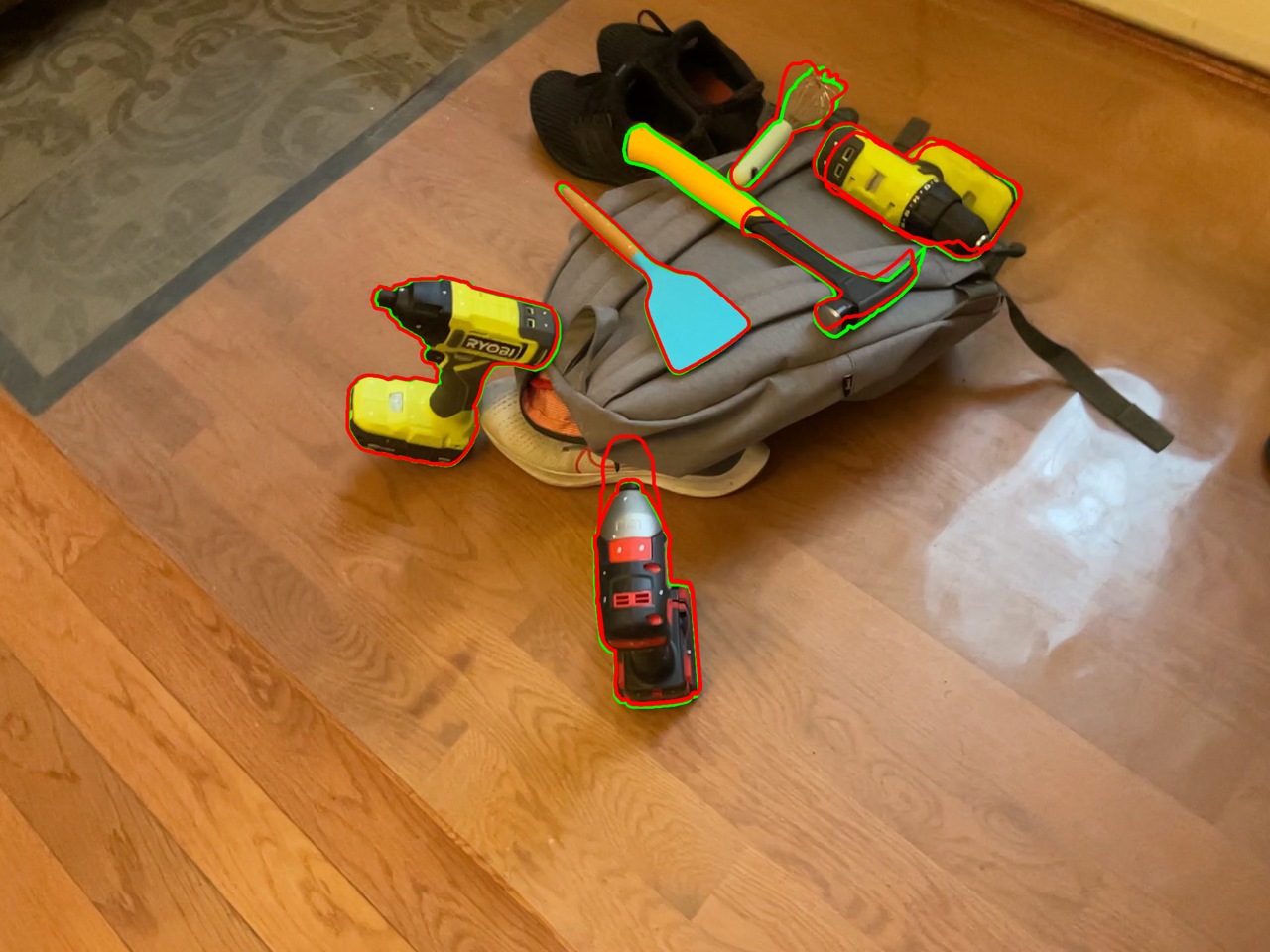}
    \end{tabular}
   \caption{
       \textbf{Results of the top methods for model-based 6D detection:} FreeZeV2.1~\cite{freeze} on BOP-Classic-Core (first 7 rows), and GigaPose~\cite{gigaPose} with GenFlow refinement~\cite{genflow} on BOP-H3 (last 3 rows).
       The methods perform well across diverse datasets, despite not being trained on them.
       Contours of 3D models in the ground-truth and estimated poses are in green and red respectively. Shown are estimates with confidence $\geq 0.3$.
       } 
       \label{fig:qualitative}
\end{figure}

\subsection{Experimental setup}

A method had to use a fixed set of hyperparameters across all objects and datasets. In the model-based onboarding, a method could render images of the 3D models or use a subset of the BlenderProc images (originally provided for BOP 2020~\cite{hodan2020bop}), assuming that rendering a single BlenderProc image takes 2 seconds.
The onboarding phase had to be done within 5 minutes on a single GPU.
Note that methods marked as RGB-only could only use RGB/monochrome channels for onboarding and inference.
Not a single pixel of test images may have been used for training and onboarding, nor ground-truth annotations that are available for test images of some datasets.
Ranges of the azimuth and elevation camera angles, and the range of camera-object distances determined by the ground-truth poses from test images are the only information about the test set that may have been used during training and onboarding.
Only subsets of test images were used (see Tab.~\ref{tab:dataset_params}) to remove redundancies and speed up the evaluation.

\subsection{Model-based 6D localization of unseen objects}
\label{sec:results_track_1}
Among the 44 new entries from 2024 for model-based 6D object localization on BOP-Classic-Core (Tab.~\ref{tab:track1}), 20 entries (19 RGB-D, 1 RGB) outperform the best method from 2023, an RGB-D variant of GenFlow~\cite{genflow} (\#21 in Tab.~\ref{tab:track1}). The best method from 2024, FreeZeV2.1~\cite{freeze}, is 22\% more accurate (82.1 \vs 67.4\,AR) and 39\% faster than GenFlow.
Notably, FreeZeV2.1 is only 3.5 AR behind GPose2023, currently the best method for seen objects.

FreeZeV2.1 is an RGB-D method which, in the winning variant, relies on 2D segmentation masks from three methods: SAM6D~\cite{lin2023sam}, NIDS~\cite{lu2024adapting}, and CNOS~\cite{nguyen2023cnos}. FreeZeV2.1 estimates a 6D pose with a confidence score for each mask and keeps only the best estimates by applying non-maximum suppression.
 Given a segmentation mask, the pose is estimated in a RANSAC-fitting scheme from 3D-3D correspondences established by matching features (2D visual features from DINOv2~\cite{oquab2023dinov2} concatenated with 3D geometry features from GeDi~\cite{poiesi2022learning}) between the query image and the 3D model (the features are extracted from renderings of the model and aggregated on its surface).
Pose estimates are further refined by depth-based ICP and symmetry-aware refinement.

Another notable method is FRTPose.v1, ranked 2nd, which quickly trains a ResNet34 network (4.3 min per object on RTX 4090) at onboarding. Poses are estimated from predicted dense 2D-3D correspondences
 and refined by FoundationPose~\cite{foundationPose}.

A more practical method is Co-op~\cite{Moon2025Coop} (\#9 in Tab.~\ref{tab:track1}), which is 25 times faster (0.8\,s per image) and 13\% more accurate than GenFlow~\cite{genflow}. Co-op estimates poses with a template-based approach, followed by a flow-based refinement similar to GenFlow.

\setlength{\tabcolsep}{2pt}
\begin{table}[t!]
    \renewcommand{\arraystretch}{0.95}
    \tiny
    \centering
    \begin{tabularx}{\linewidth}{rlclllllYYYYY}
        \toprule
        \# & Method & Awards\hspace{-1pt} & Year\hspace{-1pt} & Det./seg. & Refine. & Train im.\hspace{-2pt} & ...type\hspace{-1pt} & \rotatebox{45}{HOT3D}  & \rotatebox{45}{HOPEv2} & \rotatebox{45}{HANDAL} & \rotatebox{45}{AP} & \rotatebox{45}{Time}  \\
        \midrule

        1 & GigaPose+GenFlow~\cite{gigaPose,genflow}\hspace{-4pt}   
        & \iconBest & \cellcolor{ccol!100}2024 & \cellcolor{ccol!100}CNOS-FastSAM & GenFlow &  RGB-D & \cellcolor{ccol!100}PBR & \cellcolor{arcol!26.8}26.8 & \cellcolor{arcol!41.1}41.1 & \cellcolor{arcol!25.6}25.6 & \cellcolor{avgcol!31.2}31.2 & \cellcolor{timecol!5.3}5.3 \\

        2 & GigaPose~\cite{gigaPose} & \iconFast \, \iconOpen & \cellcolor{ccol!100}2024 & \cellcolor{ccol!100}CNOS-FastSAM & -- &  \cellcolor{ccol!100}RGB & \cellcolor{ccol!100}PBR & \cellcolor{arcol!7.2}$\phantom{0}$7.2 & \cellcolor{arcol!16.7}16.7 & \cellcolor{arcol!4.1}$\phantom{0}$4.1 & \cellcolor{avgcol!9.4}$\phantom{0}$9.4 & \cellcolor{timecol!0.9}0.9 \\

        3 & OPFormer-MegaPose & & \cellcolor{ccol!100}2024 & \cellcolor{ccol!100}CNOS-FastSAM & \cellcolor{ccol!100}MegaPose\hspace{-1pt} &  - & - & - & \cellcolor{arcol!39.2}39.2 & \cellcolor{arcol!26.2}26.2 & - & - \\

        4 & OPFormer-Coarse &  & \cellcolor{ccol!100}2024 & \cellcolor{ccol!100}CNOS-FastSAM & - &  - & - & - & \cellcolor{arcol!35.1}35.1 & \cellcolor{arcol!19.2}19.2 & - & -\\

        \bottomrule

    \end{tabularx}
    \vspace{-3pt}
    \caption{\textbf{Track 4: Model-based 6D det.\ of unseen objects on BOP-H3.}
    }
    \label{tab:track4}
\end{table}

\setlength{\tabcolsep}{3pt}
\begin{table}[t!]
    \renewcommand{\arraystretch}{0.95}
    \tiny
    \centering
    \begin{tabularx}{\linewidth}{rlclllYYYYY}
        \toprule
        \# & Method & Awards & Year  & Onboarding im. & ...type & \rotatebox{45}{HOT3D}  & \rotatebox{45}{HOPEv2} & \rotatebox{45}{HANDAL} & \rotatebox{0}{AP} & \rotatebox{0}{Time}  \\
        \midrule

        1 & MUSE  & \iconBest & \cellcolor{ccol!100}2024 & N/A & N/A & \cellcolor{arcol!42.6}42.6 & \cellcolor{arcol!47.4}47.4 & \cellcolor{arcol!27.0}27.0 & \cellcolor{avgcol!39.0}39.0 & \cellcolor{timecol!1.5}1.5 \\

        2 & CNOS (FastSAM)~\cite{nguyen2023cnos}  & \iconFast \; \iconOpen & \cellcolor{ccol!100}2024 & \cellcolor{ccol!100}RGB & \cellcolor{ccol!100}PBR & \cellcolor{arcol!35.0}35.0 & \cellcolor{arcol!31.3}31.3 & \cellcolor{arcol!24.6}24.6 & \cellcolor{avgcol!30.3}30.3 & \cellcolor{timecol!0.3}0.3 \\

        3 & CNOS (SAM)~\cite{nguyen2023cnos} &  & \cellcolor{ccol!100}2024 & \cellcolor{ccol!100}RGB & \cellcolor{ccol!100}PBR & \cellcolor{arcol!31.7}31.7 & \cellcolor{arcol!36.5}36.5 & \cellcolor{arcol!19.7}19.7 & \cellcolor{avgcol!29.3}29.3 & \cellcolor{timecol!1.8}1.8 \\

        \bottomrule

    \end{tabularx}
    \vspace{-3pt}
    \caption{\textbf{Track 5: Model-based 2D det.\ of unseen objects on BOP-H3.}
    }
    \label{tab:track5}
\end{table}

\setlength{\tabcolsep}{3pt}
\begin{table}[t!]
    \renewcommand{\arraystretch}{0.95}
    \tiny
    \centering
    \begin{tabularx}{\linewidth}{rlcllYYYYY}
        \toprule
        \# & Method & Awards & Year & Onboarding type & \rotatebox{0}{HOT3D}  & \rotatebox{0}{HOPEv2} & \rotatebox{0}{HANDAL} & \rotatebox{0}{AP} & \rotatebox{0}{Time} \\
        \midrule

        1 & GFreeDet (FastSAM)~\cite{liu2024gfreedet} & \iconBest \; \iconFast & \cellcolor{ccol!100}2024 & Static & \cellcolor{arcol!33.8}33.8 & \cellcolor{arcol!36.4}36.4 & \cellcolor{arcol!25.5}25.5 & \cellcolor{avgcol!31.9}31.9 & \cellcolor{timecol!0.3}0.3 \\

        2 & GFreeDet (SAM)~\cite{liu2024gfreedet} & & \cellcolor{ccol!100}2024 & Static & \cellcolor{arcol!30.9}30.9 & \cellcolor{arcol!38.4}38.4 & \cellcolor{arcol!26.4}26.4 & \cellcolor{avgcol!31.9}31.9 & \cellcolor{timecol!2.1}2.1 \\

        \bottomrule

    \end{tabularx}
    \vspace{-3pt}
    \caption{\textbf{Track 7: Model-free 2D det.\ of unseen objects on BOP-H3.}
    }
    \label{tab:track7}
\end{table}

\subsection{Model-based 6D detection of unseen objects}
\label{sec:results_track_1}
\noindent\textbf{BOP-Classic-Core (Tab.~\ref{tab:track2}).} Ranking of methods on Track 2 is consistent with Track 1, suggesting that the 6D detection and 6D localization tasks present similar difficulties and that learnings from previous BOP challenges, focused on the 6D localization task, are also relevant for the newly introduced and more practical 6D detection task.
On both tracks, FreeZeV2.1 is the best method,
although the improvement of FreeZeV2.1 over Co-op is more significant on Track 2 than on Track 1: 18.8\% (\#1 and \#4 in Tab.~\ref{tab:track2}) \vs only 6.5\% (\#1 and \#8 in Tab.~\ref{tab:track1}).
Adding refinement to Co-op tends to decrease the 6D detection score (see coarse poses on \#6 in Tab.~\ref{tab:track2} \vs \#7 with single- and \#8 with multi-hypotheses refinement). This is a surprising result that is worth investigating.

\customparagraph{BOP-H3 (Tab.~\ref{tab:track4}).}
The best entry is GigaPose~\cite{gigaPose}, a coarse pose estimation method using 2D-2D correspondences established between the input image and pre-rendered templates, followed by the GenFlow~\cite{genflow} refinement.
GigaPose and GigaPose+GenFlow achieve 12.3 and 50.4\,AP on BOP-Classic-Core (\#17 and \#15 in Tab.~\ref{tab:track2}), while only 9.4 and 31.2\,AP on BOP-H3, suggesting that BOP-H3 is more difficult.
Another method on this track is OPFormer, which follows a similar pipeline as FoundPose~\cite{foundPose}, but predicts correspondences using a trained transformer head instead of relying on a simple kNN matching of DINOv2 features~\cite{oquab2023dinov2}.

\subsection{Model-based 2D detection of unseen objects}
\label{sec:results_track_3}

\noindent\textbf{BOP-Classic-Core (Tab.~\ref{tab:track3}).} The best 2024 method, MUSE, follows a similar pipeline as the best 2023 method, CNOS~\cite{nguyen2023cnos}: detection proposals are generated using SAM2~\cite{ravi2024sam} and Grounding DINO~\cite{liu2024grounding} and then matched to templates using DINOv2~\cite{oquab2023dinov2}.
The key difference is in the matching stage, where MUSE introduces a novel similarity metric that leverages both class and patch tokens, whereas CNOS only uses class tokens.
MUSE is 21\% more accurate than CNOS
(\#1 \vs \#10 in Tab.~\ref{tab:track3}),
however, MUSE is still -35\% behind the best method for seen objects (GDet2023).

\customparagraph{BOP-H3 (Tab.~\ref{tab:track5}).}
MUSE achieves a noticeably lower score on BOP-H3 than on BOP-Classic-Core (39.0 \vs 52.0\,AP), while still outperforming CNOS by 29\% (\#1 \vs \#2 in Tab.~\ref{tab:track5}).

\subsection{Model-free 2D detection of unseen objects} \label{sec:results_track_7}

The only entry, GFreeDet~\cite{liu2024gfreedet}, reconstructs a 3DGS model~\cite{kerbl20233d} from onboarding videos and renders templates from equidistantly sampled viewpoints. At inference, GFreeDet uses SAM~\cite{kirillov2023segment} to generate proposals and DINOv2~\cite{oquab2023dinov2} for proposal-template matching.
Notably, GFreeDet outperforms the model-based CNOS on HOPEv2 and HANDAL (\#1 in Tab.~\ref{tab:track7} \vs \#2 Tab.~\ref{tab:track5}).

\section{Awards} \label{sec:awards}

The 2024 challenge awards are based on the results analyzed in Sec.~\ref{sec:evaluation}. Methods receiving awards are marker with icons: \iconBest\; for the best overall method, \iconFast\; for the best fast method (the most accurate method with the average running time per image below 1\,s), \iconOpen\; for the best open-source method, \iconRGB\; for the best RGB-only method, and \iconDefault\; for the best method using default detections.

\vspace{1ex}
\noindent Authors of awarded entries:
\textbf{FreeZeV2.1}~\cite{freeze}~by Andrea Caraffa, Davide Boscaini, Amir Hamza, Fabio Poiesi;
\textbf{Co-op}~\cite{Moon2025Coop} and \textbf{GenFlow}~\cite{genflow} by Sungphill Moon, Hyeontae Son, Dongcheol Hur, Sangwook Kim;
\textbf{FoundationPose}~\cite{foundationPose} by Bowen Wen, Wei Yang, Jan Kautz, Stan Birchfield;
\textbf{GigaPose}~\cite{gigaPose} by Van Nguyen Nguyen, Thibault Groueix, Mathieu Salzmann, Vincent Lepetit;
\textbf{SAM6D}~\cite{lin2023sam} by Jiehong Lin, Lihua Liu, Dekun Lu, Kui Jia;
\textbf{CNOS}~\cite{nguyen2023cnos} by Van Nguyen Nguyen, Thibault Groueix, Georgy Ponimatkin, Vincent Lepetit, Tomas Hodan;
\textbf{GFreeDet}~\cite{liu2024gfreedet} by Xingyu Liu, Yingyue Li, Chengxi Li, Gu Wang, Chenyangguang Zhang, Ziqin Huang, Xiangyang Ji;
\textbf{FRTPose.v1} and \textbf{MUSE} are anonymous for now.

\section{Conclusions} \label{sec:conclusion}

In 2024, methods for 6D localization of unseen objects have almost reached the accuracy of methods for seen objects. While some of these methods now take less than 1 second per image, further speed up is needed for real-time applications. 
Rankings on 6D localization and 6D detection are consistent, suggesting that learnings from past BOP challenges
are relevant also for the newly introduced and more practical 6D detection task. 2D detection of unseen objects has improved significantly, but is still noticeably behind 2D detection of seen objects.
The evaluation system
stays open and awaits new submissions on the model-free tasks.

{\small
\bibliographystyle{ieee_fullname}
\bibliography{references}

\begin{thebibliography}{10}\itemsep=-1pt

\bibitem{banerjee2024hot3d}
Prithviraj Banerjee, Sindi Shkodrani, Pierre Moulon, Shreyas Hampali, Shangchen Han, Fan Zhang, Linguang Zhang, Jade Fountain, Edward Miller, Selen Basol, Richard Newcombe, Robert Wang, Jakob~Julian Engel, and Tomas Hodan.
\newblock {HOT3D}: Hand and object tracking in {3D} from egocentric multi-view videos.
\newblock {\em CVPR}, 2025.

\bibitem{brachmann2014learning}
Eric Brachmann, Alexander Krull, Frank Michel, Stefan Gumhold, Jamie Shotton, and Carsten Rother.
\newblock Learning {6D} object pose estimation using {3D} object coordinates.
\newblock In {\em ECCV}, 2014.

\bibitem{freeze}
Andrea Caraffa, Davide Boscaini, Amir Hamza, and Fabio Poiesi.
\newblock Freeze: Training-free zero-shot 6d pose estimation with geometric and vision foundation models.
\newblock {\em ECCV}, 2024.

\bibitem{chang2015shapenet}
Angel~X Chang, Thomas Funkhouser, Leonidas Guibas, Pat Hanrahan, Qixing Huang, Zimo Li, Silvio Savarese, Manolis Savva, Shuran Song, Hao Su, et~al.
\newblock Shapenet: An information-rich {3D} model repository.
\newblock {\em arXiv preprint arXiv:1512.03012}, 2015.

\bibitem{chen20233d}
Jianqiu Chen, Mingshan Sun, Tianpeng Bao, Rui Zhao, Liwei Wu, and Zhenyu He.
\newblock 3d model-based zero-shot pose estimation pipeline.
\newblock {\em arXiv preprint arXiv:2305.17934}, 2023.

\bibitem{denninger2020blenderproc}
Maximilian Denninger, Martin Sundermeyer, Dominik Winkelbauer, Dmitry Olefir, Tom{\'a}{\v{s}} Hoda{\v{n}}, Youssef Zidan, Mohamad Elbadrawy, Markus Knauer, Harinandan Katam, and Ahsan Lodhi.
\newblock {BlenderProc:} {R}educing the reality gap with photorealistic rendering.
\newblock {\em RSS Workshops}, 2020.

\bibitem{denninger2019blenderproc}
Maximilian Denninger, Martin Sundermeyer, Dominik Winkelbauer, Youssef Zidan, Dmitry Olefir, Mohamad Elbadrawy, Ahsan Lodhi, and Harinandan Katam.
\newblock Blenderproc.
\newblock {\em arXiv preprint arXiv:1911.01911}, 2019.

\bibitem{denninger2023blenderproc2}
Maximilian Denninger, Dominik Winkelbauer, Martin Sundermeyer, Wout Boerdijk, Markus~Wendelin Knauer, Klaus~H Strobl, Matthias Humt, and Rudolph Triebel.
\newblock Blenderproc2: A procedural pipeline for photorealistic rendering.
\newblock {\em Journal of Open Source Software}, 8(82):4901, 2023.

\bibitem{doumanoglou2016recovering}
Andreas Doumanoglou, Rigas Kouskouridas, Sotiris Malassiotis, and Tae-Kyun Kim.
\newblock Recovering {6D} object pose and predicting next-best-view in the crowd.
\newblock In {\em CVPR}, 2016.

\bibitem{downs2022google}
Laura Downs, Anthony Francis, Nate Koenig, Brandon Kinman, Ryan Hickman, Krista Reymann, Thomas~B McHugh, and Vincent Vanhoucke.
\newblock Google scanned objects: A high-quality dataset of {3D} scanned household items.
\newblock {\em ICRA}, 2022.

\bibitem{drost2017introducing}
Bertram Drost, Markus Ulrich, Paul Bergmann, Philipp Hartinger, and Carsten Steger.
\newblock Introducing {MVTec} {ITODD} -- {A} dataset for {3D} object recognition in industry.
\newblock In {\em ICCVW}, 2017.

\bibitem{guo2023handal}
Andrew Guo, Bowen Wen, Jianhe Yuan, Jonathan Tremblay, Stephen Tyree, Jeffrey Smith, and Stan Birchfield.
\newblock Handal: A dataset of real-world manipulable object categories with pose annotations, affordances, and reconstructions.
\newblock In {\em IROS}, 2023.

\bibitem{hampali2023inhand}
Shreyas Hampali, Tom{\'a}{\v{s}} Hoda{\v{n}}, Luan Tran, Lingni Ma, Cem Keskin, and Vincent Lepetit.
\newblock In-hand {3D} object scanning from an {RGB} sequence.
\newblock {\em Conference on Computer Vision and Pattern Recognition (CVPR)}, 2023.

\bibitem{han2022umetrack}
Shangchen Han, Po-chen Wu, Yubo Zhang, Beibei Liu, Linguang Zhang, Zheng Wang, Weiguang Si, Peizhao Zhang, Yujun Cai, Tomas Hodan, et~al.
\newblock Umetrack: Unified multi-view end-to-end hand tracking for vr.
\newblock {\em SIGGRAPH Asia}, 2022.
\newblock \url{https://github.com/facebookresearch/hand_tracking_toolkit?tab=readme-ov-file\#evaluation}.

\bibitem{hinterstoisser2012accv}
S. Hinterstoisser, V. Lepetit, S. Ilic, S. Holzer, G. Bradski, K. Konolige, and N. Navab.
\newblock Model based training, detection and pose estimation of texture-less 3{D} objects in heavily cluttered scenes.
\newblock {\em ACCV}, 2012.

\bibitem{hodan2021phd}
Tom{\'a}{\v{s}} Hoda{\v{n}}.
\newblock Pose estimation of specific rigid objects.
\newblock {\em PhD Thesis, Czech Technical University in Prague}, 2021.

\bibitem{hodan2019bop}
Tom{\'a}{\v{s}} Hoda{\v{n}}, Eric Brachmann, Bertram Drost, Frank Michel, Martin Sundermeyer, Ji{\v{r}}{\'\i} Matas, and Carsten Rother.
\newblock {BOP} {C}hallenge 2019.
\newblock \url{https://bop.felk.cvut.cz/media/bop_challenge_2019_results.pdf}, 2019.

\bibitem{hodan2017tless}
Tom{\'a}{\v{s}} Hoda{\v{n}}, Pavel Haluza, {\v{S}}t{\v{e}}p{\'a}n Obdr{\v{z}}{\'a}lek, Ji{\v{r}}{\'\i} Matas, Manolis Lourakis, and Xenophon Zabulis.
\newblock {T-LESS}: {A}n {RGB-D} dataset for {6D} pose estimation of texture-less objects.
\newblock {\em WACV}, 2017.

\bibitem{hodan2018bop}
Tom{\'a}{\v{s}} Hoda{\v{n}}, Frank Michel, Eric Brachmann, Wadim Kehl, Anders Glent~Buch, Dirk Kraft, Bertram Drost, Joel Vidal, Stephan Ihrke, Xenophon Zabulis, Caner Sahin, Fabian Manhardt, Federico Tombari, Tae-Kyun Kim, Ji{\v{r}}{\'i} Matas, and Carsten Rother.
\newblock {BOP}: {B}enchmark for {6D} object pose estimation.
\newblock {\em ECCV}, 2018.

\bibitem{hodan2020bop}
Tom{\'a}{\v{s}} Hoda{\v{n}}, Martin Sundermeyer, Bertram Drost, Yann Labb{\'e}, Eric Brachmann, Frank Michel, Carsten Rother, and Ji{\v{r}}{\'\i} Matas.
\newblock {BOP Challenge} 2020 on {6D} object localization.
\newblock In {\em ECCV}, 2020.

\bibitem{hodan2023bop}
Tom{\'a}{\v{s}} Hoda{\v{n}}, Martin Sundermeyer, Yann Labb{\'e}, Van~Nguyen Nguyen, Gu Wang, Eric Brachmann, Bertram Drost, Vincent Lepetit, Carsten Rother, and Ji{\v{r}}{\'i} Matas.
\newblock {BOP} challenge 2023 on detection, segmentation and pose estimation of seen and unseen rigid objects.
\newblock {\em CVPRW}, 2024.

\bibitem{kaskman2019homebreweddb}
Roman Kaskman, Sergey Zakharov, Ivan Shugurov, and Slobodan Ilic.
\newblock {HomebrewedDB}: {RGB-D} dataset for {6D} pose estimation of {3D} objects.
\newblock {\em ICCVW}, 2019.

\bibitem{kerbl20233d}
Bernhard Kerbl, Georgios Kopanas, Thomas Leimk{\"u}hler, and George Drettakis.
\newblock {3D} gaussian splatting for real-time radiance field rendering.
\newblock {\em ACM Trans. Graph.}, 2023.

\bibitem{kirillov2023segment}
Alexander Kirillov, Eric Mintun, Nikhila Ravi, Hanzi Mao, Chloe Rolland, Laura Gustafson, Tete Xiao, Spencer Whitehead, Alexander~C Berg, Wan-Yen Lo, et~al.
\newblock Segment anything.
\newblock In {\em ICCV}, 2023.

\bibitem{megapose}
Yann Labb\'e, Lucas Manuelli, Arsalan Mousavian, Stephen Tyree, Stan Birchfield, Jonathan Tremblay, Justin Carpentier, Mathieu Aubry, Dieter Fox, and Josef Sivic.
\newblock {{MegaPose}: {6D} Pose Estimation of Novel Objects via Render \& Compare}.
\newblock In {\em CoRL}, 2022.

\bibitem{lin2023sam}
Jiehong Lin, Lihua Liu, Dekun Lu, and Kui Jia.
\newblock Sam-6d: Segment anything model meets zero-shot 6d object pose estimation.
\newblock In {\em CVPR}, 2024.

\bibitem{lin2014microsoft}
Tsung-Yi Lin, Michael Maire, Serge Belongie, James Hays, Pietro Perona, Deva Ramanan, Piotr Doll{\'a}r, and C~Lawrence Zitnick.
\newblock Microsoft {COCO}: {C}ommon objects in context.
\newblock {\em ECCV}, 2014.

\bibitem{liu2024grounding}
Shilong Liu, Zhaoyang Zeng, Tianhe Ren, Feng Li, Hao Zhang, Jie Yang, Qing Jiang, Chunyuan Li, Jianwei Yang, Hang Su, et~al.
\newblock Grounding dino: Marrying dino with grounded pre-training for open-set object detection.
\newblock In {\em European Conference on Computer Vision}, pages 38--55. Springer, 2024.

\bibitem{liu2024gfreedet}
Xingyu Liu, Yingyue Li, Chengxi Li, Gu Wang, Chenyangguang Zhang, Ziqin Huang, and Xiangyang Ji.
\newblock Gfreedet: Exploiting gaussian splatting and foundation models for model-free unseen object detection in the bop challenge 2024.
\newblock {\em arXiv preprint arXiv:2412.01552}, 2024.

\bibitem{lu2024adapting}
Yangxiao Lu, Yunhui Guo, Nicholas Ruozzi, Yu Xiang, et~al.
\newblock Adapting pre-trained vision models for novel instance detection and segmentation.
\newblock {\em arXiv preprint arXiv:2405.17859}, 2024.

\bibitem{mildenhall2021nerf}
Ben Mildenhall, Pratul~P Srinivasan, Matthew Tancik, Jonathan~T Barron, Ravi Ramamoorthi, and Ren Ng.
\newblock {NeRF}: Representing scenes as neural radiance fields for view synthesis.
\newblock {\em Communications of the ACM}, 2021.

\bibitem{genflow}
Sungphill Moon, Hyeontae Son, Dongcheol Hur, and Sangwook Kim.
\newblock {GenFlow: Generalizable Recurrent Flow for 6D Pose Refinement of Novel Objects}.
\newblock In {\em arXiv preprint arXiv:2403.11510}, 2024.

\bibitem{Moon2025Coop}
Sungphill Moon, Hyeontae Son, Dongcheol Hur, and Sangwook Kim.
\newblock Co-op: Correspondence-based novel object pose estimation.
\newblock {\em arXiv preprint arXiv:2503.17731}, 2025.

\bibitem{newcombe2011kinectfusion}
Richard~A Newcombe, Shahram Izadi, Otmar Hilliges, David Molyneaux, David Kim, Andrew~J Davison, Pushmeet Kohi, Jamie Shotton, Steve Hodges, and Andrew Fitzgibbon.
\newblock {K}inect{F}usion: {R}eal-time dense surface mapping and tracking.
\newblock {\em ISMAR}, 2011.

\bibitem{nguyen2023cnos}
Van~Nguyen Nguyen, Thibault Groueix, Georgy Ponimatkin, Vincent Lepetit, and Tomas Hodan.
\newblock {CNOS: A Strong Baseline for CAD-based Novel Object Segmentation}.
\newblock In {\em ICCVW}, 2023.

\bibitem{gigaPose}
Van~Nguyen Nguyen, Thibault Groueix, Mathieu Salzmann, and Vincent Lepetit.
\newblock Gigapose: Fast and robust novel object pose estimation via one correspondence.
\newblock In {\em CVPR}, 2024.

\bibitem{oquab2023dinov2}
Maxime Oquab, Timoth{\'e}e Darcet, Th{\'e}o Moutakanni, Huy Vo, Marc Szafraniec, Vasil Khalidov, Pierre Fernandez, Daniel Haziza, Francisco Massa, Alaaeldin El-Nouby, et~al.
\newblock Dinov2: Learning robust visual features without supervision.
\newblock {\em arXiv preprint arXiv:2304.07193}, 2023.

\bibitem{foundPose}
Evin~P{\i}nar {\"O}rnek, Yann Labb{\'e}, Bugra Tekin, Lingni Ma, Cem Keskin, Christian Forster, and Tomas Hodan.
\newblock Foundpose: Unseen object pose estimation with foundation features.
\newblock In {\em European Conference on Computer Vision}, 2024.

\bibitem{poiesi2022learning}
Fabio Poiesi and Davide Boscaini.
\newblock Learning general and distinctive 3d local deep descriptors for point cloud registration.
\newblock {\em PAMI}, 2022.

\bibitem{ravi2024sam}
Nikhila Ravi, Valentin Gabeur, Yuan-Ting Hu, Ronghang Hu, Chaitanya Ryali, Tengyu Ma, Haitham Khedr, Roman R{\"a}dle, Chloe Rolland, Laura Gustafson, et~al.
\newblock Sam 2: Segment anything in images and videos.
\newblock {\em arXiv preprint arXiv:2408.00714}, 2024.

\bibitem{rennie2016dataset}
Colin Rennie, Rahul Shome, Kostas~E Bekris, and Alberto~F De~Souza.
\newblock A dataset for improved {RGBD}-based object detection and pose estimation for warehouse pick-and-place.
\newblock {\em RA-L}, 2016.

\bibitem{schonberger2016structure}
Johannes~L Schonberger and Jan-Michael Frahm.
\newblock Structure-from-motion revisited.
\newblock In {\em CVPR}, 2016.

\bibitem{sundermeyer2022bop}
Martin Sundermeyer, Tomas Hodan, Yann Labb{\'e}, Gu Wang, Eric Brachmann, Bertram Drost, Carsten Rother, and Jiri Matas.
\newblock {BOP} challenge 2022 on detection, segmentation and pose estimation of specific rigid objects.
\newblock {\em CVPRW}, 2023.

\bibitem{tejani2014latent}
Alykhan Tejani, Danhang Tang, Rigas Kouskouridas, and Tae-Kyun Kim.
\newblock Latent-class hough forests for {3D} object detection and pose estimation.
\newblock {\em ECCV}, 2014.

\bibitem{tyree2022hope}
Stephen Tyree, Jonathan Tremblay, Thang To, Jia Cheng, Terry Mosier, Jeffrey Smith, and Stan Birchfield.
\newblock {6-DoF} pose estimation of household objects for robotic manipulation: {A}n accessible dataset and benchmark.
\newblock {\em IROS}, 2022.

\bibitem{wen2023bundlesdf}
Bowen Wen, Jonathan Tremblay, Valts Blukis, Stephen Tyree, Thomas M{\"u}ller, Alex Evans, Dieter Fox, Jan Kautz, and Stan Birchfield.
\newblock {BundleSDF}: Neural 6-{DoF} tracking and {3D} reconstruction of unknown objects.
\newblock In {\em CVPR}, 2023.

\bibitem{foundationPose}
Bowen Wen, Wei Yang, Jan Kautz, and Stan Birchfield.
\newblock Foundationpose: Unified 6d pose estimation and tracking of novel objects.
\newblock In {\em CVPR}, 2024.

\bibitem{xiang2017posecnn}
Yu Xiang, Tanner Schmidt, Venkatraman Narayanan, and Dieter Fox.
\newblock {PoseCNN:} {A} convolutional neural network for {6D} object pose estimation in cluttered scenes.
\newblock {\em RSS}, 2018.

\bibitem{gpose2023}
Ruida Zhang, Ziqin Huang, Gu Wang, Xingyu Liu, Chenyangguang Zhang, and Xiangyang Ji.
\newblock {{GPose2023}}, a submission to the {{BOP Challenge}} 2023.
\newblock {\em Unpublished}, 2023.
\newblock \url{http://bop.felk.cvut.cz/method_info/410/}.

\end{thebibliography}
}

\end{document}